\documentclass{article}

\usepackage{booktabs} 
\usepackage{bbm}
\usepackage{mathtools}
\usepackage{natbib}
\usepackage{verbatim}
\usepackage[linesnumbered, ruled,vlined, algo2e]{algorithm2e}
\usepackage{algorithm,algorithmic}
\usepackage{enumitem}
\usepackage[dvipsnames]{xcolor}
\setlist[itemize]{topsep=2pt, leftmargin=5mm}

\usepackage{microtype}
\usepackage{subfigure}
\usepackage{booktabs} 
\usepackage{wrapfig}

\usepackage{amsthm,amsmath,amssymb, amsfonts,bm}

\allowdisplaybreaks

\definecolor{darkblue}{RGB}{1, 21, 115}
\usepackage[colorlinks]{hyperref}
\hypersetup{
    colorlinks=magenta,
    linkcolor=magenta,
    filecolor=magenta,      
    urlcolor=blue,
    citecolor=darkblue,
    pdfpagemode=FullScreen,
}   

\newcommand{\method}[0]{ACCEL}
\newcommand{\accmethod}[0]{ACCEL}
\newcommand{\methodlong}[0]{Adversarially Compounding Complexity by Editing Levels}

\newcounter{commentCounter}
\newif\iftrvar
\trvarfalse
\iftrvar
\newcommand{\tim}[1]{{\small \color{red} \refstepcounter{commentCounter}\textsf{[TR]$_{\arabic{commentCounter}}$:{#1}}}}
\newcommand{\minqi}[1]{{\small \color{blue} \refstepcounter{commentCounter}\textsf{[MJ]$_{\arabic{commentCounter}}$:{#1}}}}
\newcommand{\ed}[1]{{\small \color{magenta} \refstepcounter{commentCounter}\textsf{[ETG]$_{\arabic{commentCounter}}$:{#1}}}}

\newcommand{\jcom}[1]{\textcolor{purple}{JPH: #1}}

\newcommand{\michael}[1]{{\small \color{cyan} \refstepcounter{commentCounter}\textsf{[MDD]$_{\arabic{commentCounter}}$:{#1}}}}

\newcommand{\jakob}[1]{{\small \color{green} \refstepcounter{commentCounter}\textsf{[JF]$_{\arabic{commentCounter}}$:{#1}}}}

\else
\newcommand{\tim}[1]{}
\newcommand{\ed}[1]{}
\newcommand{\minqi}[1]{}
\newcommand{\jcom}[1]{}
\newcommand{\michael}[1]{}
\newcommand{\jakob}[1]{}

\fi

\usepackage{amsmath}
\usepackage{amsthm}
\usepackage{stmaryrd}

        {\medskip}

\usepackage{amsthm}

\newtheorem{innercustomgeneric}{\customgenericname}
\providecommand{\customgenericname}{}
\newcommand{\newcustomtheorem}[2]{%
  \newenvironment{#1}[1]
  {%
   \renewcommand\customgenericname{#2}%
   \renewcommand\theinnercustomgeneric{##1}%
   \innercustomgeneric
  }
  {\endinnercustomgeneric}
}

\newcustomtheorem{customthm}{Theorem}

        {\hspace*{\fill}$\Box$\par\vspace{4mm}}
        {\hspace*{\fill}$\Box$\par}
        
\DeclareMathOperator*{\argmax}{arg\!\max}
\DeclareMathOperator*{\argmin}{arg\!\min}
\newcommand{\EO}{\mathop{\mathbb{E}}}

\newcommand{\Specialize}[2]{{#1}^{#2}}

\newcommand{\PPOMDP}{\mathcal{M}}

\newcommand{\As}{A}
\newcommand{\Os}{O}
\newcommand{\Ss}[1]{\Specialize{S}{#1}}
\newcommand{\Tf}[1]{\Specialize{\mathcal{T}}{#1}}
\newcommand{\Of}[1]{\Specialize{\mathcal{I}}{#1}}
\newcommand{\Rf}[1]{\Specialize{\mathcal{R}}{#1}}
\newcommand{\discount}{\gamma}

\newcommand{\Ns}{\Theta}
\newcommand{\apply}[2]{#1_{#2}}

\usepackage[accepted]{icml2022}

\usepackage{comment}

\icmltitlerunning{Evolving Curricula with Regret-Based Environment Design}

\begin{document}
\twocolumn[

\icmltitle{Evolving Curricula with Regret-Based Environment Design}

\icmlsetsymbol{equal}{*}

\begin{icmlauthorlist}
\icmlauthor{Jack Parker-Holder}{equal,meta,ox}
\icmlauthor{Minqi Jiang}{equal,meta,ucl}
\icmlauthor{Michael Dennis}{ucb}
\icmlauthor{Mikayel Samvelyan}{meta,ucl}
\icmlauthor{Jakob Foerster}{ox}%
\icmlauthor{Edward Grefenstette}{meta,ucl}
\icmlauthor{Tim Rocktäschel}{meta,ucl}
\end{icmlauthorlist}

\icmlaffiliation{meta}{Meta AI}
\icmlaffiliation{ox}{University of Oxford}
\icmlaffiliation{ucl}{UCL}
\icmlaffiliation{ucb}{UC Berkeley}

\icmlcorrespondingauthor{Jack Parker-Holder}{jackph@robots.ox.ac.uk}
\icmlcorrespondingauthor{Minqi Jiang}{msj@fb.com}

\vskip 0.3in
]

\printAffiliationsAndNotice{\icmlEqualContribution}

\begin{abstract}
Training generally-capable agents with reinforcement learning (RL) remains a significant challenge. A promising avenue for improving the robustness of RL agents is through the use of curricula. One such class of methods frames environment design as a game between a student and a teacher, using regret-based objectives to produce environment instantiations (or levels) at the frontier of the student agent's capabilities. These methods benefit from theoretical robustness guarantees at equilibrium, yet they often struggle to find effective levels in challenging design spaces in practice. By contrast, evolutionary approaches incrementally alter environment complexity, resulting in potentially open-ended learning, but often rely on domain-specific heuristics and vast amounts of computational resources. This work proposes harnessing the power of evolution in a principled, regret-based curriculum. Our approach, which we call \emph{\methodlong{}} (\method{}), seeks to constantly produce levels at the frontier of an agent's capabilities, resulting in curricula that start simple but become increasingly complex. \method{} maintains the theoretical benefits of prior regret-based methods, while providing significant empirical gains in a diverse set of environments. An interactive version of this paper is available at \url{https://accelagent.github.io}.
\end{abstract}

\section{Introduction}
\label{sec:intro}

Reinforcement Learning (RL, \citet{Sutton1998}) considers the problem of an agent learning through experience to maximize reward in a given environment. The past decade has seen a surge of interest in RL, with high profile successes in games \citep{alphastar, dota, alphago, dqn2013, hu2019simplified} and robotics \citep{rubics_cube, dexterity}, with some believing RL may be sufficient for producing generally capable agents \citep{SILVER2021103535}. Despite the promise of RL, it is often a challenge to train agents capable of systematic generalization \citep{kirk2021generalisation}. 

This work focuses on the use of adaptive curricula for training more generally-capable agents. By adapting the training distribution over the parameters of an environment, adaptive curricula have been shown to produce more robust policies in fewer training steps \citep{portelas2019teacher, plr}. For example, these parameters may correspond to friction coefficients in a robotics simulator or maze layouts for a navigation task. Each concrete setting of parameters results in an environment instance called a \emph{level}. Indeed in many prominent RL successes, adaptive curricula have played a key role, acting over opponents \citep{alphastar}, game levels \citep{xland}, or parameters of a simulator \citep{rubics_cube, dexterity}. 

\emph{Unsupervised Environment Design} (UED, \citet{paired}) formalizes the problem of finding adaptive curricula, whereby a teacher agent designs levels using feedback from a student, which seeks to solve them. When using \emph{regret} as feedback, \citet{paired} showed that if the system reaches equilibrium, then the student must follow a minimax regret strategy, i.e. the student would be capable of solving all solvable environments. This approach produces student policies exhibiting impressive zero-shot transfer to challenging human designed environments \citep{paired, jiang2021robustplr, gur2021adversarial}. However, training  such an adversarial teacher remains a challenge, and so far, the strongest empirical results come from \emph{curating} randomly sampled levels for high-regret, rather than learning to directly design such levels \citep{jiang2021robustplr}. This approach is unable to take advantage of any previously discovered level structures, and its performance can be expected to degrade as the size of the design space grows. 

\begin{figure*}[t!]
    \centering
    \begin{minipage}{0.99\textwidth}
    \centering\subfigure{\includegraphics[width=.99\linewidth]{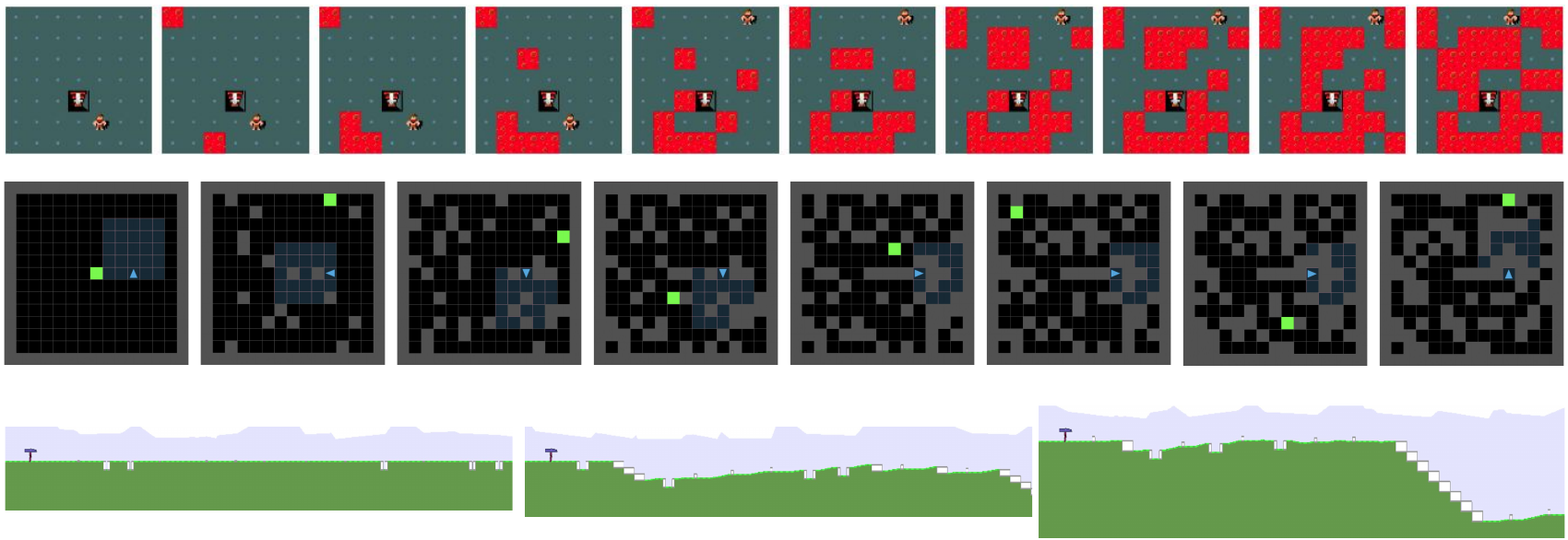}} 
    \vspace{-3mm}
    \caption{The evolution of a level in three different environments: MiniHack lava grids, MiniGrid mazes and BipedalWalker terrains. In each case, the far left shows a base level, acting as a parent for subsequent edited levels to the right. Each level along the evolutionary path has a high regret for the student agent at that point in time. Thus the level difficulty co-evolves with the agent's capabilities. In each environment, we see that despite starting with simple levels, the pursuit of high regret leads to increasingly complex challenges. This complexity emerges entirely without relying on any environment-specific exploration heuristics. Note that since the agent can move diagonally in the lava environment, the final level in the top row is solvable.
    }
    \label{figure:lava7_levels}
    \end{minipage}
    \vspace{-5mm}
\end{figure*}

An important benefit of adaptive curricula is the possibility of open-ended learning \citep{chromaria}, given the curriculum can be steered toward constantly designing novel tasks for the agent to solve. While generating truly open-ended learning remains a grand challenge \citep{stanley2017open}, recent works in the evolutionary community have taken the first steps in this direction, through methods such as Minimal Criteria Coevolution (MCC, \citet{mcc_og}) and POET \citep{poet, enhanced_poet}. These approaches show that evolving levels can effectively produce agents capable of solving a diverse range of challenging tasks. In contrast to prior UED works,  these evolutionary methods directly take advantage of the most useful structures found so far in a constant process of mutation and selection. However, the key drawbacks of these methods are their reliance on domain specific heuristics and need for vast computational resources, making it challenging for the community to make further progress in this direction.
    
\begin{figure}[t!]
    \centering
    \centering\subfigure{\includegraphics[width=.99\linewidth]{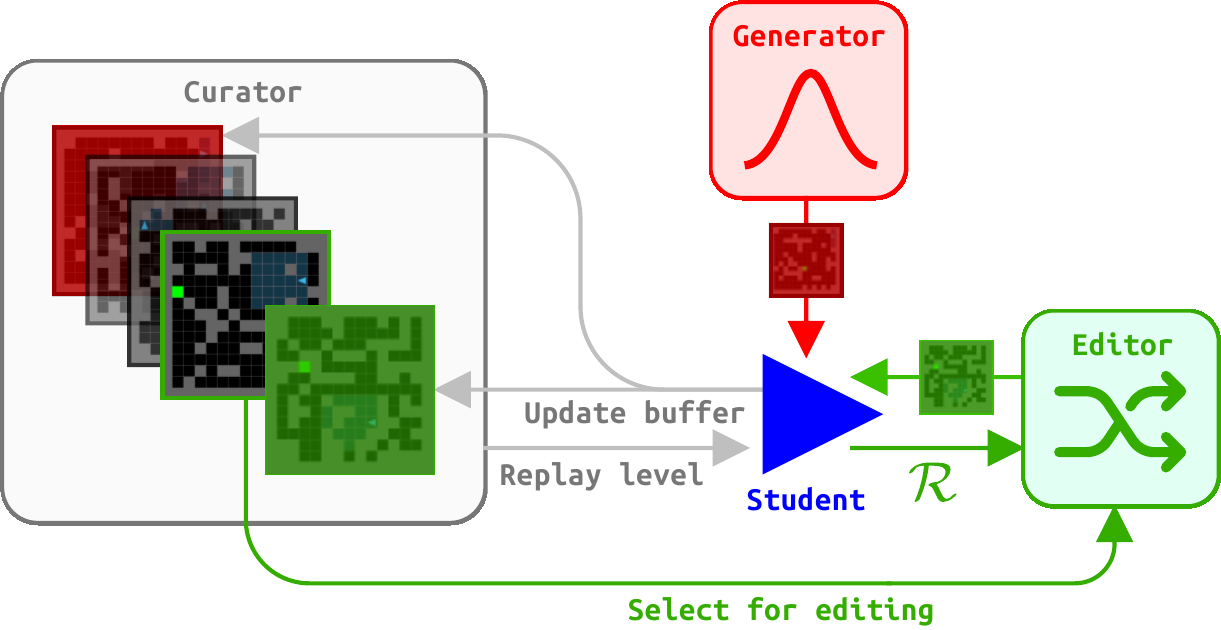}}
    \vspace{-3mm}
    \caption{An overview of \method{}. Levels are randomly sampled from a generator and evaluated, with high-regret levels added to the level replay buffer. The curator selects levels to replay, and the student only trains on replay levels. After training, the regret of replayed levels are edited and evaluated again for level replay. }
    \label{figure:accel}
    \vspace{-5mm}
\end{figure}

In this work, we seek to harness the power and potential open-endedness of evolution in a principled regret-based curriculum. We introduce a new algorithm, called \emph{\methodlong{}}, or \method{}. \method{} evolves a curriculum by making small \emph{edits} (e.g. mutations) to previously high-regret levels, thus constantly producing new levels at the frontier of the student agent's capabilities (see Figure~\ref{figure:accel}). Levels generated by ACCEL begin simple but quickly become more complex. This dynamic benefits the beginning of training where the student can then learn more quickly \citep{Berthouze04adaptive,powerplay2013}, and encourages the policy to rapidly co-evolve with the environment to solve increasingly complex levels (see Figure~\ref{figure:lava7_levels}). 

We believe \method{} provides the best of both worlds: an evolutionary approach that can generate increasingly complex environments, combined with a regret-based curator that reduces the need for domain-specific heuristics and provides theoretical robustness guarantees in equilibrium. \method{} leads to strong empirical gains in both sparse-reward navigation tasks and a 2D bipedal locomotion task over challenging terrain. In both domains, \method{} demonstrates the ability to rapidly increase level complexity while producing highly capable agents. \method{} produces and solves highly challenging levels with a fraction of the compute of previous approaches, reaching comparable level complexity as POET while training on less than 0.05\% of the total number of environment interaction samples, on a single GPU. An open source implementation of \method{} reproducing our experiments is available at \mbox{\url{https://github.com/facebookresearch/dcd}}.

\section{Background}

\subsection{From MDPs to Underspecified POMDPs}

A Markov Decision Process ($\mathrm{MDP}$) is defined as a tuple $\langle \Ss{}, \As, \Tf{}, \Rf{}, \discount \rangle$, where $\Ss{}$ and $\As$ are the set of states and actions respectively, $\Tf{}: S \times A \rightarrow S$ is the transition function from state $s_t$ to state $s_{t+1}$ given action $a_t$, $\Rf{}: S \rightarrow \mathbb{R}$ is the reward function, and $\discount$ is the discount factor. Given an MDP, the goal of reinforcement learning (RL, \citet{Sutton1998}) is to learn a policy $\pi$ that maximizes the expected discounted return, i.e. $\EO[\sum_{t}\gamma^t r_t]$. 

Despite its generality, the MDP framework is often an unrealistic model for real-world environments. First, it assumes full observability of the state, which is often impossible in practice. This limitation is addressed by the \emph{partially observable} MDP (POMDP), which includes an observation function $\Of{}: S \rightarrow O$ mapping the true state (unknown to the agent) to a (potentially noisy) set of observations $\Os$. Secondly, the traditional MDP framework assumes a single reward and transition function, which are fixed throughout training. Instead, in the real world, agents may experience variations not seen during training, making robust transfer crucial in practice.

To address the latter issue, we use the recently introduced \emph{Underspecified} POMDP, or UPOMDP \citep{paired}, given by $\PPOMDP = \langle \As,\Os, \Ns, S, \mathcal{T},\mathcal{I},\mathcal{R},\discount \rangle$. This definition is identical to a POMDP with the addition of $\Theta$ to represent the free parameters of the environment, similar to the context in a Contextual MDP \citep{modi2017markov}. These parameters can be distinct at every time step and incorporated into the transition function $\mathcal{T}: S \times A \times \Ns \rightarrow S$. Following \citet{jiang2021robustplr} we define a \emph{level} $\apply{\PPOMDP}{\theta}$ as an environment resulting from a fixed $\theta \in \Theta$. We define the value of $\pi$ in $\apply{\PPOMDP}{\theta}$ to be
$V^{\theta}(\pi) = \EO[\sum_{i=0}^{T} r_t\gamma^t]$ where $r_t$ are the rewards achieved by $\pi$ in $\apply{\PPOMDP}{\theta}$. UPOMDPs are generally applicable, as $\Theta$ can represent possible transition dynamics and changes in observations, e.g. in sim2real \citep{peng2017dr, rubics_cube, dexterity}, as well as different reward functions or world topologies in procedurally-generated environments.

\subsection{Methods for Unsupervised Environment Design}

Unsupervised Environment Design (UED, \citet{paired}) seeks to produce a series of levels that form a curriculum for a \emph{student} agent, such that the student agent is capable of systematic generalization across all possible levels. UED typically views levels as produced by a generator (or \emph{teacher}) maximizing some utility function, $U_t(\pi, \theta)$. For example DR corresponds to a teacher with a  constant utility function, for any constant $C$:
\begin{align}
\label{eqn:dr_obj}
    U_t^U(\pi, \theta) = C.
\end{align}
Recent UED methods use a teacher that maximizes \emph{regret}, defined as the difference between the expected return of the current policy and the optimal policy. The teacher's utility is then defined as:
  \begin{align}
  \label{eqn:regret}
  U_t^R(\pi, \theta) & =\argmax_{\pi^* \in \Pi}\{\textsc{Regret}^{\theta}(\pi,\pi^*)\} \\
  & = \argmax_{\pi^* \in \Pi}\{V^\theta(\pi^*)-V^\theta(\pi)\}.
  \end{align}
Regret-based objectives are desirable, as they have been shown to promote the simplest possible levels that the student cannot currently solve \citep{paired}. More formally, if $S_t= \Pi$ is the strategy set of the student and $S_t = \Theta$ is the strategy set of the teacher, then if the learning process reaches a Nash equilibrium, the resulting student policy $\pi$ provably converges to a minimax regret policy, defined as
\begin{equation}
\pi = \argmin_{\pi \in \Pi}\{\max_{\theta,\pi^* \in \Theta, \Pi}\{\textsc{Regret}^{\theta}(\pi,\pi^*)\}\}.
\end{equation}
However, without access to $\pi^*$ for each level, UED algorithms must approximate the regret. PAIRED estimates regret as the difference in return attained by the main student agent and a second agent. By maximizing this difference, the teacher maximizes an approximation of the student's regret. Furthermore, multi-agent learning systems may not always converge in practice \citep{MazumdarRS20}. Indeed, the Achilles' heel of prior UED methods, like PAIRED \citep{paired}, is the difficulty of training the teacher, typically entailing an RL problem with sparse rewards and long-horizon credit assignment. An alternative regret-based UED approach is \emph{Prioritized Level Replay} (PLR, \citet{plr, jiang2021robustplr}). PLR trains the student on challenging levels found by curating a rolling buffer of the highest-regret levels surfaced through random search over possible level configurations. In practice, PLR has been found to outperform other UED methods that directly train a teacher. PLR approximates regret using a score function such as the \emph{positive value loss}:
\begin{equation}
\label{eqn:pvl}
\frac{1}{T}\sum_{t=0}^{T} \max \left(\sum_{k=t}^T(\gamma\lambda)^{k-t}\delta_k, 0\right)
\end{equation}
where $\lambda$ and $\gamma$ are the Generalized Advantage Estimation (GAE, \citet{schulman2016gae}) and MDP discount factors respectively, and $\delta_t$, the  TD-error at timestep $t$. Equipped with this method for approximating regret, Corollary 1 in \citet{jiang2021robustplr} finds that if the student agent only trains on curated levels, then it will follow a minimax regret strategy at equilibrium. 
Thus, counterintuitively, the student learns more effectively by training on less data.

Empirically PLR has been shown to produce policies with strong generalization capabilities, but remains limited in only curating randomly sampled levels. PLR's inability to directly extend previously discovered structures makes it unlikely to sample more complex structures to encourage further robustness and generalization. Random search suffers from the curse-of-dimensionality in higher-dimensional design spaces, where randomly encountering levels at the frontier of the agent's current capabilities can be highly unlikely, especially as the agent becomes more capable.

\section{Adversarially Compounding Complexity}

In this section we introduce a new algorithm for UED, combining an evolutionary environment generator with a principled regret-based curator. Unlike PLR which relies on random sampling to produce new batches of training levels, we instead propose to make \emph{edits} (e.g. mutations) to previously curated ones. Evolutionary methods have been effective in a variety of challenging optimization problems \citep{neuronature, qdnature}, yet typically rely on handcrafted, domain-specific rules. For example, POET manually filters BipedalWalker levels to have a return in the range $[50,300]$. The key insight in this work is that regret serves as a domain-agnostic fitness function for evolution, making it possible to consistently produce batches of levels at the frontier of agent capabilities across domains. Indeed, by iteratively editing and curating the resulting levels, the levels in the level replay buffer quickly increase in complexity. As such, we call our method \emph{\methodlong{}}, or \method{}.

\method{} does not rely on a specific editing mechanism, which could be any mutation process used in other open-ended evolutionary approaches \citep{chromaria}. In this paper, editing involves making a handful of changes (e.g. adding or removing obstacles in a maze), which can operate directly on environment elements within the level or on a more indirect encoding such as the latent-space representation of the level under a generative model of the environment. 

In general, editing may rely on more advanced mechanisms, such as search-based methods, but in this work we predominantly make use of simple, random mutations.  \method{} makes the key assumption that regret varies smoothly with the environment parameters $\Theta$, such that the regret of a level is close to the regret of others within a small edit distance. If this is the case, then small edits to a single high-regret level should lead to the discovery of entire batches of high-regret levels---which could be an otherwise challenging task in high-dimensional design spaces. 

Following PLR \citep{jiang2021robustplr}, we do not immediately train on edited levels. Instead, we first evaluate them and only add them to the level replay buffer if they have high regret, estimated by positive value loss (Equation \ref{eqn:pvl}). The full procedure is shown in Algorithm~\ref{alg:accel}.

\begin{figure}[h]
\vspace{-3mm}
\begin{minipage}{\linewidth}
\begin{algorithm}[H]

\SetAlgoLined
\caption{\accmethod{}}
\label{alg:accel}
\textbf{Input:} Level buffer size $K$, initial fill ratio $\rho$, level generator \\
\textbf{Initialize:} Initialize policy $\pi(\phi)$, level buffer $\bm{\Lambda}$ \\
    Sample $K * \rho$ initial levels to populate $\bm{\Lambda}$ \\
    \While{not converged}{
        Sample replay decision $d \sim P_{D}(d)$ \\
        \eIf{$d=0$}{
            Sample level $\theta$ from level generator\\
            Collect $\pi$'s trajectory $\tau$ on $\theta$, with stop-gradient $\phi_{\bot}$ \\  
            Compute regret score $S$ for $\theta$ (Equation \ref{eqn:pvl}) \\
            Update $\bm{\Lambda}$ with $\theta$ if score $S$ meets threshold
        }
        {
        Sample a replay level, $\theta \sim \bm{\Lambda}$ \\
        Collect policy trajectory $\tau$ on $\theta$ \\
        Update $\pi$ with rewards $\bm{R}(\tau)$ \\
        Edit $\theta$ to produce $\theta'$ \\
        Collect $\pi$'s trajectory $\tau$ on $\theta'$, with stop-gradient $\phi_{\bot}$ \\  
        Compute regret score $S$ ($S'$) for $\theta$ ($\theta'$) \\       
        Update $\bm{\Lambda}$ with $\theta$ ($\theta'$) if score $S$ ($S'$) meets threshold \\
        (Optionally) Update Editor using score $S$
        }
    }
\end{algorithm}
\end{minipage}
\vspace{-3mm}
\end{figure}

\method{} can be seen as a UED algorithm taking a step toward open-ended evolution \citep{stanley2017open}, where the evolutionary fitness is estimated regret, as levels only stay in the population (that is, the level replay buffer) if they meet the high-regret criterion for curation. However, \method{} avoids some important weaknesses of evolutionary algorithms such as POET: First, \method{} maintains a population of levels, but not a population of agents. Thus, \method{} requires only a single desktop GPU for training. In contrast, evolutionary approaches typically require a CPU cluster. Moreoever, forgoing an agent population allows \method{} to avoid the agent selection problem. Instead, \method{} directly trains a single \emph{generalist} agent. Finally, since \method{} uses a minimax \emph{regret}  objective (rather than minimax as in POET), it naturally promotes levels at the frontier of agent's capabilities, without relying on domain-specific knowledge (such as reward ranges). Training on high regret levels also means that \method{} inherits the robustness guarantees in equilbrium from PLR (Corollary 1 in \citet{jiang2021robustplr}):

\newtheorem{accelremark}{Remark}
\begin{accelremark}
If \method{} reaches a Nash equilibrium, then the student follows a minimax regret strategy.
\end{accelremark}

In contrast, other evolutionary approaches primarily justify their applicability solely via empirical results on specific domains. As our experiments show, a key strength of \method{} is its generality. It can produce highly capable agents in a diverse range of environments, without domain knowledge.

\section{Experiments}
\label{sec:experiments}
In our experiments we seek to compare agents trained with \method{} with several of the  best-performing UED baselines. In all cases, we train a student agent via Proximal Policy Optimization (PPO, \citet{schulman2017proximal}). To evaluate the quality of the resulting curricula, we show all performance with respect to the number of gradient updates for the student policy, as opposed to total number of environment interactions, which is, in any case, often comparable for PLR and \method{} (see Table~\ref{table:stepcount}). For a full list of hyperparameters for each experiment please see Table~\ref{table:hyperparams} in Section \ref{sec:hparams}. Our primary baseline is Robust PLR \citep{jiang2021robustplr}, which combines the random generator with a regret-based curation mechanism. The other baselines are domain randomization (DR), PAIRED \citep{paired}, and a minimax adversarial teacher. The minimax baseline corresponds to the objective used in POET without the hand-coded constraints. We leave the comparison to population-based methods to future work due to the computational expense required. We report results in a consistent manner across environments: In each case, we show the emergent complexity during training and report test performance in terms of the aggregate inter-quartile mean (IQM) and optimality gap using the recently introduced \texttt{rliable} library \citep{agarwal2021deep}.

We begin with a partially-observable navigation environment, where we test our agents' transfer capabilities on human-designed levels. Finally, we compare each method on the continuous-control environment from \citet{poet}, featuring a highly challenging distribution of training levels that requires the agent to master multiple behaviors to achieve strong performance. We also include a proof-of-concept experiment in the Appendix (see  Section~\ref{app:minihack}).

\subsection{Partially Observable Navigation}

\begin{figure}[t!]
    \begin{minipage}{0.5\textwidth}
    \centering\subfigure{\includegraphics[width=.99\linewidth]{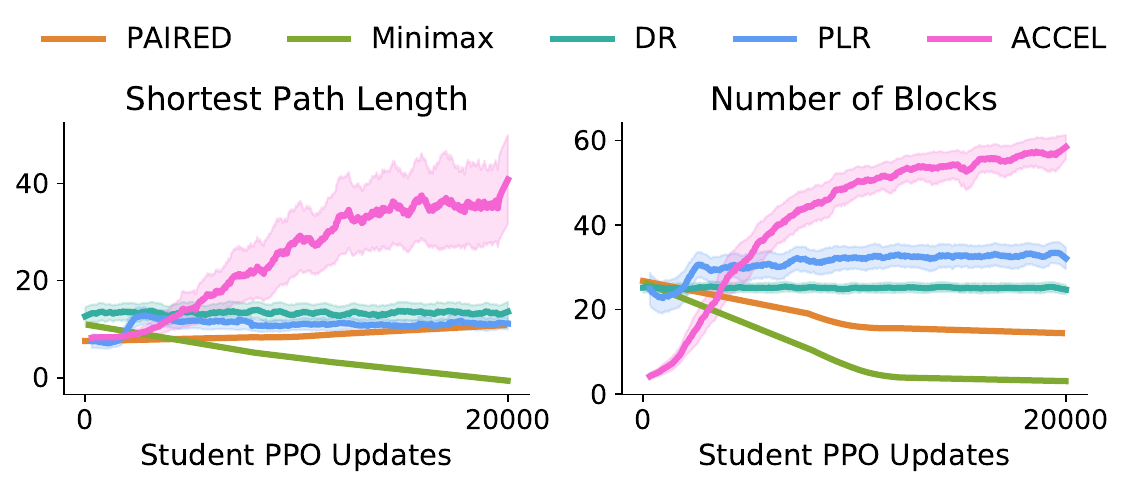}}
    \vspace{-3mm}
    \caption{Emergent complexity metrics for mazes generated during training. Mean and standard error across 5 training seeds are shown.}
    \label{figure:mg_training_meta}
    \end{minipage}
\end{figure}

We begin with a maze navigation  environment based on MiniGrid \citep{gym_minigrid}, originally introduced in \citet{paired}. Despite being a conceptually simple environment, training robust agents in this domain requires a large-scale experiment: Our agents train for 20k updates ($\approx$350M steps, see Table~\ref{table:stepcount}), learning an LSTM-based policy with a 147-dimensional partially-observable observation. Our DR baseline samples between 0 and 60 blocks to place, providing a sufficient range for PLR to form a curriculum. For \method{} we begin with empty rooms and randomly edit the block locations (by adding or removing blocks), as well as the goal location. In Figure~\ref{figure:mg_training_meta}, we report training performance and complexity metrics. We see that \method{} rapidly compounds complexity, leading to training levels with significantly higher block counts and longer solution paths than other methods.

\begin{figure}[t!]
    \centering
    \begin{minipage}{0.5\textwidth}
    \centering\subfigure{\includegraphics[width=.99\linewidth]{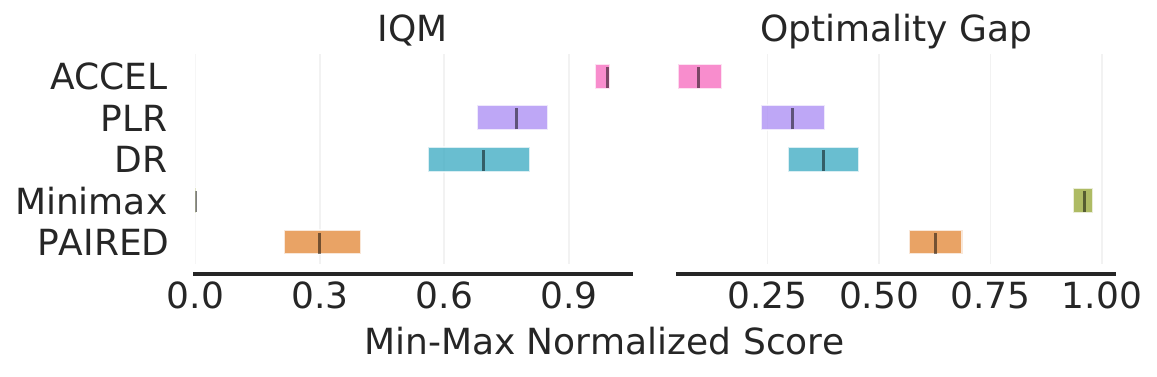}}
    \vspace{-3mm}
    \caption{Aggregate zero-shot test performance in the maze domain.}
    \vspace{-5mm}
    \label{figure:minigrid_rliable}
    \end{minipage}
\end{figure}

We evaluate the zero-shot transfer performance of each method on a series of held-out test environments, as done in prior works. For DR, PLR, and \method{}, evaluation occurs after 20k student PPO updates, focusing the  comparison on the effect of the curriculum. The minimax and PAIRED results are those reported in \citet{jiang2021robustplr} at 250M training steps ($\approx$30k updates). As we see, \method{} performs at least as well as the next best method in almost all test environments, with particularly strong performance in Labyrinth and Maze. As reported in Figure~\ref{figure:minigrid_rliable}, \method{} achieves drastically stronger performance than all other methods in aggregate across all test environments: Its IQM approaches a perfect solved rate compared to below 80\% for the next best method, PLR, with an 80.2\% probability of improvement over PLR. Detailed, per environment test results are provided in Figures~\ref{figure:minigrid_zs_levels} and \ref{figure:mg_zs_results} in Appendix~\ref{appendix:full_results}. Figure~\ref{figure:minigridlevels} shows example levels generated by each method. We see \method{} produces more structured mazes than the baselines.

\begin{figure}[t!]
    \centering
    \begin{minipage}{0.5\textwidth}
    \centering\subfigure{\includegraphics[width=.99\linewidth]{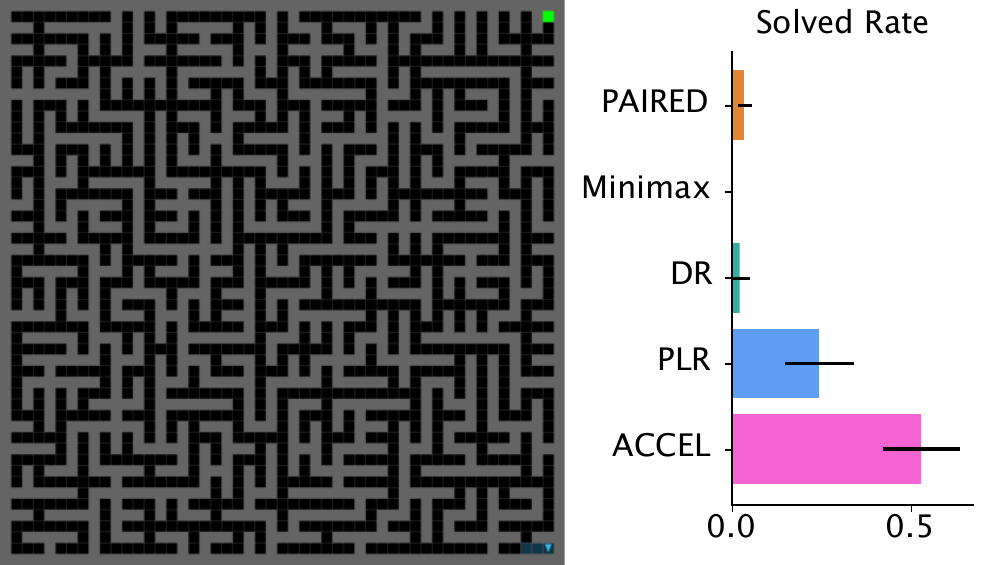}}
    \vspace{-3mm}
   \caption{Zero-shot performance on a large procedurally-generated maze environment. The bars show mean and standard error over 5 training seeds, each evaluated over 100 episodes. \method{} achieves over twice the success rate of the next best method.}
    \label{figure:accelmazes}
    \end{minipage}
    \vspace{-2mm}
\end{figure}

Next, we consider an even more challenging setting based on a larger version of PerfectMaze, a procedurally-generated maze environment, shown in Figure~\ref{figure:accelmazes}, where levels have $51\times51$ tiles with a maximum episode length of over 5k steps---an order of magnitude larger than training levels. 
We evaluate agents for 100 episodes (per training seed), using the same checkpoints in  Figure~\ref{figure:minigrid_rliable}. The results in Figure~\ref{figure:accelmazes} show \method{} significantly outperforms all baselines with a success rate of 53\% compared to the next best method, PLR, which has a success rate of 25\%, while all other methods fail. Notably, successful agents approximately follow the left-hand rule for solving single-component mazes.

\begin{figure}[t!]
    \centering
    \begin{minipage}{0.5\textwidth}
    \centering\subfigure{\includegraphics[width=.8\linewidth]{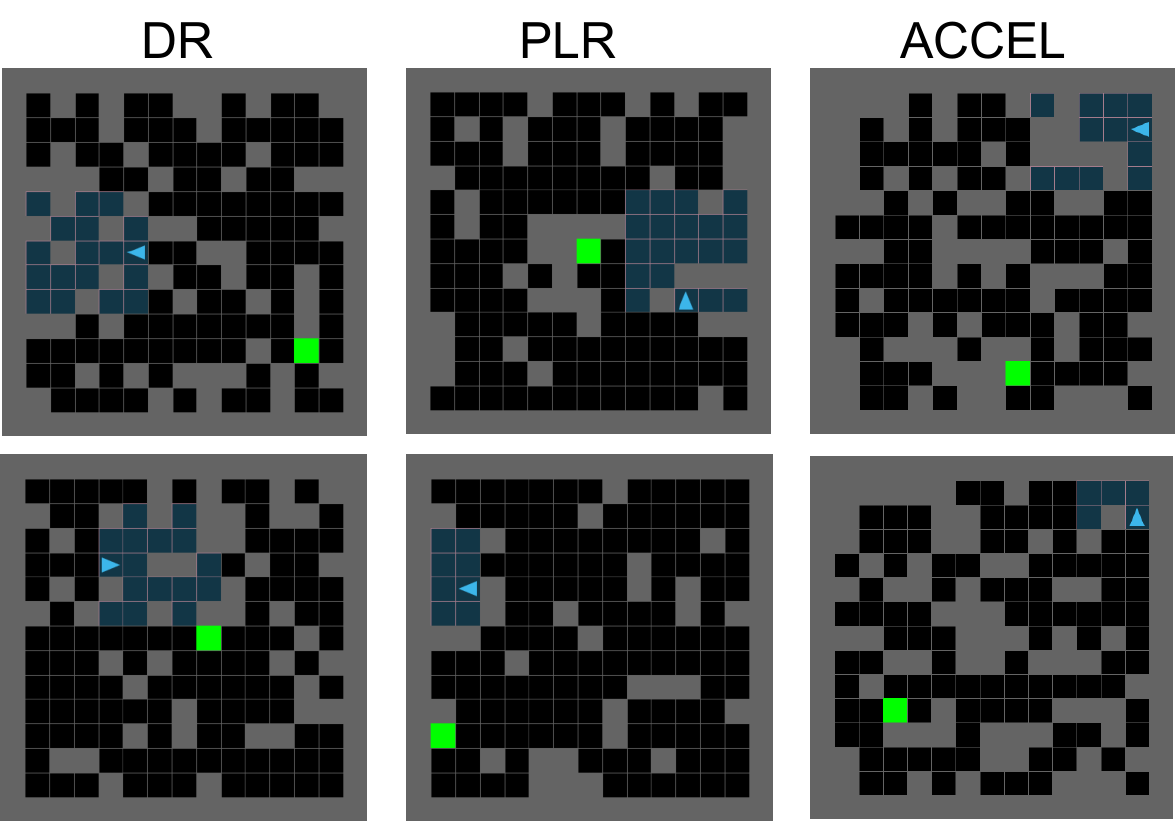}}
    \vspace{-3mm}
    \caption{Example levels generated by DR, PLR, and \method{}.}
    \label{figure:minigridlevels}
    \end{minipage}
    \vspace{-2mm}
\end{figure}

\begin{figure}[t!]
    \centering
    \begin{minipage}{0.5\textwidth}
    \centering\subfigure{\includegraphics[width=.8\linewidth]{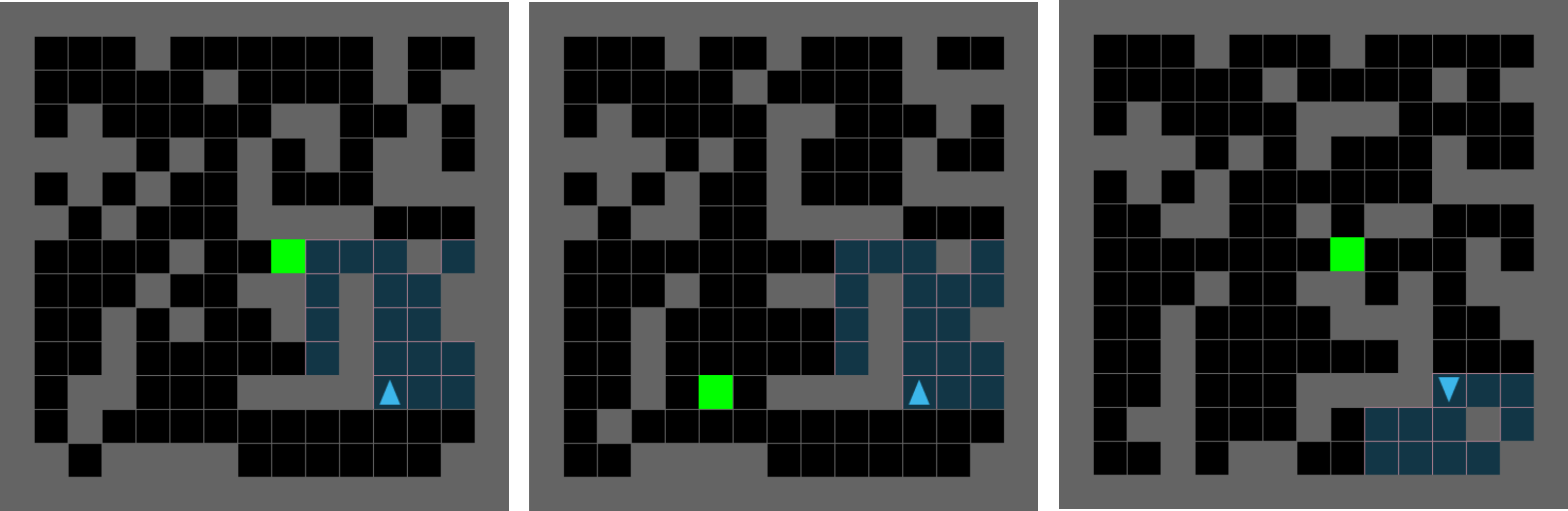}}
    \vspace{-3mm}
    \caption{Despite sharing a common ancestor, each of these levels requires different behaviors to solve. Left: The agent can approach the goal by moving upwards or leftwards. Middle: The goal is on the left. Right: The left path is blocked.}
    \label{figure:threemazes}
    \end{minipage}
\end{figure}

We seek to understand the key drivers of \method{}'s outperformance: Incremental changes to a level can lead to a diverse batch of new ones \citep{sturtevant2020unexpected}, which may move those that are currently too hard or too easy towards the frontier of the agent's capabilities. This diversity may prevent overfitting. For example, in Figure~\ref{figure:threemazes}, we see three edits of the same level produced by \method{}. Each has a similar initial observation, yet requires the agent to explore in different directions to reach the goal, thereby pressuring the agent to actively explore the environment. Further, making edits that do not change the optimal solution path can be seen as a form of data augmentation that changes the observation but not the optimal policy. Data augmentation has been shown to improve sample efficiency and robustness in RL \citep{rad, drq, ucb_drac}.

\subsection{Walking in Challenging Terrain}
\begin{figure}[H]
    \vspace{-2mm}
    \centering
    \begin{minipage}{0.5\textwidth}
    \centering\subfigure{\includegraphics[width=.95\linewidth]{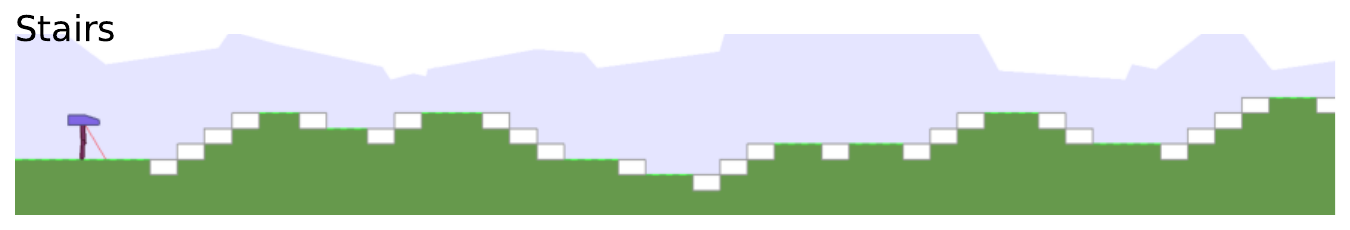}}
    \centering\subfigure{\includegraphics[width=.95\linewidth]{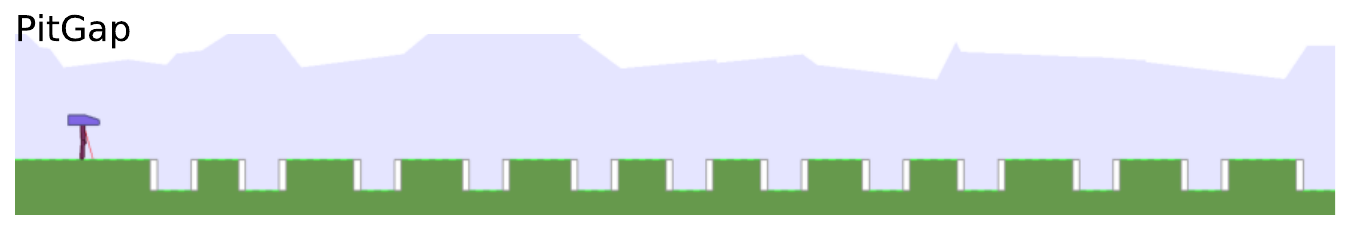}}
    \centering\subfigure{\includegraphics[width=.95\linewidth]{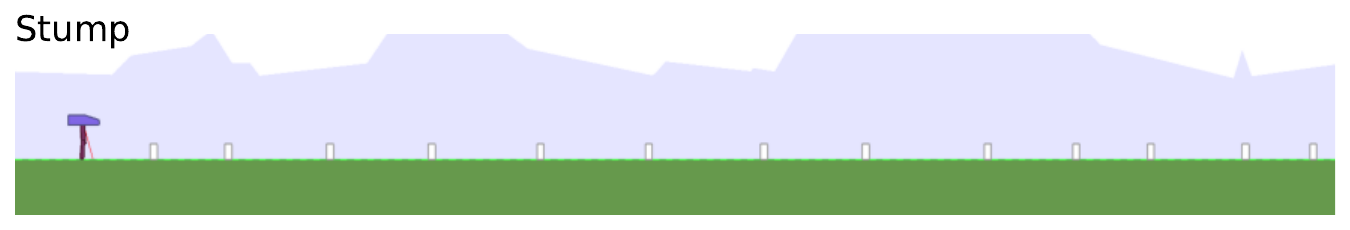}}
    \centering\subfigure{\includegraphics[width=.95\linewidth]{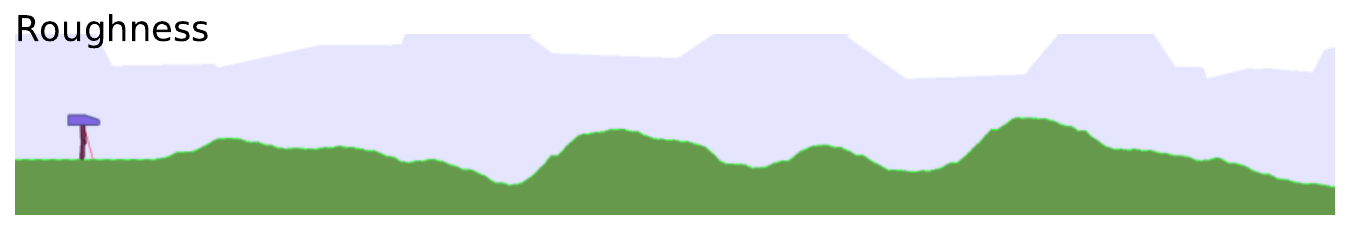}}
    \centering\subfigure{\includegraphics[width=.99\linewidth]{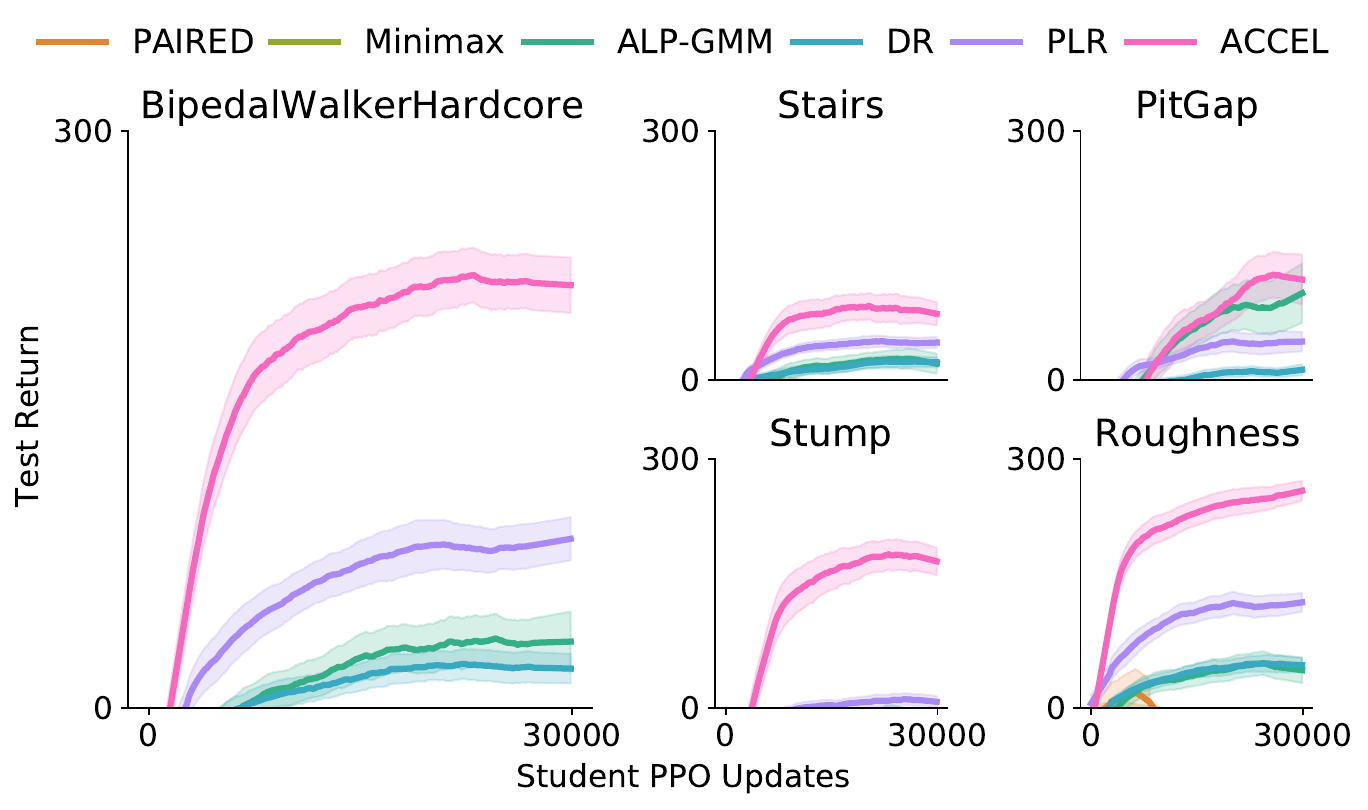}}
    \vspace{-3mm}
    \caption{Top: Examples of the four individual challenges in BipedalWalker. Bottom: Performance on test environments during training (mean and standard error). Negative returns are omitted.}
    \vspace{-3mm}
    \label{figure:bipedal_trainperf}
    \end{minipage}
\end{figure}

Finally, we evaluate \method{} in the \texttt{BipedalWalker} environment from \citet{poet}, a continuous-control environment with dense rewards. As in \citet{poet}, we use a modified version of \texttt{BipedalWalkerHardcore} from OpenAI Gym \citep{Gym}. We include all eight parameters in the design space, rather than only the subset used in \citet{poet}. This environment is detailed at length in Appendix \ref{sec:env_details}. We run all baselines from previous experiments, in addition to ALP-GMM \citep{portelas2019teacher}, which was originally tested on BipedalWalker. We train agents for 30k student updates, equivalent to between 1B to 2B total environment steps, depending on the method (see Table \ref{table:stepcount}). During training we evaluate agents on both the simple \texttt{BipedalWalker} and more challenging \texttt{BipedalWalker-Hardcore} environments, in addition to four environments testing the agent's effectiveness against specific, isolated challenges otherwise present to varying degrees in training levels: \{\texttt{Stump}, \texttt{PitGap}, \texttt{Stairs}, \texttt{Roughness}\}, shown in Figure \ref{figure:bipedal_trainperf}.

\begin{figure}[h!]
    \centering
    \vspace{-0mm}
    \begin{minipage}{0.5\textwidth}
    \centering\subfigure{\includegraphics[width=.99\linewidth]{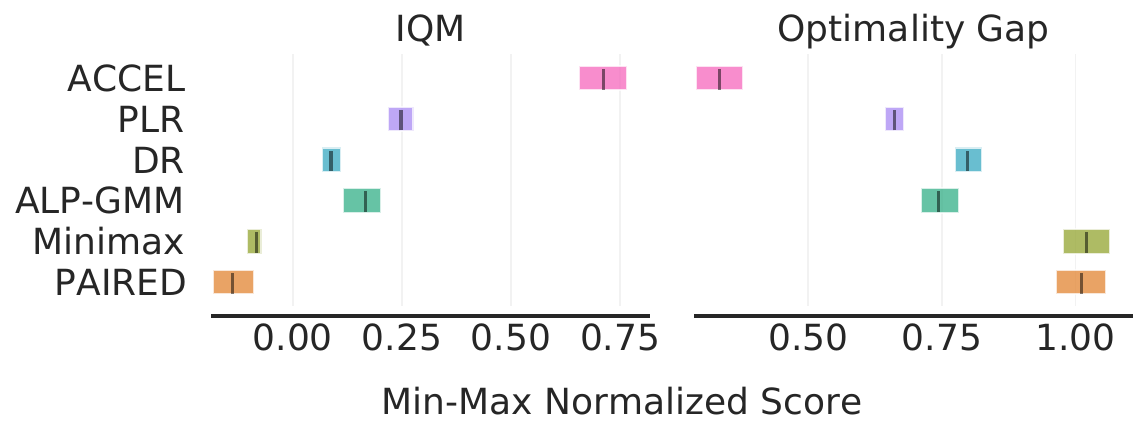}}
    \vspace{-4mm}
    \caption{Aggregate performance for ten seeds across all five BipedalWalker test environments.}
    \vspace{-3mm}
    \label{figure:bipedal_rliable}
    \end{minipage}
\end{figure}

After 30k PPO updates, we conduct a more rigorous evaluation based on 100 test episodes in each test environment. Figure \ref{figure:bipedal_rliable} reports the aggregate results, normalized according to a return range of $[0,300]$. \method{} significantly outperforms all baselines, achieving close to 75\% of optimal performance, almost three times the performance of the best baseline, PLR. All other baselines struggle, likely due to the environment design space containing a high proportion of levels not useful for learning. Faced with such challenging levels, agents may learn to resort to the locally optimal behavior of preventing itself from falling (avoiding a -100 penalty), rather than attempt forward locomotion. Finally, we see ALP-GMM performs poorly when the design space is increased from 2D (as in \citet{portelas2019teacher}) to 8D.

\begin{figure}[h!]
    \centering
    \begin{minipage}{0.5\textwidth}
    \centering\subfigure{\includegraphics[width=.99\linewidth]{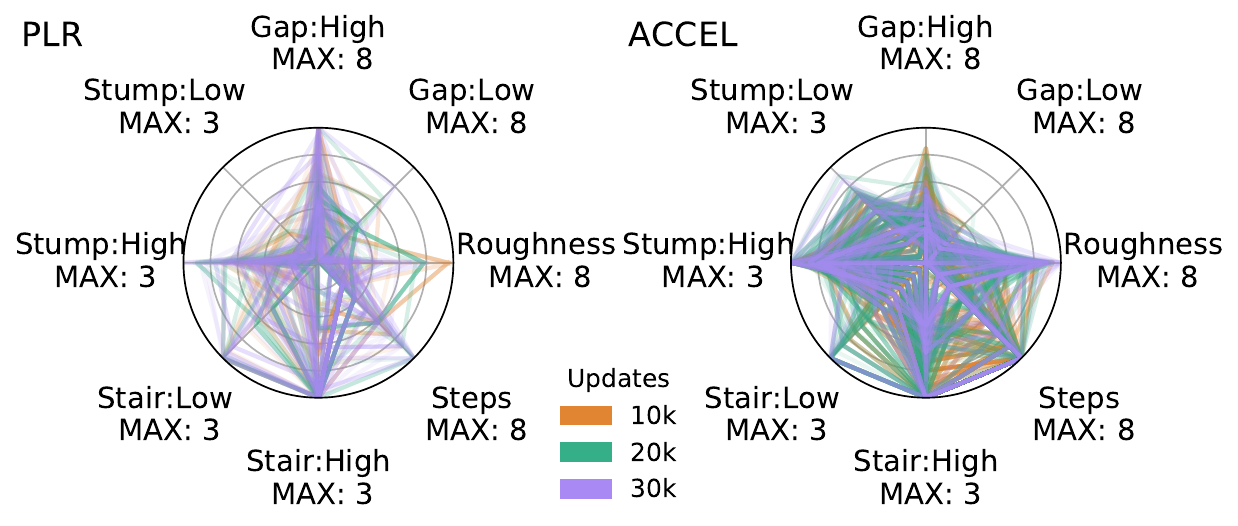}}
    \centering\subfigure{\includegraphics[width=.99\linewidth]{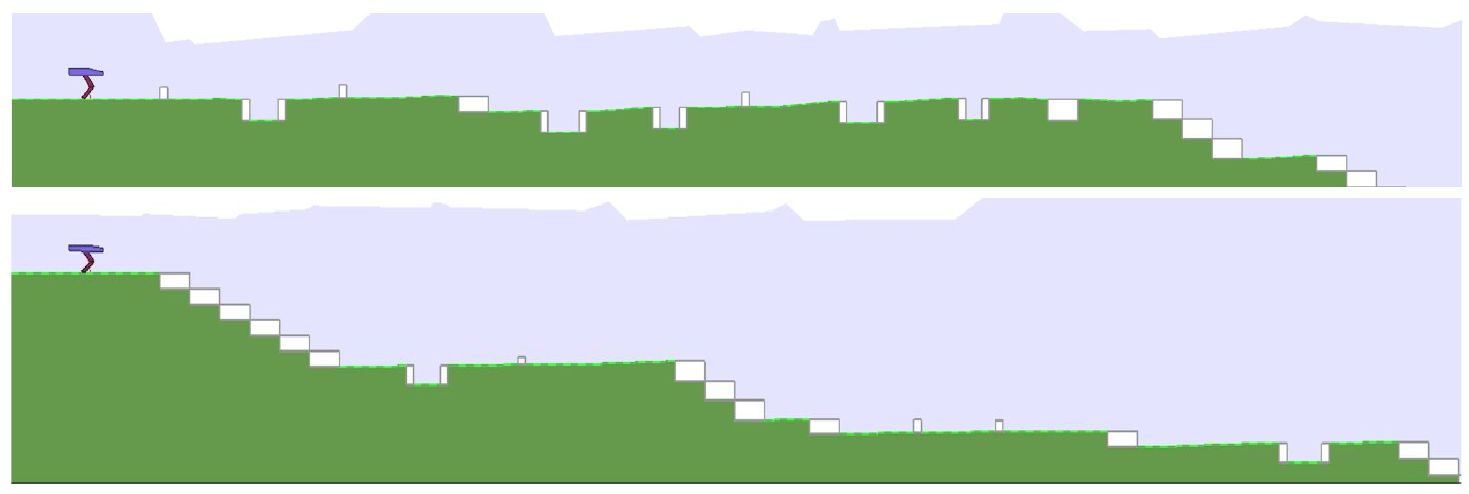}}
    \caption{Top: Rose plots of complexity metrics of BipedalWalker levels discovered by PLR and ACCEL. Each line represents a solved level from the associated checkpoint. All levels are among the top-100 highest regret levels for the given checkpoint. Bottom: Two levels created and solved by \method{}.}
    \label{figure:bipedal_complexity}
    \end{minipage}
\end{figure}

Next we seek to understand the properties of the evolving distribution of high-regret levels. We include all solved levels from the top-100 regret levels after 10k, 20k, and 30k student updates. For each level we show all eight parameters in Figure~\ref{figure:bipedal_complexity} (top), with the PLR agent as a comparison. As we see, \method{} solves a large quantity of difficult levels of comparable difficulty with other methods such as POET, but uses a fraction of the compute. For comparison, \method{} sees 2.07B environment steps at 30k student updates, less than 0.5\% of that used in \citet{poet}. 

\subsection{POET Comparison}

For a more direct comparison with POET, we train each method using 10 training seeds for 50k student PPO updates with the smaller 5D BipedalWalker environment encoding used in \citet{poet}. We use the thresholds provided in \citet{poet}, summarized in Table~\ref{table:poet_thresholds}, to evaluate the difficulty of generated levels. A level meeting none of these thresholds is considered \emph{easy}, while meeting one, two or three is considered \emph{challenging}, \emph{very challenging} or \emph{extremely challenging} respectively. 

\begin{table}[h!]
\vspace{-5mm}
\begin{center}
\caption{\small{Environment encoding thresholds for 5D BipedalWalker.
}}
\label{table:poet_thresholds}
\scalebox{0.85}{
\begin{tabular}{ ccc } 
\toprule
\textbf{Stump Height (High)} & \textbf{Pit Gap (High)}  & \textbf{Ground Roughness} \\ 
\midrule 
$\geq2.4$ & $\geq6$ & $\geq4.5$ \\
\bottomrule
\end{tabular}}
\end{center}
\vspace{-2mm}
\end{table}

\begin{figure}[h!]
    \centering
    \begin{minipage}{0.5\textwidth}
    \centering\subfigure{\includegraphics[width=.9\linewidth]{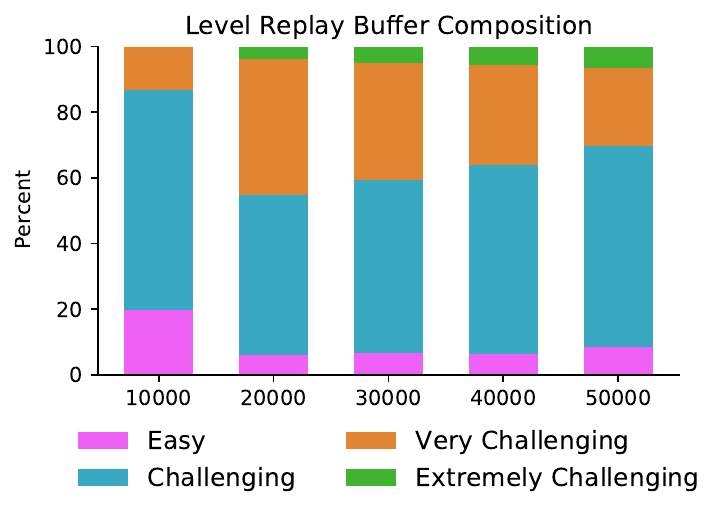}}
    \vspace{-5mm}
    \caption{\small{Percent of \method{} level replay buffer for each difficulty. The complexity emerges purely in pursuit of high-regret levels.}}
    \vspace{-1mm}
    \label{figure:bipedal_poet_progression}
    \end{minipage}
\end{figure}

In Figure~\ref{figure:bipedal_poet_progression}, we show the composition of the \method{} level replay buffer during training. As we see, \method{} produces an increasing number of extremely challenging levels as training progresses. This is a significant achievement given that POET's evolutionary curriculum is unable to create levels in this category, without including a complex stepping-stone procedure \citep{poet}. We thus see minimax-regret UED offers a computationally cheaper alternative to producing such levels. Moreover, POET explicitly encourages the environment parameters to reach high values through a novelty bonus, whereas the complexity discovered by \method{} is completely emergent, arising purely through the pursuit of high-regret levels.

While POET seeks to discover a diverse population of specialists, each capable of solving a a specific extremely challenging level, \method{} aims to train a single generalist. To evaluate the generality of the \method{} agent, we test all agents trained in the 5D BipedalWalker environment on the settings outlined in Figure~\ref{figure:bipedal_trainperf}, and report the results in Table~\ref{table:poet_exp_robustness}. Note that the \texttt{Stairs} environment is now out-of-distribution, as the agent never sees stairs during training. As we saw in the higher-dimensional setting, the \method{} agent is able to perform well across all settings.

\begin{table}[h!]
\vspace{-4mm}
\begin{center}
\caption{Test solved rates at 50k updates (mean and standard error) for 10 runs of each method on 100 episodes. Extremely challenging evaluation uses 1000 episodes due to the high diversity of levels. Bold values are within one standard error of the best mean.
}
\label{table:poet_exp_robustness}
\scalebox{0.9}{
\begin{tabular}{ l ccc } 
\toprule
 & PLR & ALP-GMM & ACCEL \\ 
\midrule 
\texttt{Stump} & $0.04 \pm 0.02$ & $0.07 \pm 0.02$ & $\mathbf{0.44 \pm 0.08}$ \\
\texttt{PitGap} & $0.2 \pm 0.09$ & $\mathbf{0.58 \pm 0.08}$ & $\mathbf{0.61 \pm 0.08}$ \\
\texttt{Roughness} & $0.23 \pm 0.04$ & $0.13 \pm 0.03$ & $\mathbf{0.73 \pm 0.03}$ \\
\texttt{Stairs} & $0.02 \pm 0.0$ & $0.01 \pm 0.0$ & $\mathbf{0.12 \pm 0.02}$ \\
\midrule
\texttt{Hardcore} & $0.21 \pm 0.04$ & $0.2 \pm 0.04$ & $\mathbf{0.65 \pm 0.02}$ \\
\texttt{Extreme} &  $0.01 \pm 0.01$ & $0.02 \pm 0.01$ & $\mathbf{0.12 \pm 0.02}$ \\
\bottomrule
\end{tabular}}
\end{center}
\vspace{-5mm}
\end{table}

We further test all methods on a held-out distribution of extremely challenging levels. In this case, we resample the level parameters for each episode so to ensure they meet all three criteria in Table~\ref{table:poet_thresholds}. This leads to a highly diverse set of test levels. To mitigate stochasticity influencing the outcome, we evaluate each method over 1000 episodes. The results are summarized in Table~\ref{table:poet_exp_robustness}, where we see \method{} attains 12\% average solved rate, while PLR and ALP-GMM see 1\% and 2\% average solved rates respectively.

Finally, we seek to evaluate our agents on specific levels produced by POET. We used the rose plots from \citet{poet} to create six extremely challenging environments, each solved by one of the three POET runs. Unsurprisingly our agents find these levels challenging and see low success rates. We note that this is not a perfect comparison---POET fixes the level generator's random seed, thereby solving a single level for each parameterization, while we repeatedly sample different levels under the same parameterization. Still, 9 out of 10 of our independent runs solved at least one of the 6 environments at least once out of 100 trials. See the Appendix (Table~\ref{table:poet_roseplot_test}) for more detail on this experiment.

In summary, we believe \method{} can produce levels of comparable complexity to POET, without a novelty bonus or domain-specific heuristics, at the fraction of the compute cost. Moreover, \method{} produces a single agent robust across environment challenges, while POET results in multiple agents, each tailored to individual challenges. Therefore, we believe our method produces agents that are more robust, and thus more generally capable. 

\begin{figure}[t!]
    \centering
    \begin{minipage}{0.5\textwidth}
    \centering\subfigure{\includegraphics[width=.99\linewidth]{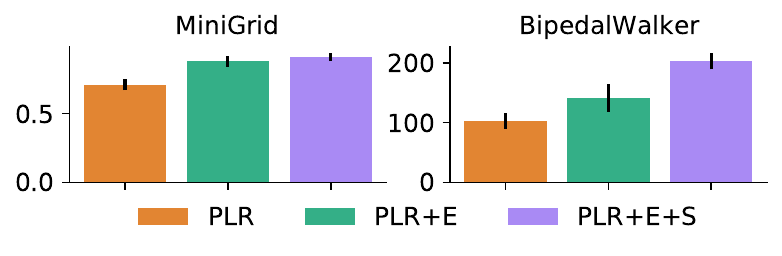}}
    \vspace{-3mm}
    \caption{Aggregate returns for Editing ablations in MiniGrid and BipedalWalker. E=editing, S=start simple.}
    \label{figure:accel_ablation_editing}
    \end{minipage}
    \vspace{-5mm}
\end{figure}

\textbf{Do we need to start simple?} We conduct a simple ablation study on \method{} to test the importance of the editing mechanism and the inductive bias of starting simple. In Figure~\ref{figure:accel_ablation_editing} we show the performance of three approaches: PLR (sample and replay DR levels), PLR+E (sample, replay, and edit DR levels) and finally PLR+E+S (i.e. \method{}). As we see, editing levels leads to improved performance, while starting simple is more important in BipedalWalker environments.

\subsection{Discussion and Limitations}

Our experiments demonstrate that \method{} is capable of forming highly effective curricula in three challenging and diverse environments. 
In MiniGrid, we showed it is possible to produce complex mazes that facilitate zero-shot transfer to out-of-distribution, human-designed ones, including those an order of magnitude larger than the training environment. Finally, in the BipedalWalker setting we produced comparable level complexity as POET using a single agent and a small fraction of the compute cost. Our experiments provide evidence that \method{} can be an effective curriculum method in more open-ended UED design spaces. 

Importantly, the goal of our work differs from \citet{poet}. The primary motivation of \method{} is to produce a single robust agent that can solve a wide range of challenges. \method{}'s regret-based curriculum seeks to prioritize the simplest levels that the agent cannot currently solve. In contrast, POET co-evolves agent-environment pairs in order to find specialized policies for solving a single highly specialized task. POET's specialized agents may likely learn to solve challenging environments outside the capabilities of ACCEL's generalist agents, but at the cost of potentially overfitting to their paired levels. Thus, unlike ACCEL, the policies produced by POET should not be expected to be robust across the full distribution of levels. The relative merits of POET and ACCEL thus highlight an important trade-off between specialization and generalization, both of which may ultimately be important for solving  more complex, larger scale problems.

While \method{}'s simplicity is appealing, larger design spaces may require additional mechanisms, like actively promoting diversity in level design. Moreover, \method{} uses an inductive bias by starting with the simplest base case (e.g. an empty room), which may not always be possible in practice. In some settings, simple levels for agents may be more complex in design, e.g. in a Hide-and-Seek game.

\section{Related Work}

\begin{table*}[t!]
\begin{center}
\caption{The components of related approaches. Like POET, ACCEL evolves levels, but only trains a single agent while using a minimax-regret objective to ensure levels are solvable. PAIRED uses minimax regret to train the generator, and does not replay levels. Finally, PLR curates levels using minimax regret, but relies solely on domain randomization for generation.}
\label{table:summary_of_diff}
\scalebox{0.95}{
\begin{tabular}{ ccccc } 
\toprule
\textbf{Algorithm}   & \textbf{Generation Strategy} & \textbf{Generator Obj} & \textbf{Curation Obj} &  \textbf{Output} \\ \midrule
POET \citep{poet}  & Evolution & Minimax  & MCC & Specialists \\ 
PAIRED \citep{paired}  & Reinforcement Learning & Minimax Regret  & None & Generalist \\ 
PLR \citep{plr, jiang2021robustplr}  & Random & None  & Minimax Regret & Generalist \\ 
\midrule
\method{}  & Random + Evolution  & None & Minimax Regret & Generalist \\ 
\bottomrule
\end{tabular}}
\vspace{-4mm}
\end{center}
\end{table*}

In this section we provide a brief overview of related work. We provide a more detailed discussion in Appendix~\ref{app:relatedwork}.

The goal of this work is to produce agents that are capable of systematic generalization across a wide range of environments \citep{whiteson2009generalized}, which has recently been a focus for the deep RL community \citep{packer2019assessing,igl2019generalization, procgen, agarwal2021contrastive, zhang2018generalizationgrid, ghosh2021generalization}. A common approach for producing robust agents is Domain Randomization (DR, \citet{evolutionary_dr, cad2rl}), widely used in robotics \citep{tobin_dr, james2017transferring, dexterity, rubics_cube}.

The evolutionary component of ACCEL is inspired by the open-ended creative potential of POET \citep{poet, enhanced_poet, mcc_og, pinsky}, which seeks to train a population of highly capable specialists. By contrast, ACCEL trains a single generally capable agent with a regret-based curriculum as in PAIRED \citep{paired} and Robust PLR \citep{jiang2021robustplr} (see Table~\ref{table:summary_of_diff}). These methods for \emph{Unsupervised Environment Design} \citep[UED,][]{paired}, naturally relate to the field of \emph{Automatic Curriculum Learning} \citep[ACL,][]{portelas2020automatic, florensa2017, riac2009}, with the key difference being that in UED \emph{all} elements of the POMDP are underspecified.

Our work also closely relates to previous environment design literature in the symbolic AI commmunity \citep{zhang2008ed,zhang2009ed, keren2017equi, keren2019efficient}, though our focus falls primarily on generating curricula. Finally, we take inspiration from the field of \emph{procedural content generation} \citep[PCG;][]{pcg, pcg_illuminating}, which seeks to produce a distribution of levels for a given environment, often using machine learning \citep{pcgml, Bhaumik2020TreeSV, DL_PCG}. We are particularly inspired by PCGRL \citep{pcgrl,controllablepcgrl2021earle} which frames level design as an RL problem, making incremental changes to a level to maximize some objective subject to game-specific constraints.

\section{Conclusion and Future Work}

We proposed ACCEL, a new method for unsupervised environment design (UED), that evolves a curriculum by \emph{editing} previously curated levels. Editing induces an evolutionary process that leads to a wide variety of environments at the frontier of the agent's capabilities, producing curricula that start simple and quickly compound in complexity. Thus, ACCEL provides a principled regret-based curriculum that exploits an evolutionary process to produce a broad spectrum of environment complexity matched to the agent's current capabilities. Importantly, ACCEL avoids the need for domain-specific heuristics. In our experiments, we showed that \method{} is capable of training robust agents in a series of challenging design spaces, where ACCEL agents outperform the best-performing baselines.

We are excited by the many possibilities for extending how ACCEL edits and maintains its population of high-regret levels.  The editing mechanism could encompass a wide variety of approaches, such as Neural Cellular Automata \citep{earle2021illuminating}, controllable editors  \citep{controllablepcgrl2021earle}, or perturbing the latent space of a generative model \citep{lsi_overcooked}. Another possibility is to actively seek levels which are likely to produce useful levels in the future \citep{evolvabilityes}. Moreover, ACCEL's evolutionary search may be expedited by introducing so-called \emph{extinction events} \citep{raup1986biological, lehman2015extinction}, believed to play a crucial role in natural evolution.  We did not explore methods to encourage level diversity, but such diversity is likely important for larger design spaces, such as 3D control tasks that transfer more directly to the real world. It remains an open question whether producing sufficient diversity would require a population, for example using the domain-agnostic, population-based novelty objective in Enhanced POET \citep{enhanced_poet}. We believe such ideas at the intersection of evolution and adaptive curricula can take us closer to producing a truly open-ended learning process between the agent and the environment \citep{stanley2017open}. Finally, we note that while \method{} may be an effective approach for automatically generating an effective curriculum, it may still be necessary to likewise adapt the agent configuration  \citep{autorl_survey} to most effectively train agents in open-ended environments.

\section*{Acknowledgements}
We thank Kenneth Stanley, Alessandro Lazaric, Ian Fox, and Iryna Korshunova for insightful discussions, as well as the anonymous reviewers for their useful feedback. This work was funded by Meta AI.




\bibliography{refs}
\bibliographystyle{apalike}

\newpage
\appendix
\onecolumn

\section{Extended Related Work}
\label{app:relatedwork}

Our paper focuses on testing agents on distributions of environments, long known to be crucial for evaluating the generality of RL agents \citep{whiteson2009generalized}. The inability of deep RL agents to reliably generalize across distributions of environment configurations has drawn considerable attention \citep{packer2019assessing,igl2019generalization, procgen, agarwal2021contrastive, zhang2018generalizationgrid, ghosh2021generalization}, with policies often failing to adapt to changes in the observation \citep{observational_overfitting}, dynamics \citep{ball2021augwm}, or reward \citep{zhang2018generalizationcont}. In this work, we seek to provide agents with a set of training levels to produce a policy that is robust to such variations. 

In particular, we focus on the \emph{Unsupervised Environment Design} \citep[UED,][]{paired} paradigm, which aims to design environment directly, such that they induce experiences that result in learning more robust policies. Domain Randomization  \citep[DR,][]{evolutionary_dr, cad2rl}, which simply randomizes the environment configuration, can be viewed as the most basic form of UED. DR has been particularly successful in areas such as robotics \citep{tobin_dr, james2017transferring, dexterity, rubics_cube}, with extensions that actively update the DR distribution \citep{adr2020, adr2_2020}. This paper directly extends \emph{Prioritized Level Replay} \citep[PLR,][]{plr, jiang2021robustplr}), a method for curating DR levels such that only those with high learning potential are revisited by the student agent for training. PLR is related to TSCL \citep{tscl}, self-paced learning \citep{selfpace2019klink, space}, and ALP-GMM \citep{portelas2019teacher}, which all seek to maintain and update distributions over informative environment parameterizations based on the recent performance of the agent. Recently, a method similar to PLR was shown to be capable of producing generally-capable agents in a simulated game world with a smooth space of levels \citep{xland}.

\citet{paired} first formalized UED and introduced the PAIRED algorithm, a minimax-regret \citep{minimax_regret} UED algorithm whereby an environment adversary learns to present levels that maximize regret, approximated as the difference in performance between the main student agent and a second student agent. PAIRED produces agents with improved zero-shot transfer to unseen environments and has been extended to train agents that can robustly navigate websites \citep{gur2021adversarial}. Adversarial objectives have also been considered in robotics \citep{pinto2017advrobotics} and in directly searching for situations in which the agent sees the weakest performance \citep{everett2019worlds}. POET \citep{poet, enhanced_poet} considers co-evolving a population of minimax environments and agents that solve them. ACCEL harnesses the evolutionary potential of POET while training only a single agent, which takes significantly less resources, while also avoiding the agent selection problem. 

UED is related to \emph{Automatic Curriculum Learning} \citep[ACL,][]{portelas2020automatic, florensa2017, riac2009}, which seeks to provide an adaptive curriculum of increasingly challenging tasks or goals \citep{her2017}. This setting differs from a general UPOMDP, which aims to actively generate the entire environment given a domain specification and where the free parameters, e.g. the task or goal specifier, are typically not fully observed. In Asymmetric Self-Play \citep{sukhbaatar2018asp,openai2021asymmetric}, the agent's goal is based on reversing the trajectory of another; this process leads to effective curricula for robotic manipulation tasks. AMIGo \citep{amigo} and APT-Gen \citep{fang2021adaptive} produce adaptive curricula for hard-exploration, goal-conditioned problems. Many ACL methods focus on learning to reach goals or states with high uncertainty \citep{Racaniere2020Automated, skewfit2020, frontier_zhang2020}, including latent states inside a generative model \citep{goalgan, lexa2021}.

Automatic environment design has also been considered in the symbolic AI community as a means to shape an agent's decisions \citep{zhang2008ed,zhang2009ed} Automated environment design \citep{keren2017equi, keren2019efficient} seeks to redesign specific levels to improve the agent's performance within them. In contrast, UED adapts curricula that improves performance across levels.

Our work also relates to the field of \emph{procedural content generation} \citep[PCG;][]{pcg, pcg_illuminating}, which has studied the algorithmic generation of game levels for over a decade \citep{togelius2008evolving}. Popular PCG environments used in RL include the Procgen Benchmark \citep{procgen}, MiniGrid \citep{gym_minigrid}, Obstacle Tower \citep{obstacletower}, GVGAI \citep{perez2019general}, and the NetHack Learning Environment \citep{kuettler2020nethack}. This work uses the MiniHack environment \citep{samvelyan2021minihack}, which provides a flexible framework for creating diverse environments. Many recent PCG methods use machine learning \citep{pcgml, Bhaumik2020TreeSV, DL_PCG}. PCGRL \citep{pcgrl,controllablepcgrl2021earle} frames level design as an RL problem, whereby the design policy incrementally changes the level to maximize some objective subject to game-specific constraints. Unlike \method{}, it makes use of hand-designed dense rewards and focuses on the design of levels for human players, rather than as an avenue to training highly-robust agents.

\newpage
\section{Additional Experimental Results}
\subsection{Learning with Lava}
\label{app:minihack}

Here we explore a simple proof of concept: a grid environment, where the agent must navigate to a goal in the presence of lava blocks. The grid is only $7\times7$, but remains challenging as the episode terminates with zero reward if the agent touches the lava. This dynamic makes exploration more difficult by penalizing random actions. While toy, such challenges may be relevant in real-world, safety-critical settings, where agents may wish to avoid events causing early termination during training. For DR and PLR, the random generator samples the number of lava tiles to place from the range $[0,20]$. For \method{}, we use a generator that produces empty rooms and then proceeds to edit the levels by adding or removing lava tiles. The environment is built with MiniHack \citep{samvelyan2021minihack} and is fully observable with a high-dimensional input. The full environment details are provided in \ref{sec:env_details}. 

\begin{figure}[H]
    \centering
    \begin{minipage}{0.6\textwidth}
    \vspace{-2mm}
    \centering\subfigure{\includegraphics[width=.99\linewidth]{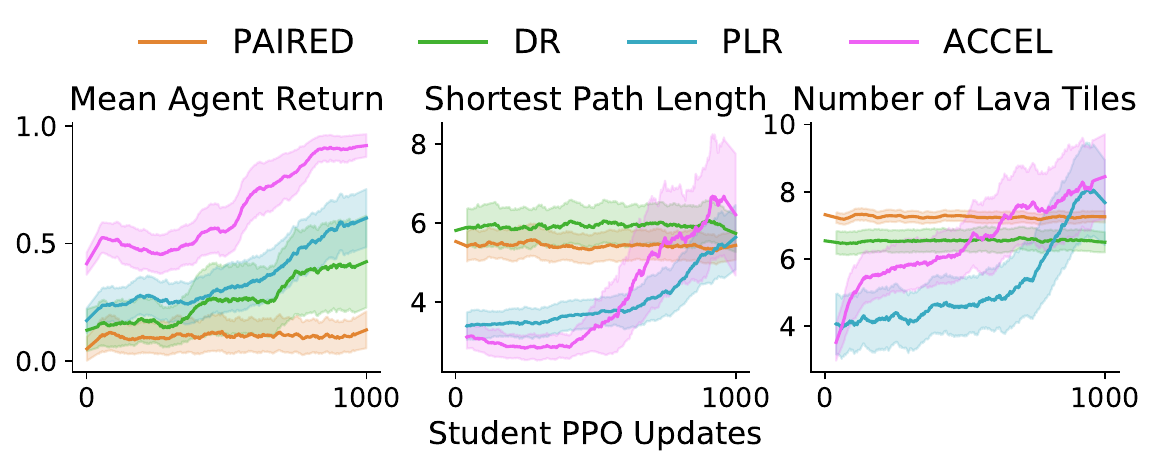}}    
    \vspace{-5mm}
    \caption{Training return and emergent complexity in Lava Grid. The plots report the mean and standard error over 5 seeds. 
    }
    \label{figure:lava7_metadata}
    \vspace{-3mm}
    \end{minipage}
\end{figure}

Figure \ref{figure:lava7_metadata} shows the results of running each method over 5 random seeds. Despite starting with empty rooms, \method{} quickly produces levels with more lava than the other methods, while also attaining higher training returns, reaching near-perfect performance on its training distribution. This behavior is entirely driven by the pursuit of high-regret levels, which constantly seeks the frontier of the agent's capabilities. PLR is able to produce a similar training profile to \method{}, but achieves lower values for each complexity metric.

\begin{figure}[H]
    \centering
    \vspace{-3mm}
    \begin{minipage}{0.5\textwidth}
    \centering\subfigure{\includegraphics[width=.99\linewidth]{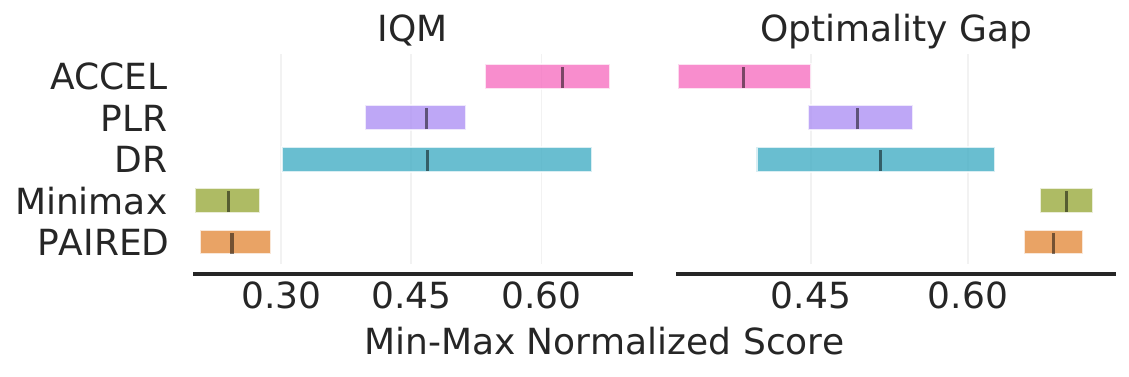}}
    \vspace{-5mm}
    \caption{Lava Grid aggregate test performance.}
    \vspace{-7mm}
    \label{figure:minihack_rliable}
    \end{minipage}
\end{figure}

We evaluate each agent on a series of test levels after 1000 PPO updates (approximately 20M timesteps), and report the aggregate results in Figure \ref{figure:minihack_rliable}, where we see that \method{} is the best performing method. Extended results are shown in Table \ref{table:lava7x7results}. The first three test environments (Empty, 10 Tiles and 20 Tiles) evaluate the performance of the agent within its training distribution, while we also include a held-out human designed environment, LavaCrossing S9N1, ported from MiniGrid \citep{gym_minigrid}. 
As we see, \method{} performs best on all of the in-distribution environments (whose levels can be directly sampled in the training distribution), while also being only one of two approaches to get meaningfully above zero in the human designed task.

\vspace{-2mm}

\begin{table}[h!]
\begin{center}
\caption{Test performance in in-distribution and out-of-distribution environments. Each entry is the mean (and standard error) of 5 independent training runs, where each run is evaluated for 100 trials on each environment. $\dagger$ indicates the generator's per-tile lava distribution is binomial and the generator can place lava in 20 blocks. $\ddagger$ indicates the generator first samples the number of lava tiles to place (in $[0,20]$), then places that many at random locations. Bold values are within one standard error of the best mean.
}
\label{table:lava7x7results}
\scalebox{0.77}{
\begin{tabular}{ l | ccccccc } 
\toprule
\textbf{Test Environment} & PAIRED  & Minimax & DR$\dagger$ & PLR$\dagger$ & DR$\ddagger$ & PLR$\ddagger$ & \method{} \\ 
\midrule 
Empty & $0.77 \pm 0.03$ & $0.76 \pm 0.02$ & $0.81 \pm 0.03$ & $0.97 \pm 0.03$ & $0.89 \pm 0.05$ & $0.96 \pm 0.04$ & $\mathbf{1.0 \pm 0.0}$ \\ 
10 Tiles & $0.12 \pm 0.03$ & $0.05 \pm 0.01$ & $0.12 \pm 0.02$ & $0.35 \pm 0.18$ & $0.33 \pm 0.15$ & $0.3 \pm 0.05$ & $\mathbf{0.49 \pm 0.07}$ \\ 
20 Tiles & $0.06 \pm 0.01$ & $0.11 \pm 0.04$ & $0.06 \pm 0.01$ & $0.15 \pm 0.09$ & $0.23 \pm 0.12$ & $0.25 \pm 0.06$ & $\mathbf{0.35 \pm 0.08}$ \\ 
LavaCrossing S9N1 & $0.0 \pm 0.0$ & $0.0 \pm 0.0$ & $0.0 \pm 0.0$ & $0.01 \pm 0.01$ & $\mathbf{0.05 \pm 0.05}$ & $0.01 \pm 0.0$ & $\mathbf{0.05 \pm 0.04}$ \\ 
\bottomrule
\end{tabular}}
\end{center}
\end{table}

\newpage
\subsection{Level Evolution}

\begin{figure}[h!]
    \vspace{-10mm}
    \centering
    \begin{minipage}{0.99\textwidth}
    \centering\subfigure{\includegraphics[width=.092\linewidth]{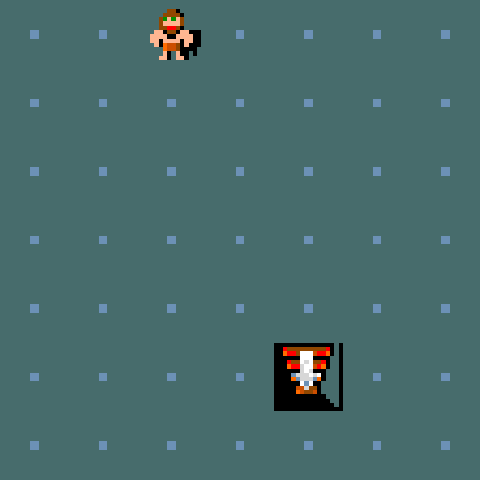}}
    \centering\subfigure{\includegraphics[width=.092\linewidth]{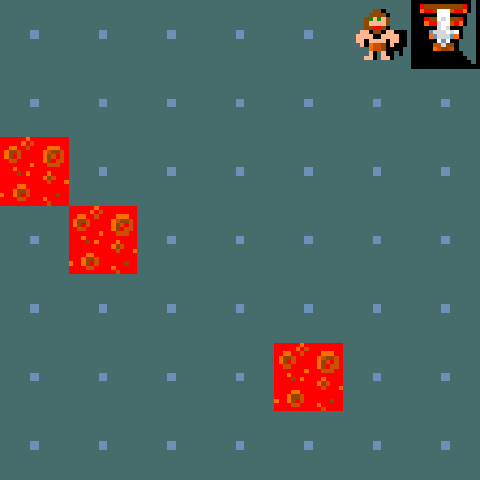}}
    \centering\subfigure{\includegraphics[width=.092\linewidth]{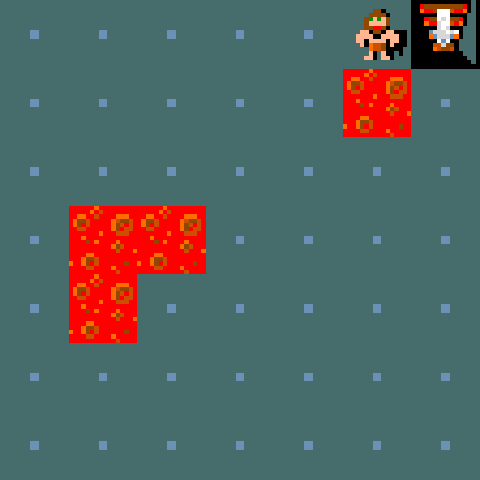}}
    \centering\subfigure{\includegraphics[width=.092\linewidth]{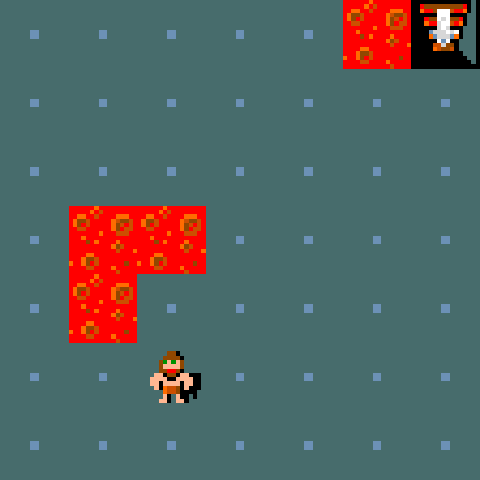}} 
    \centering\subfigure{\includegraphics[width=.092\linewidth]{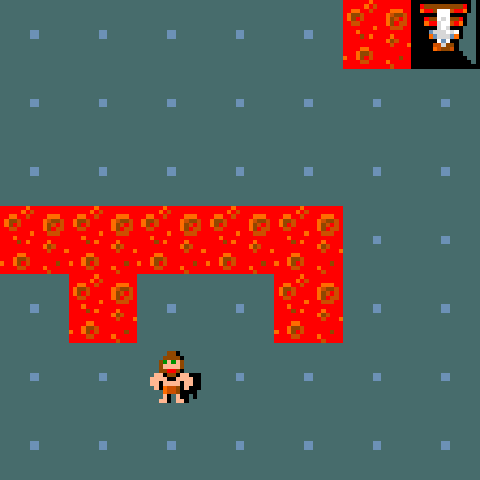}}
    \centering\subfigure{\includegraphics[width=.092\linewidth]{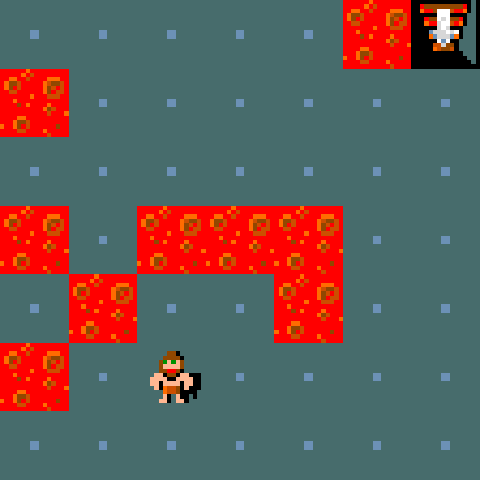}}    
    \centering\subfigure{\includegraphics[width=.092\linewidth]{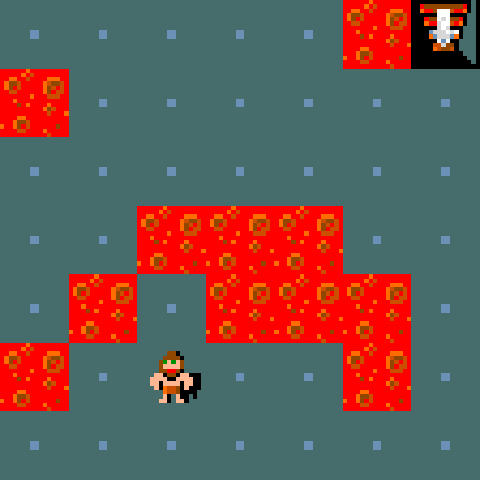}}    
    \centering\subfigure{\includegraphics[width=.092\linewidth]{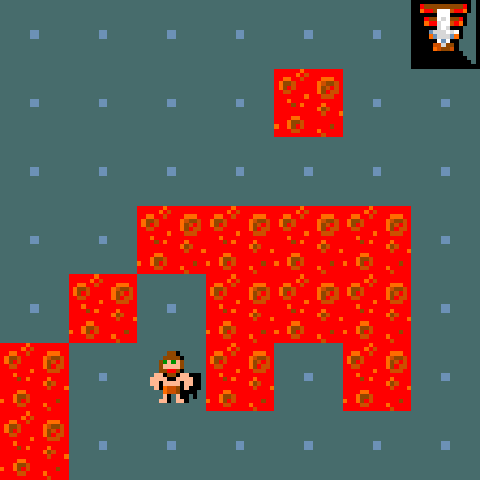}}    
    \centering\subfigure{\includegraphics[width=.092\linewidth]{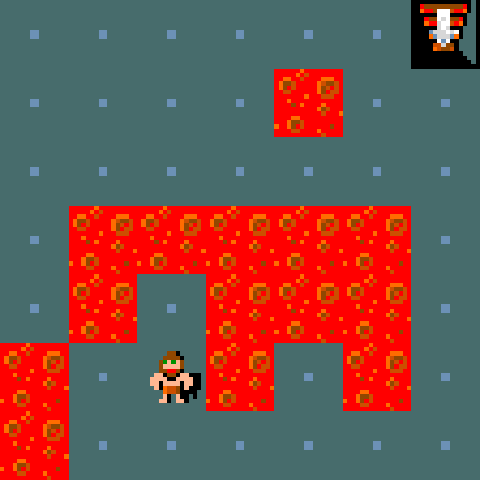}}  
    \centering\subfigure{\includegraphics[width=.092\linewidth]{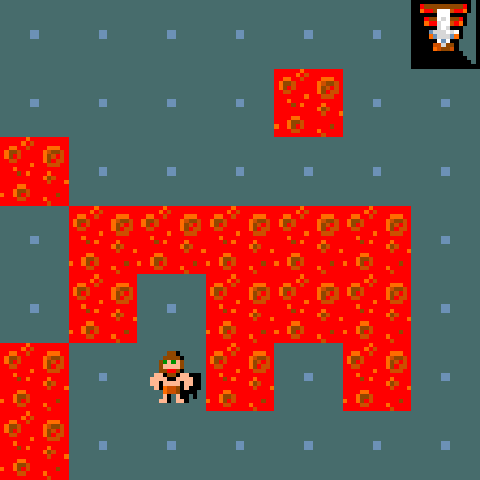}}

    \centering\subfigure{\includegraphics[width=.092\linewidth]{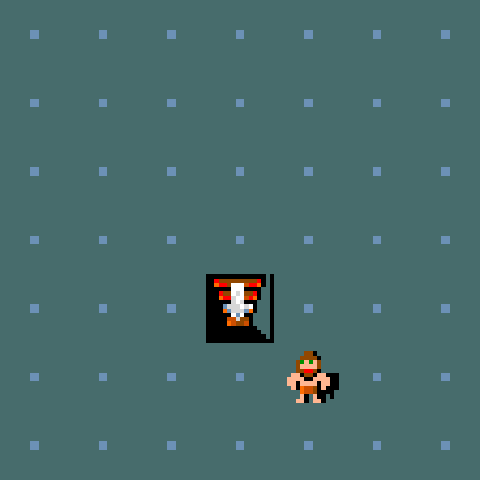}}
    \centering\subfigure{\includegraphics[width=.092\linewidth]{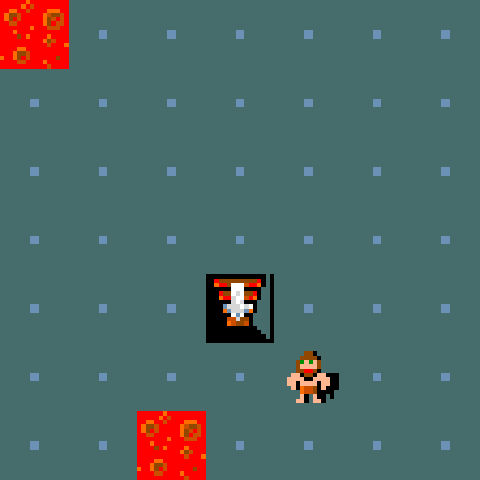}}
    \centering\subfigure{\includegraphics[width=.092\linewidth]{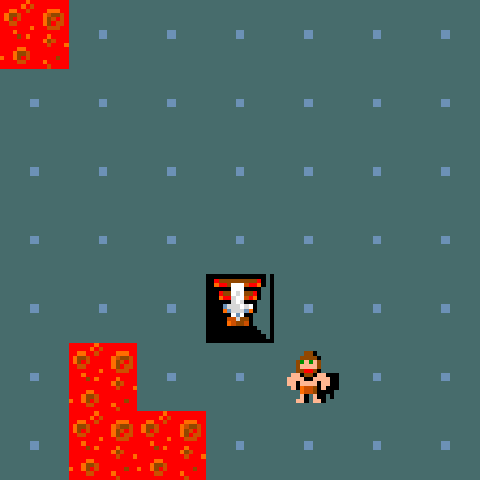}}
    \centering\subfigure{\includegraphics[width=.092\linewidth]{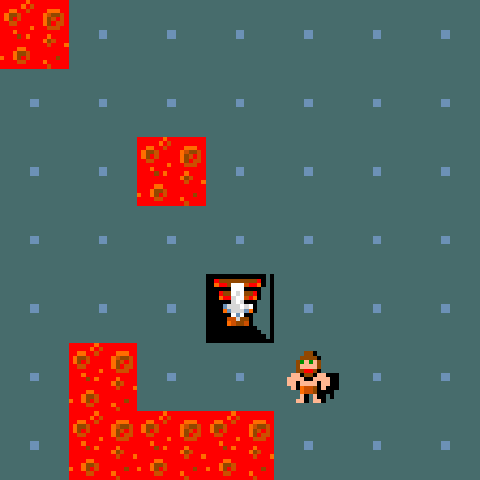}} 
    \centering\subfigure{\includegraphics[width=.092\linewidth]{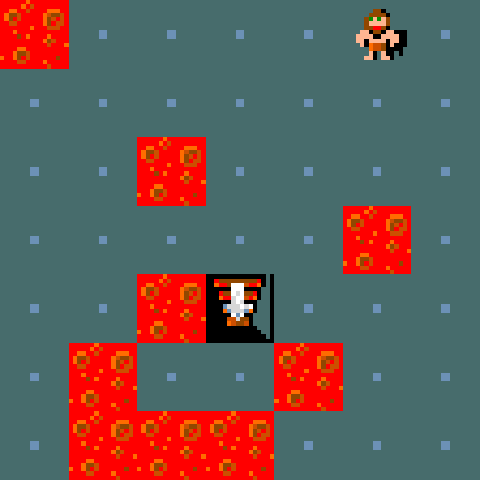}}
    \centering\subfigure{\includegraphics[width=.092\linewidth]{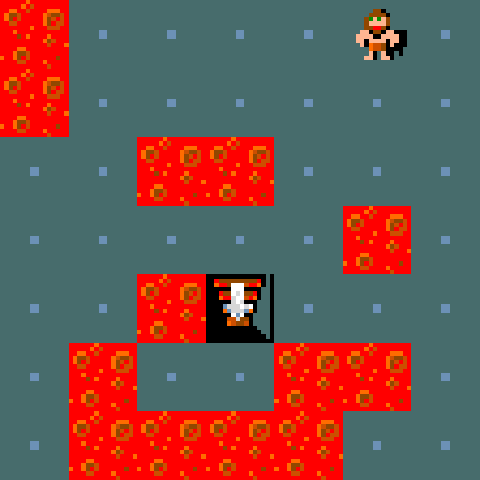}}    
    \centering\subfigure{\includegraphics[width=.092\linewidth]{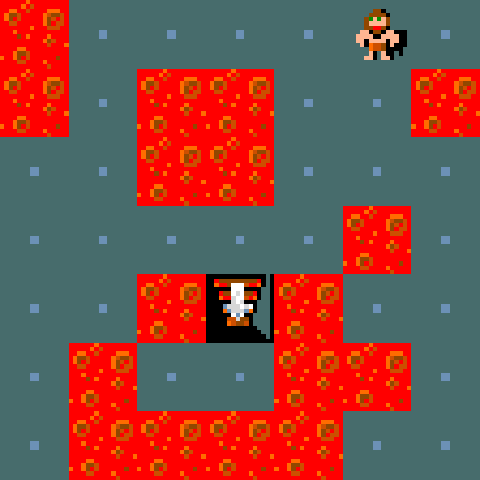}}    
    \centering\subfigure{\includegraphics[width=.092\linewidth]{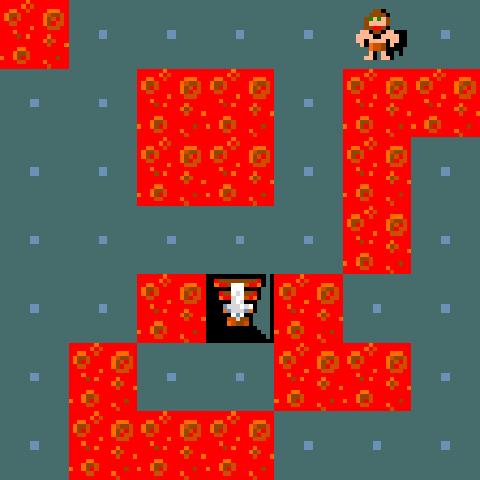}}    
    \centering\subfigure{\includegraphics[width=.092\linewidth]{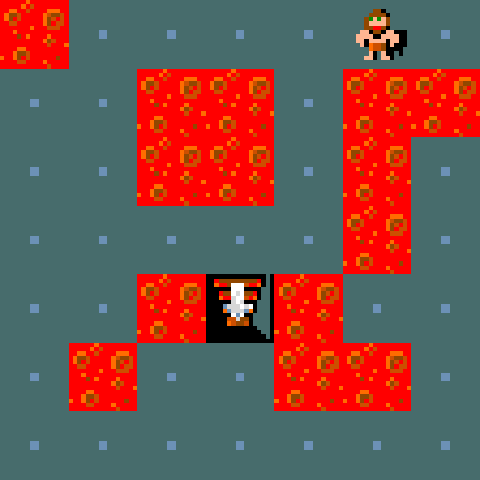}}  
    \centering\subfigure{\includegraphics[width=.092\linewidth]{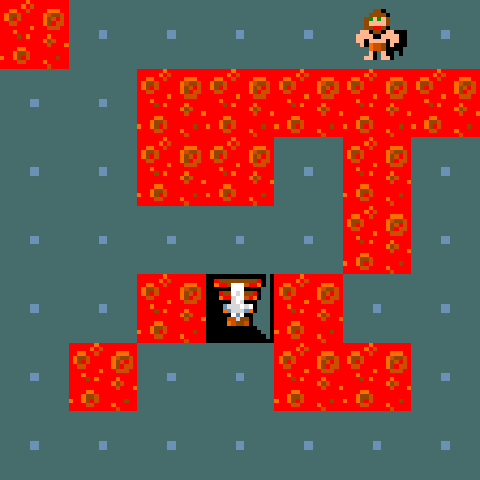}} 
    
    \centering\subfigure{\includegraphics[width=.092\linewidth]{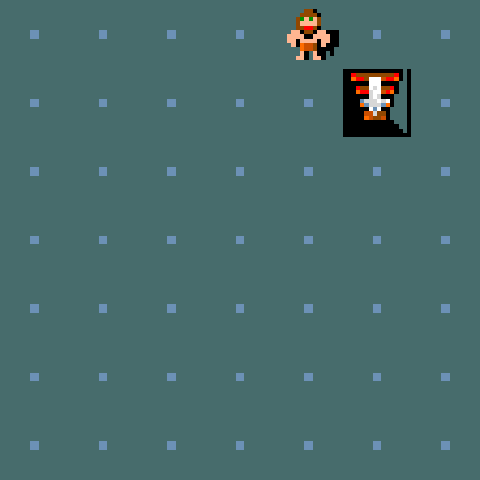}}
    \centering\subfigure{\includegraphics[width=.092\linewidth]{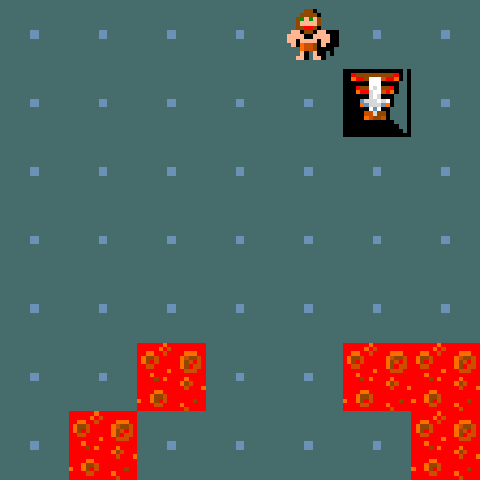}}
    \centering\subfigure{\includegraphics[width=.092\linewidth]{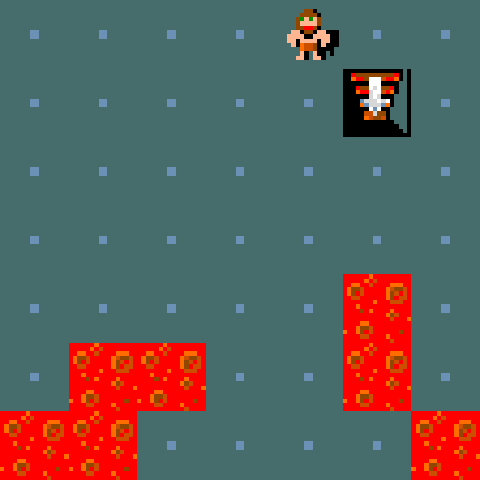}}
    \centering\subfigure{\includegraphics[width=.092\linewidth]{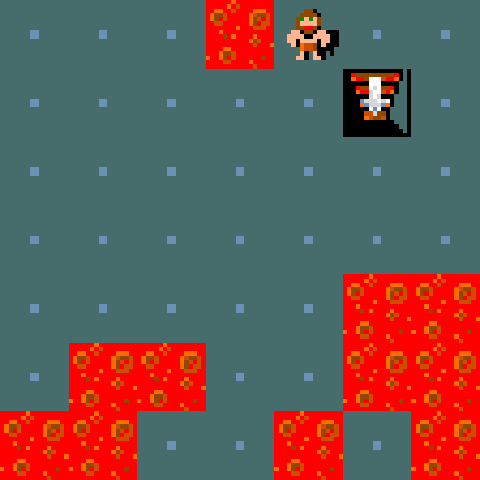}} 
    \centering\subfigure{\includegraphics[width=.092\linewidth]{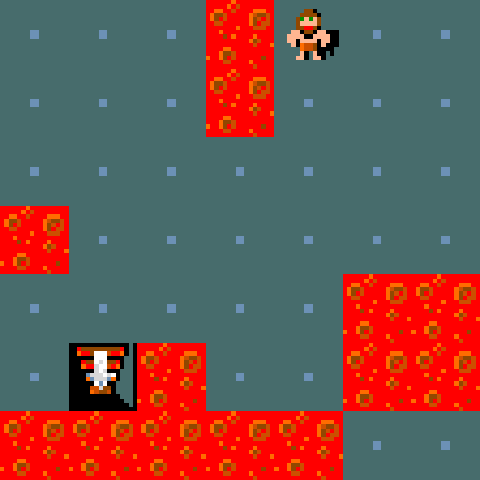}}
    \centering\subfigure{\includegraphics[width=.092\linewidth]{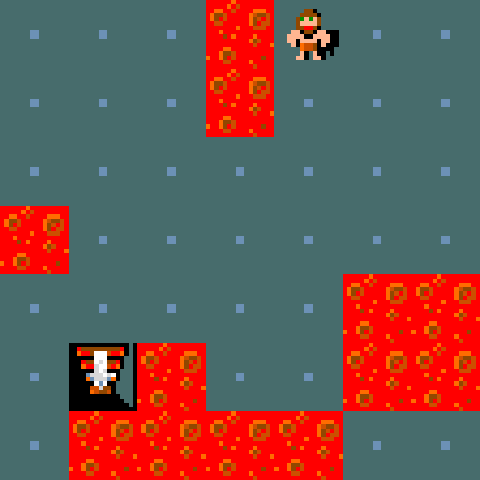}}    
    \centering\subfigure{\includegraphics[width=.092\linewidth]{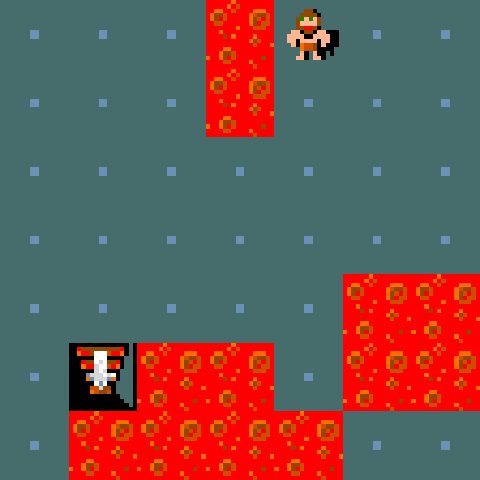}}    
    \centering\subfigure{\includegraphics[width=.092\linewidth]{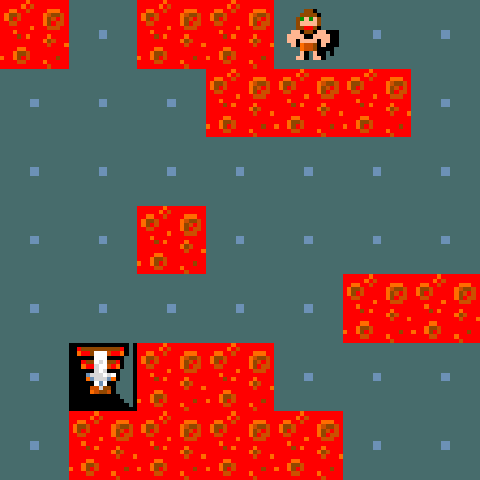}}    
    \centering\subfigure{\includegraphics[width=.092\linewidth]{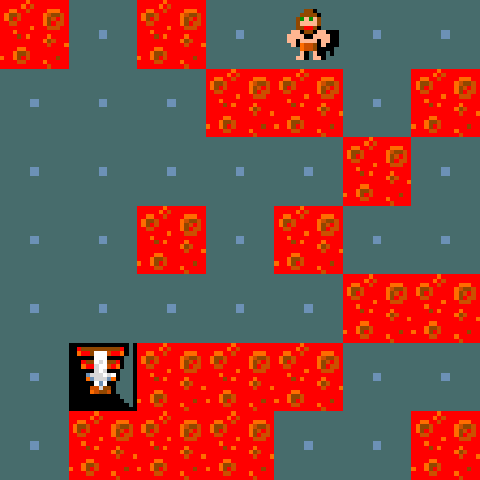}}  
    \centering\subfigure{\includegraphics[width=.092\linewidth]{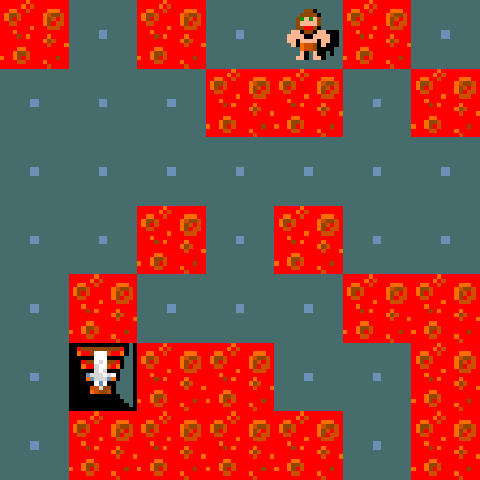}} 
    \end{minipage}
    \caption{Levels generated by \method{}. Each level along the evolutionary path is at the frontier for the student agent at that stage of training. As we can see, the edits compound to produce a series of challenges: In the first level the lava gradually surrounds the agent, such that they can initially explore in multiple directions, but near the end the task can only be solved by going down and to the right. In the middle row, we see a level where the agent has a direct path to the goal, but a corridor is evolved over time to become increasingly narrow, before being filled in so the agent has to go around it. Finally in the bottom row the level begins with simple augmentations before moving the agent behind a barrier, which results in a challenging task where the agent has to move in a diagonal direction to escape the lava.}
    \label{figure:lava7_extra_levels}
    \vspace{5mm}
    \begin{minipage}{0.99\textwidth}
    
    \centering\subfigure{\includegraphics[width=.105\linewidth]{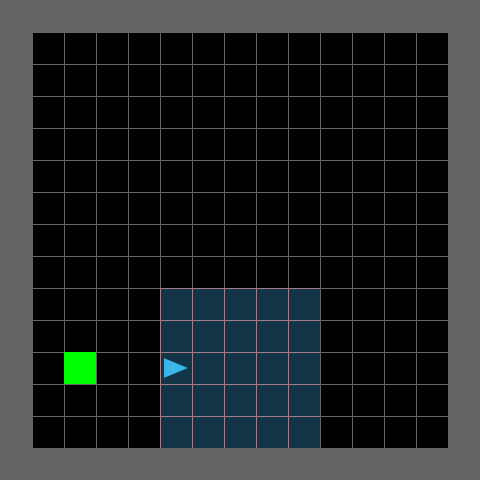}}
    \centering\subfigure{\includegraphics[width=.105\linewidth]{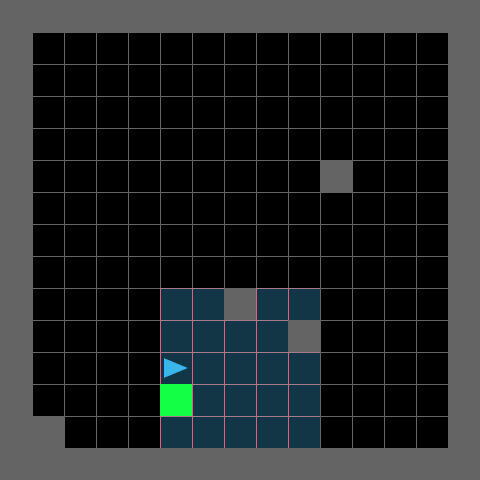}}
    \centering\subfigure{\includegraphics[width=.105\linewidth]{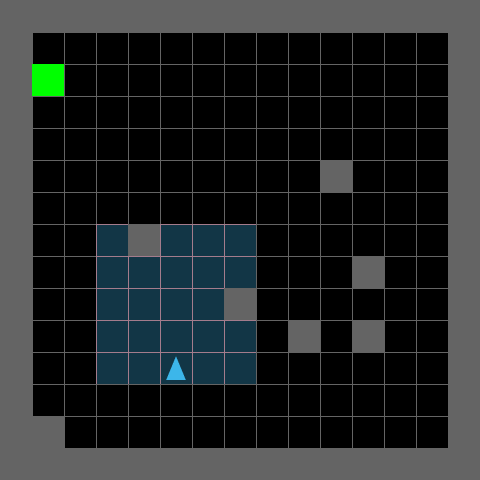}}
    \centering\subfigure{\includegraphics[width=.105\linewidth]{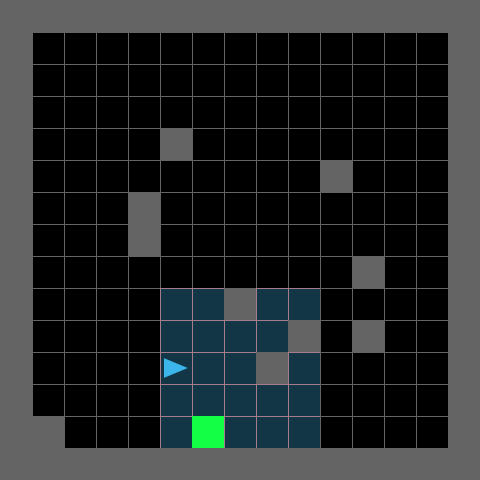}}
    \centering\subfigure{\includegraphics[width=.105\linewidth]{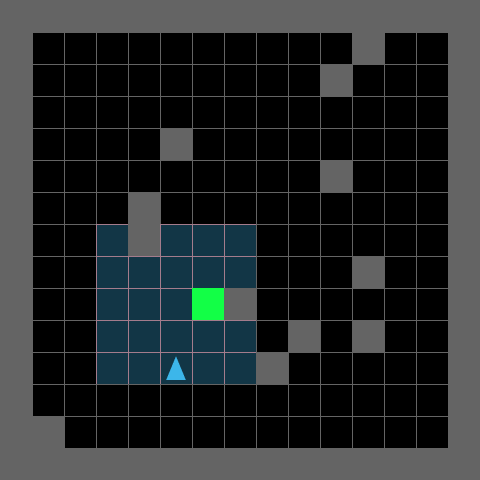}}
    \centering\subfigure{\includegraphics[width=.105\linewidth]{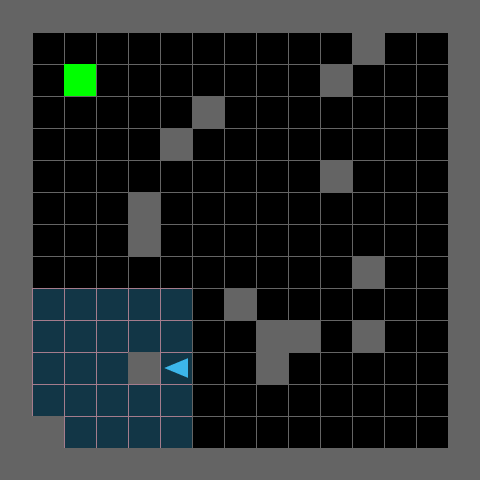}}
    \centering\subfigure{\includegraphics[width=.105\linewidth]{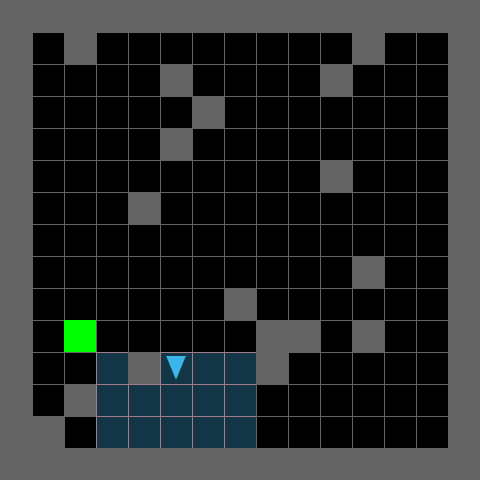}}
    \centering\subfigure{\includegraphics[width=.105\linewidth]{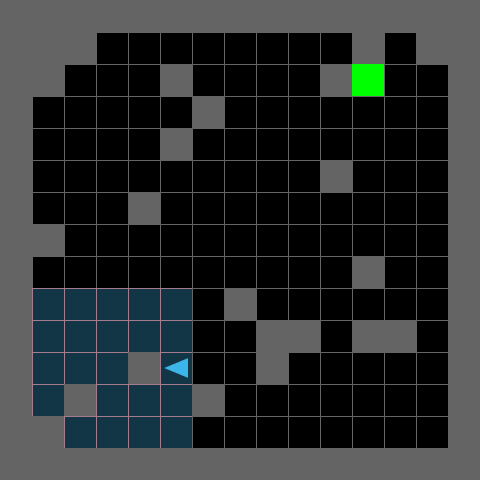}}
    \centering\subfigure{\includegraphics[width=.105\linewidth]{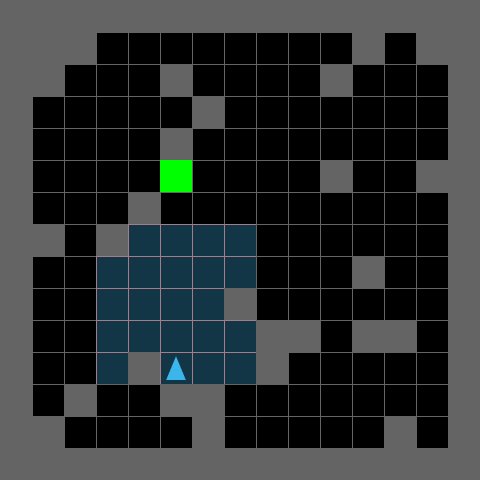}}
    \centering\subfigure{\includegraphics[width=.105\linewidth]{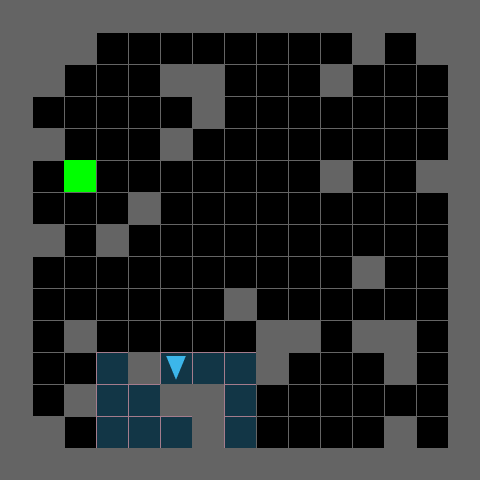}}
    \centering\subfigure{\includegraphics[width=.105\linewidth]{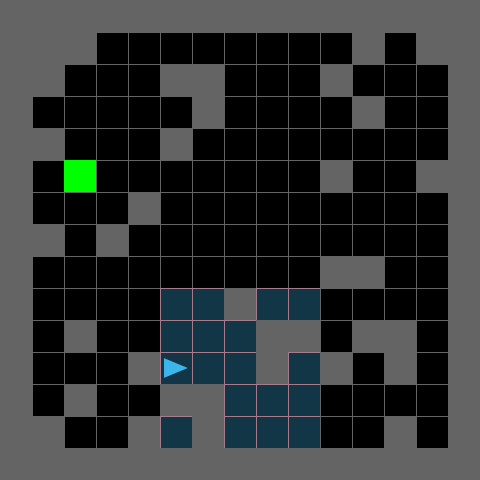}}
    \centering\subfigure{\includegraphics[width=.105\linewidth]{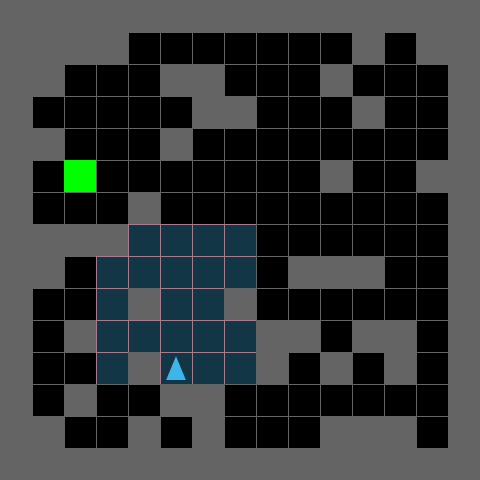}}
    \centering\subfigure{\includegraphics[width=.105\linewidth]{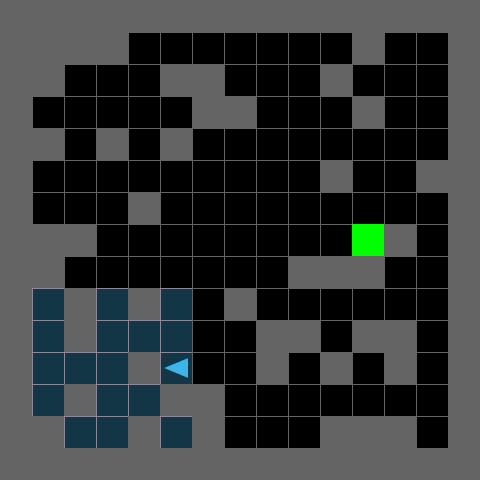}}
    \centering\subfigure{\includegraphics[width=.105\linewidth]{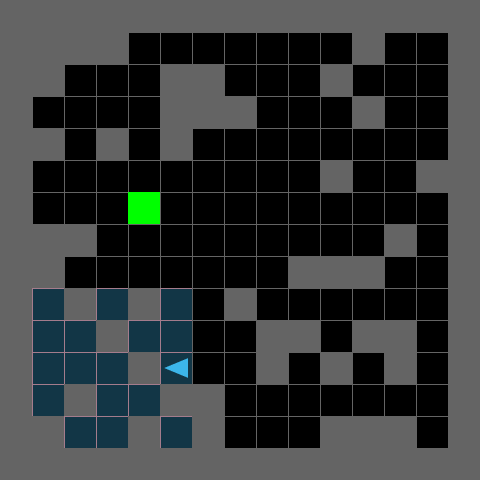}}
    \centering\subfigure{\includegraphics[width=.105\linewidth]{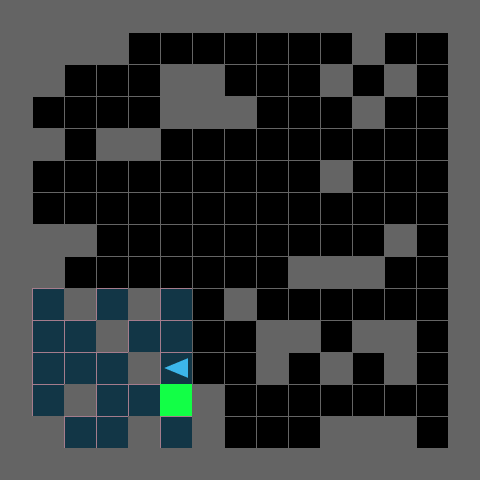}}
    \centering\subfigure{\includegraphics[width=.105\linewidth]{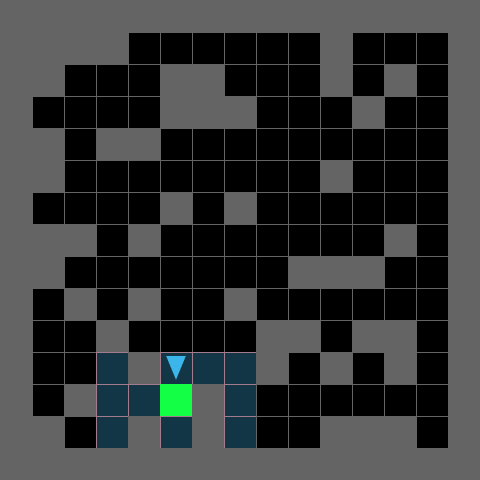}}
    \centering\subfigure{\includegraphics[width=.105\linewidth]{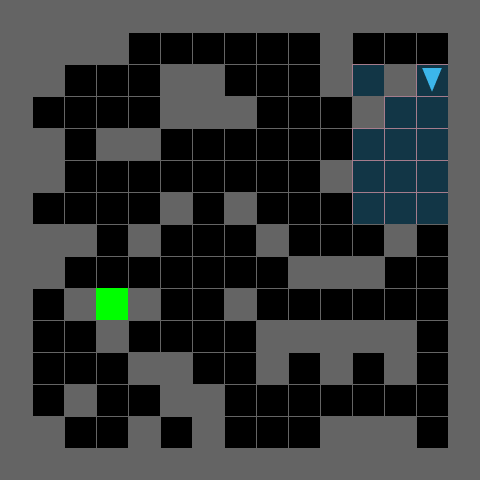}}
    \centering\subfigure{\includegraphics[width=.105\linewidth]{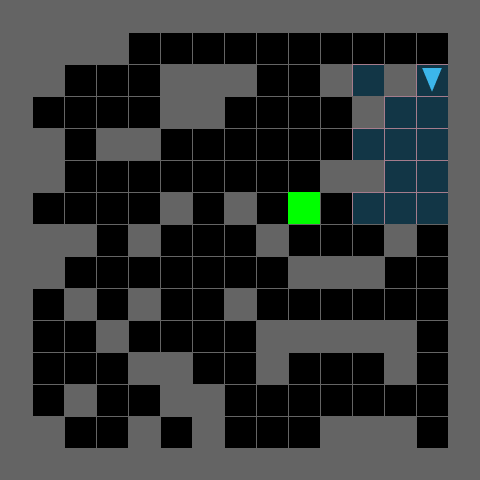}}
    \centering\subfigure{\includegraphics[width=.105\linewidth]{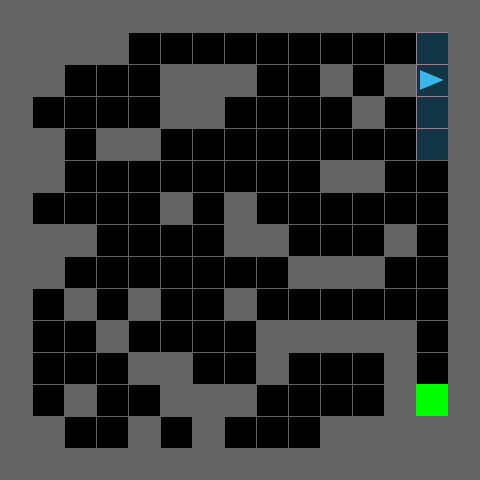}}
    \centering\subfigure{\includegraphics[width=.105\linewidth]{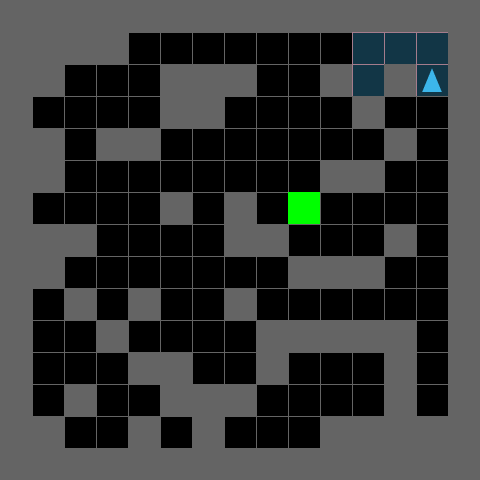}}
    \centering\subfigure{\includegraphics[width=.105\linewidth]{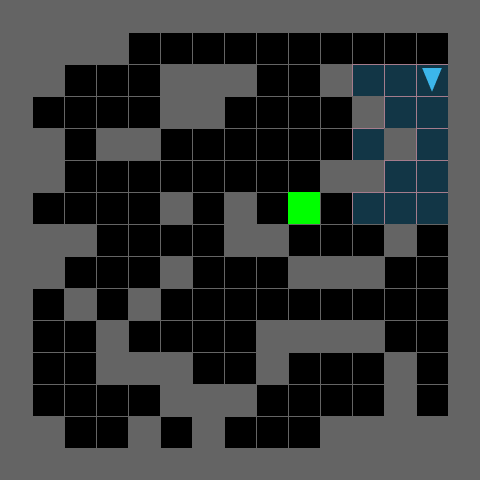}}
    \centering\subfigure{\includegraphics[width=.105\linewidth]{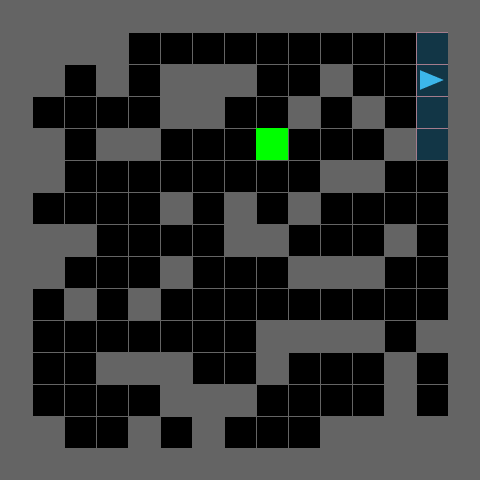}}
    \centering\subfigure{\includegraphics[width=.105\linewidth]{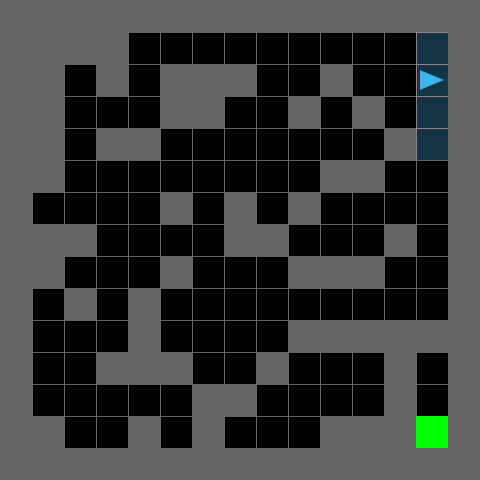}}
    \centering\subfigure{\includegraphics[width=.105\linewidth]{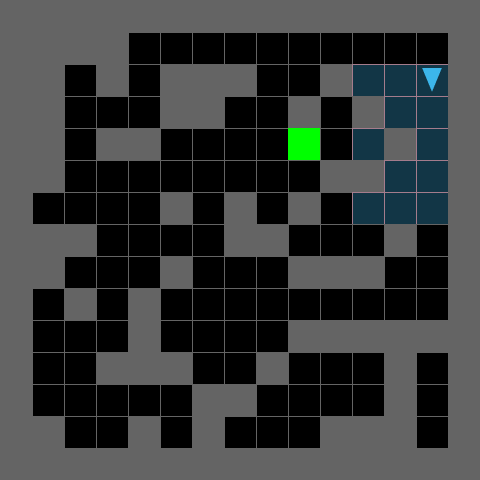}}
    \centering\subfigure{\includegraphics[width=.105\linewidth]{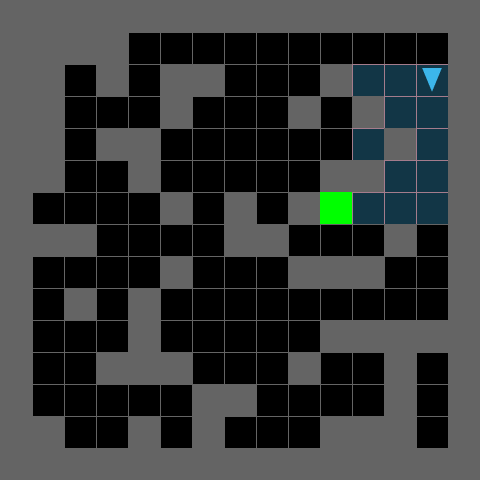}}
    \centering\subfigure{\includegraphics[width=.105\linewidth]{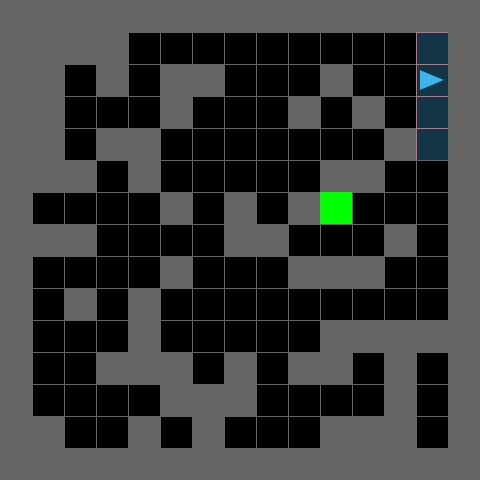}}
    \centering\subfigure{\includegraphics[width=.105\linewidth]{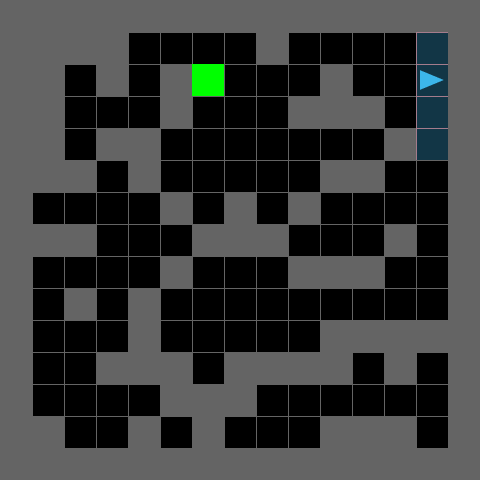}}
    \centering\subfigure{\includegraphics[width=.105\linewidth]{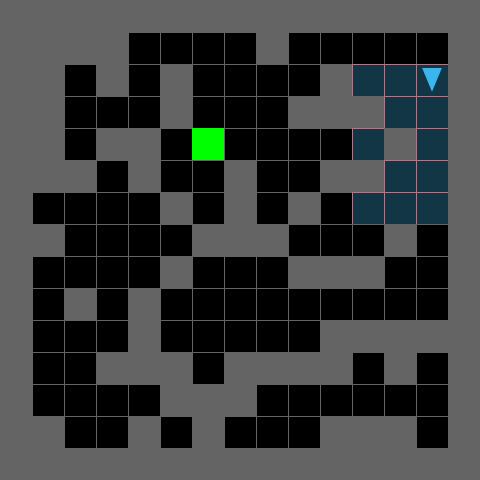}}
    \centering\subfigure{\includegraphics[width=.105\linewidth]{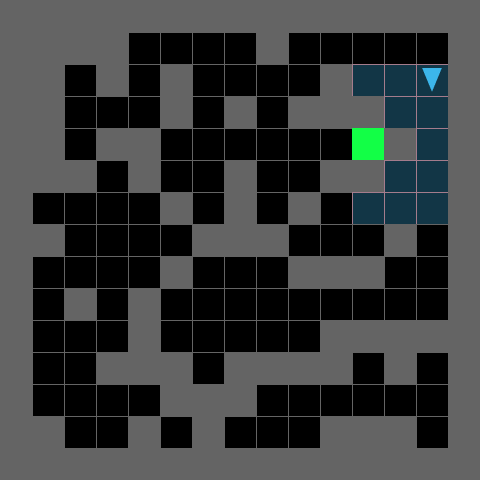}}
    \centering\subfigure{\includegraphics[width=.105\linewidth]{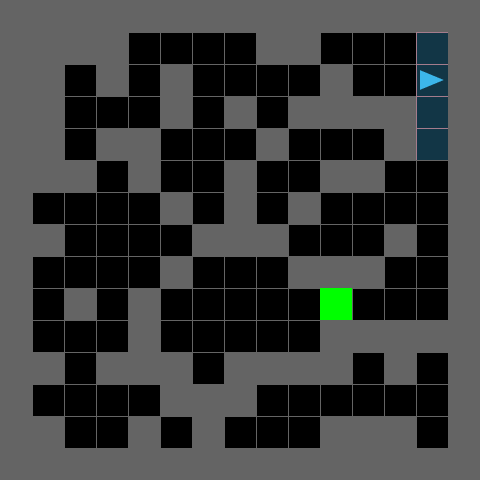}}
    \centering\subfigure{\includegraphics[width=.105\linewidth]{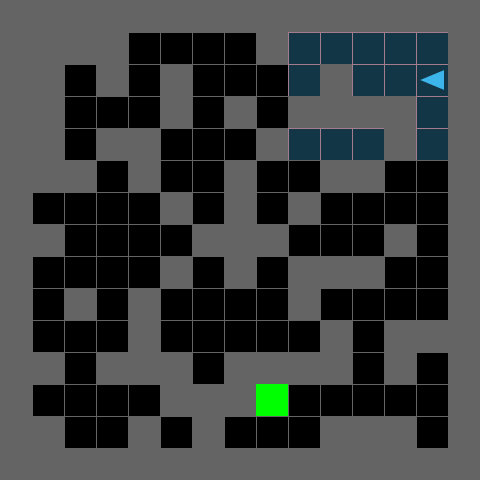}}
    \centering\subfigure{\includegraphics[width=.105\linewidth]{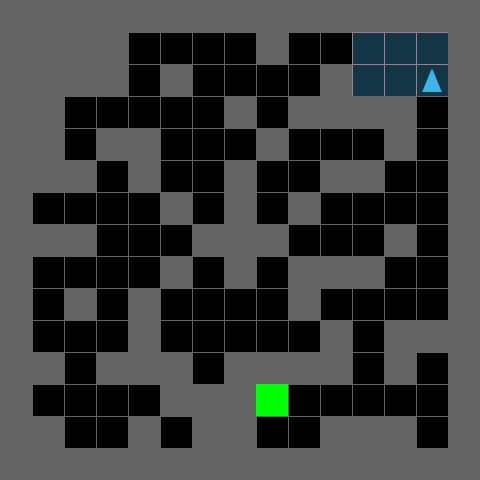}}
    \centering\subfigure{\includegraphics[width=.105\linewidth]{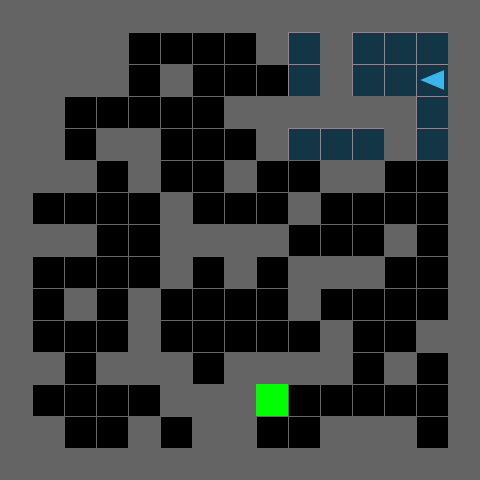}}
    \centering\subfigure{\includegraphics[width=.105\linewidth]{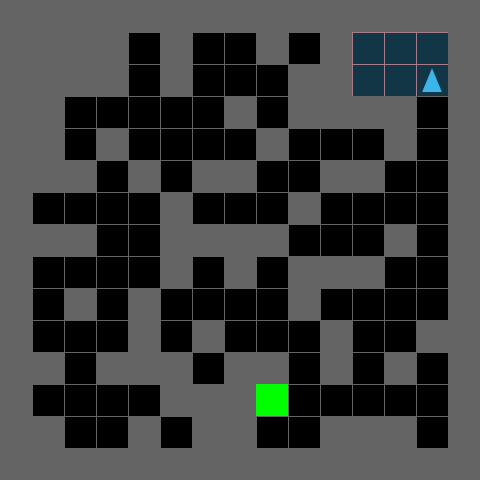}}
    \centering\subfigure{\includegraphics[width=.105\linewidth]{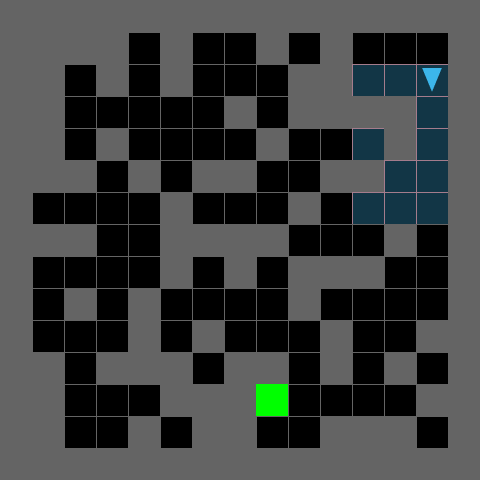}}
    \centering\subfigure{\includegraphics[width=.105\linewidth]{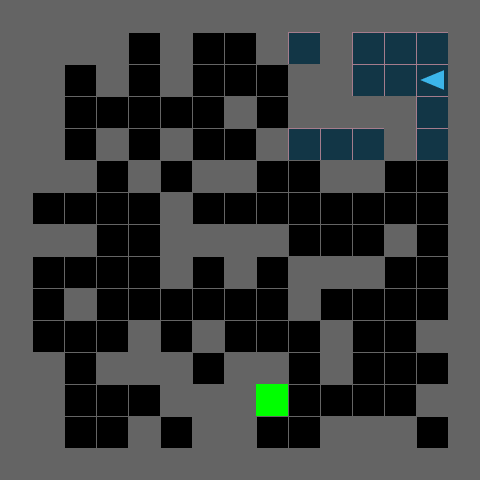}}
    \caption{The evolution of a single level the MiniGrid environment, starting from top-left, ending bottom-right. Throughout the process the agent experiences a diverse set of challenges.}
    \label{figure:mg_extra_levels}
    \end{minipage}
\end{figure}

In Fig~\ref{figure:lava7_extra_levels} and \ref{figure:mg_extra_levels}, we show levels produced by \method{} for the MiniHack lava environment and MiniGrid mazes respectively. Each step along the evolutionary process produces a level that has high learning potential at that point in training.

\begin{figure}[h!]
    \centering
    \begin{minipage}{0.99\textwidth}
    \centering\subfigure{\includegraphics[width=.99\linewidth]{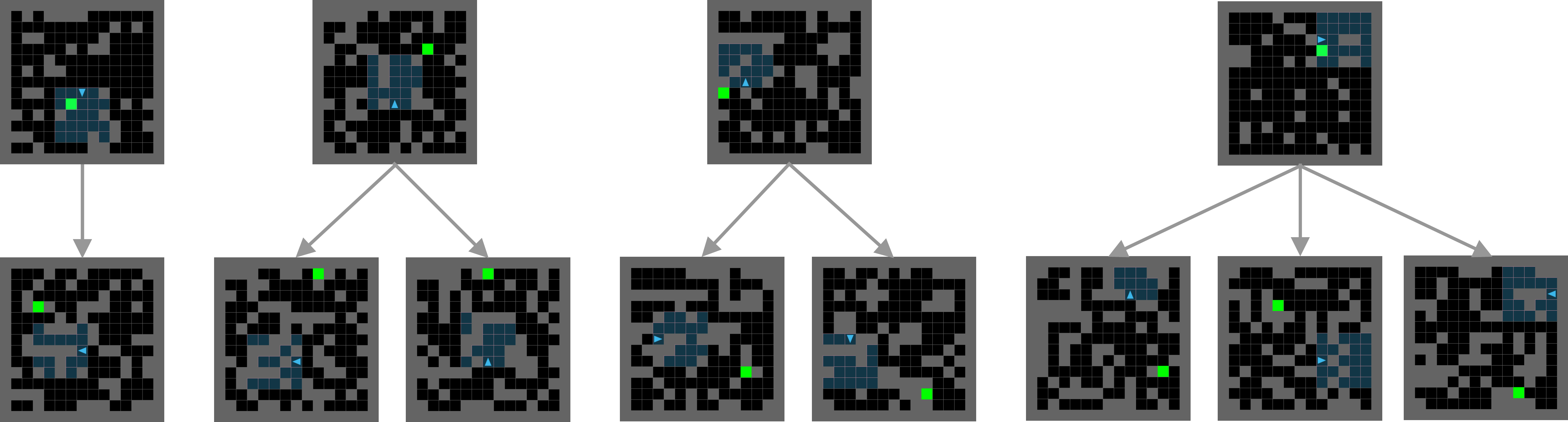}}
    \caption{Maze evolution. Top row shows starting levels, originally included in the PLR buffer due to having high positive value loss. After many edits (up to 40), ACCEL produces the bottom row, which were all selected from the top-50 levels in terms of PLR scores after 10k gradient steps. As we see, the same level can produce distinct future levels, in some cases multiple high-regret levels. 
    }
    \label{figure:mg_maze_evo}
    \end{minipage}
\end{figure}

\begin{figure}[H]
    \centering
    \begin{minipage}{0.99\textwidth}
    \centering\subfigure{\includegraphics[width=.49\linewidth]{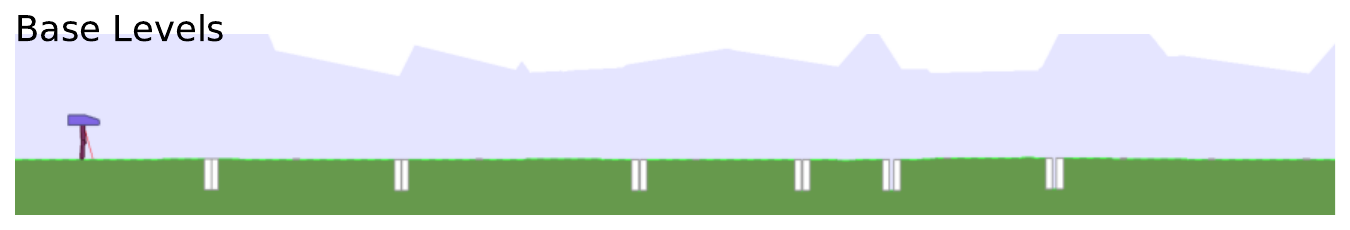}}
    \centering\subfigure{\includegraphics[width=.49\linewidth]{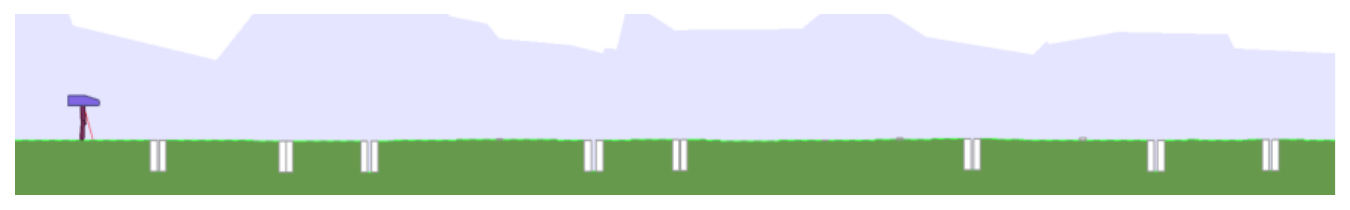}}
    \centering\subfigure{\includegraphics[width=.49\linewidth]{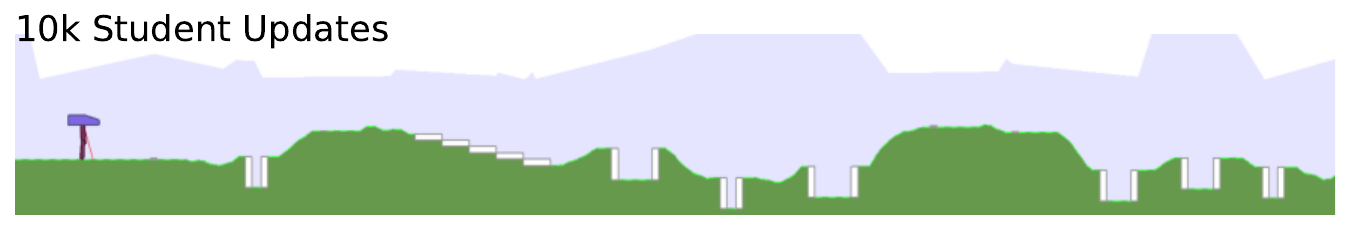}}
    \centering\subfigure{\includegraphics[width=.49\linewidth]{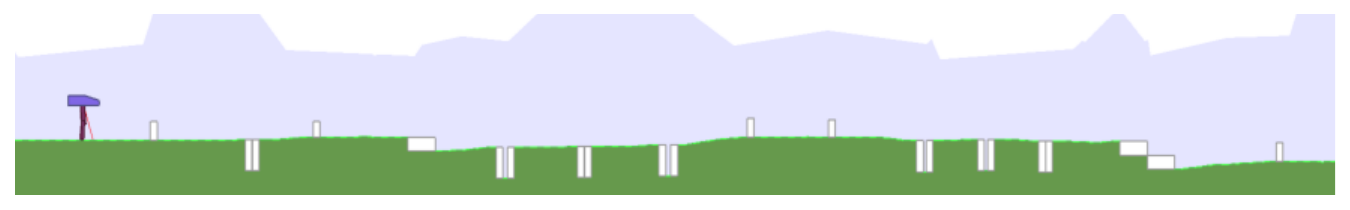}}
    \centering\subfigure{\includegraphics[width=.49\linewidth]{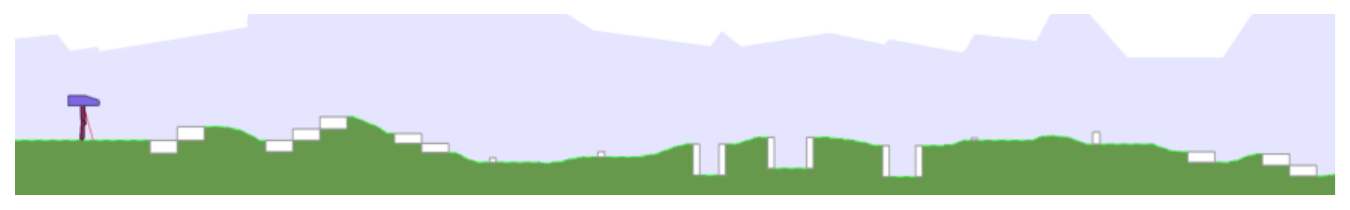}}
    \centering\subfigure{\includegraphics[width=.49\linewidth]{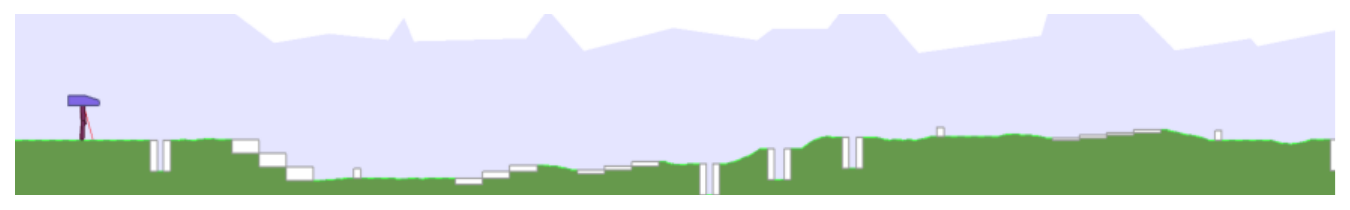}}
    \centering\subfigure{\includegraphics[width=.49\linewidth]{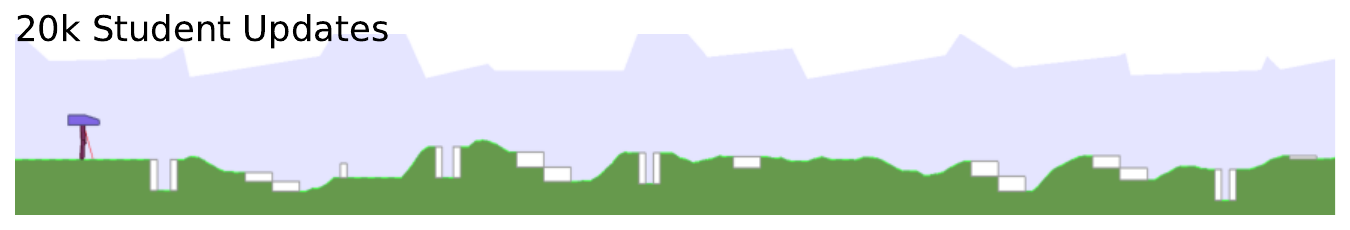}}
    \centering\subfigure{\includegraphics[width=.49\linewidth]{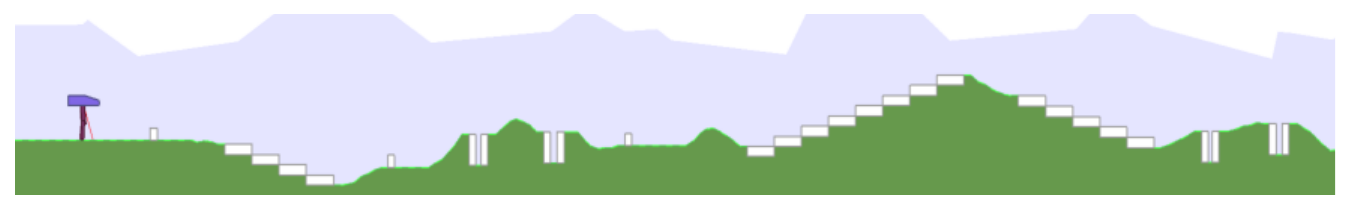}}
    \centering\subfigure{\includegraphics[width=.49\linewidth]{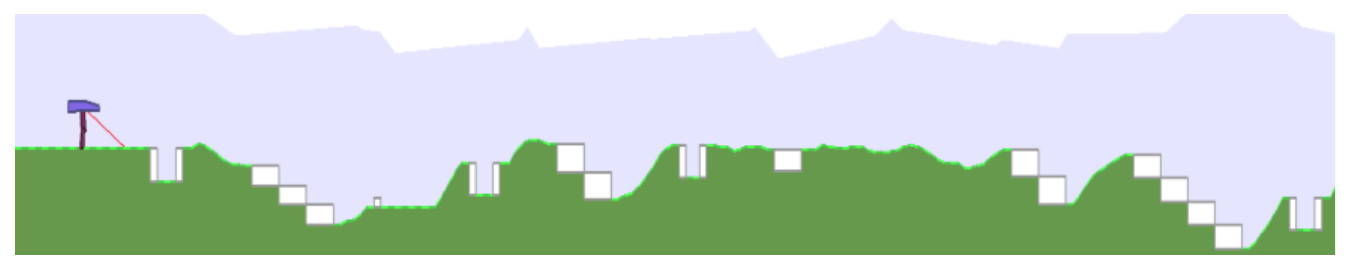}}
    \centering\subfigure{\includegraphics[width=.49\linewidth]{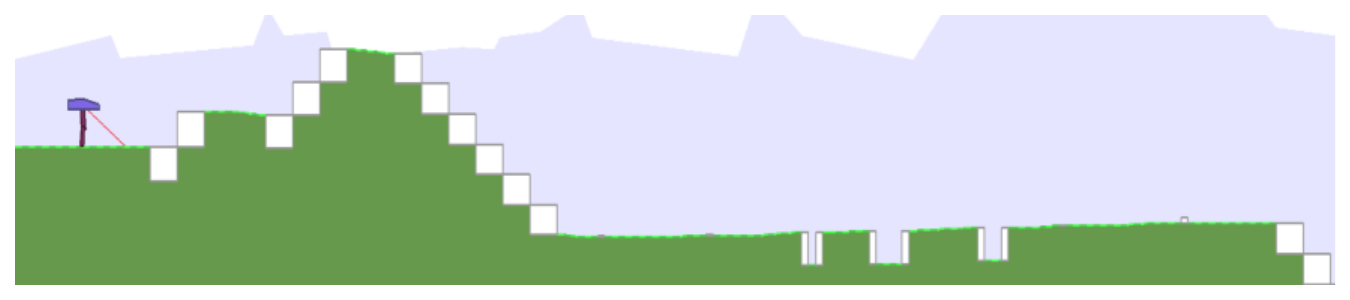}}
    \centering\subfigure{\includegraphics[width=.49\linewidth]{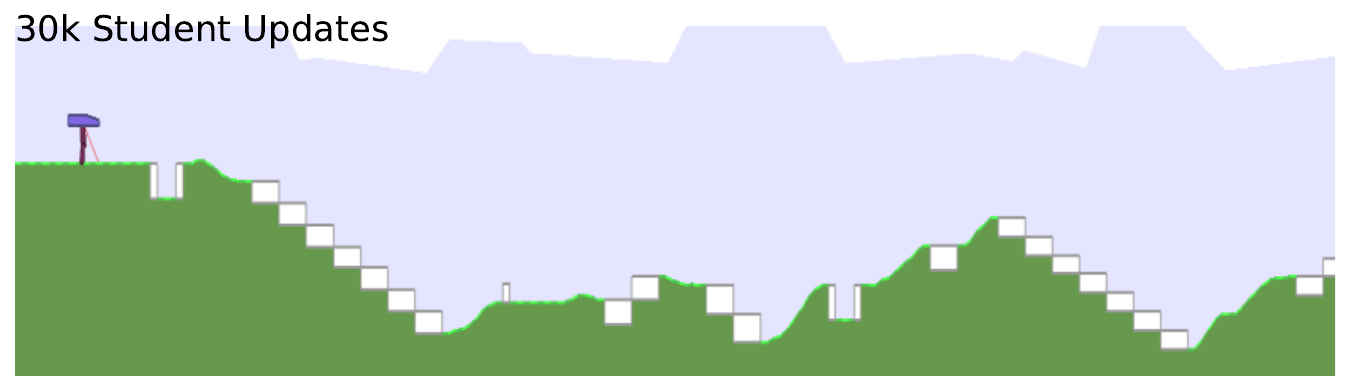}}
    \centering\subfigure{\includegraphics[width=.49\linewidth]{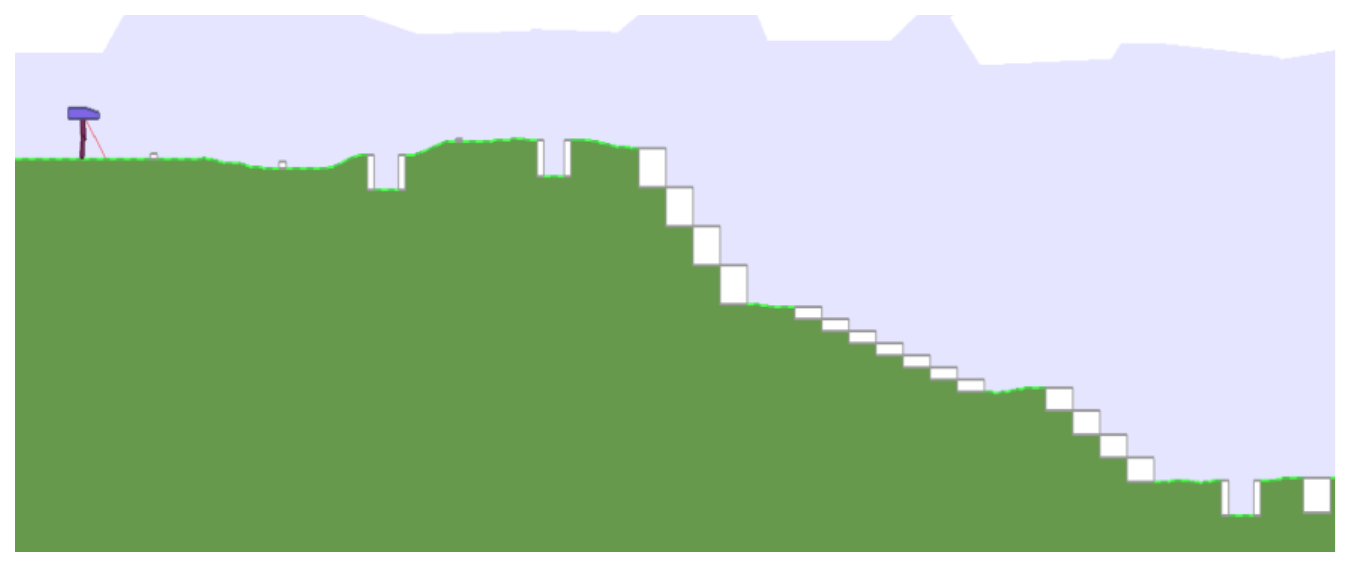}}
    \centering\subfigure{\includegraphics[width=.49\linewidth]{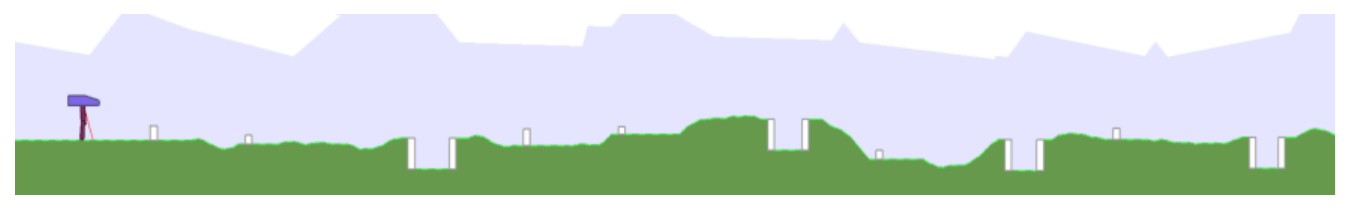}}
    \centering\subfigure{\includegraphics[width=.49\linewidth]{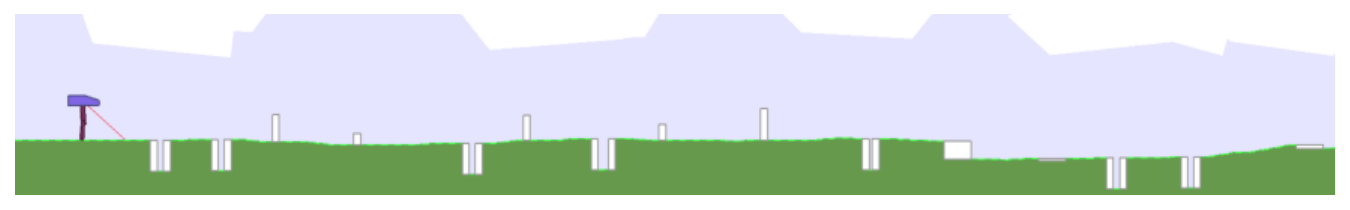}}

    \caption{Levels created and solved by \method{} in the BipedalWalker environment. At the beginning (top row) levels are initialized with low values for environment parameters, encoding the range of obstacle sizes. After several edits, we reach challenging scenarios like steep staircases or high stumps. By the end of training, we see a diverse combination of multiple challenges.
    }
    \label{figure:bipedal_evo}
    \end{minipage}
\end{figure}

\newpage
\subsection{The Expanding Frontier}

\begin{figure}[h!]
    \centering
    \begin{minipage}{0.99\textwidth}
    \centering\subfigure{\includegraphics[width=.9\linewidth]{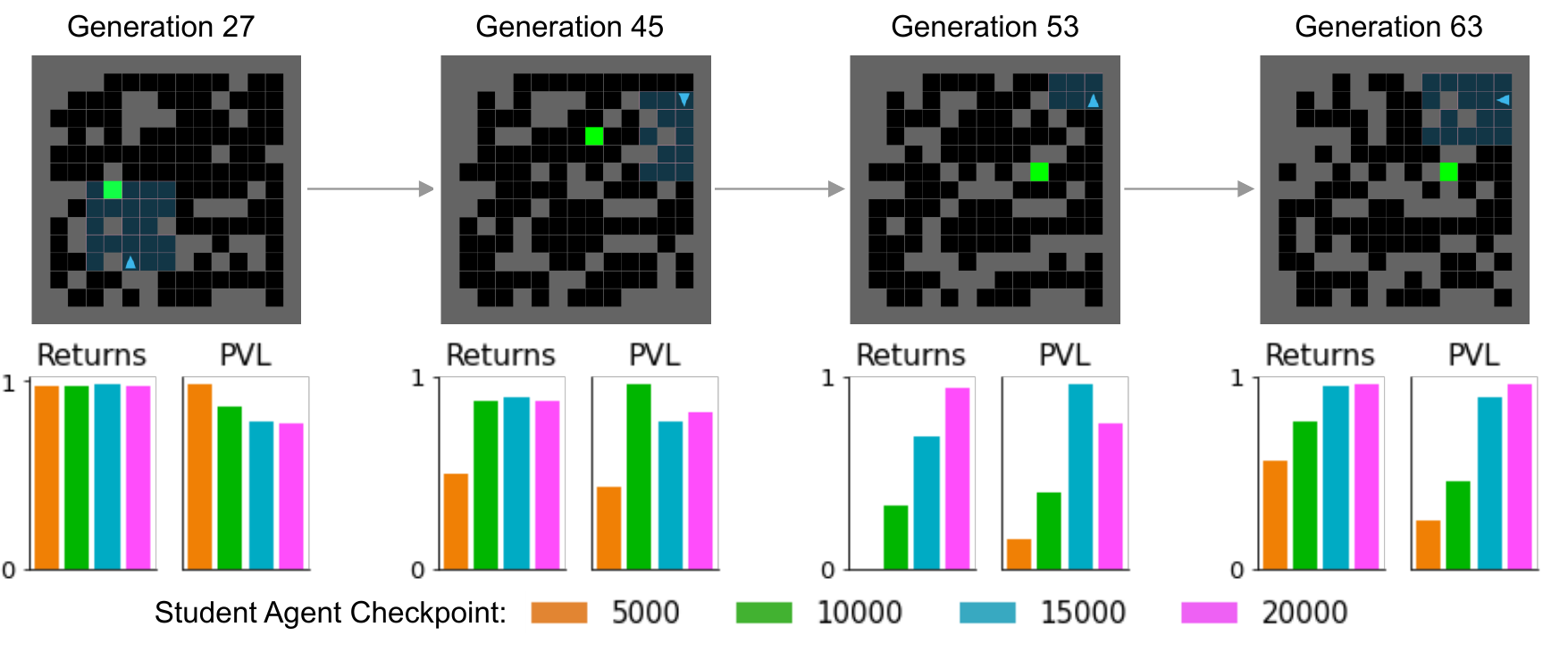}}
    \vspace{-3mm}
    \caption{The Evolving Frontier. The top row shows four levels from the same lineage, at generations 27, 45, 53 and 63. Underneath each is a bar plot showing the return and positive value loss (PVL) for four different \method{} policies, checkpointed at 5k, 10k, 15k and 20k PPO updates. At generation 27, all four checkpoints can solve the level, but the 5k checkpoint has the highest learning potential (PVL). On the right we see that by generation 63, only the 15k and 20k checkpoints are able to achieve a high return on the level.
    }
    \label{figure:mg_maze_frontier}
    \end{minipage}
\end{figure}

Here we analyze the performance of agents on levels produced by ACCEL. We inspect four agent checkpoints, from 5k, 10k, 15k and 20k student updates. In Figure~\ref{figure:mg_maze_frontier} we show four generations of a level. We see that the later generations become harder for the 5k checkpoint, while the 20k checkpoint sees the highest return from the more complex level in Gen 63. 

\begin{figure}[h!]
    \centering
    \begin{minipage}{0.99\textwidth}
    \centering\subfigure{\includegraphics[width=.9\linewidth]{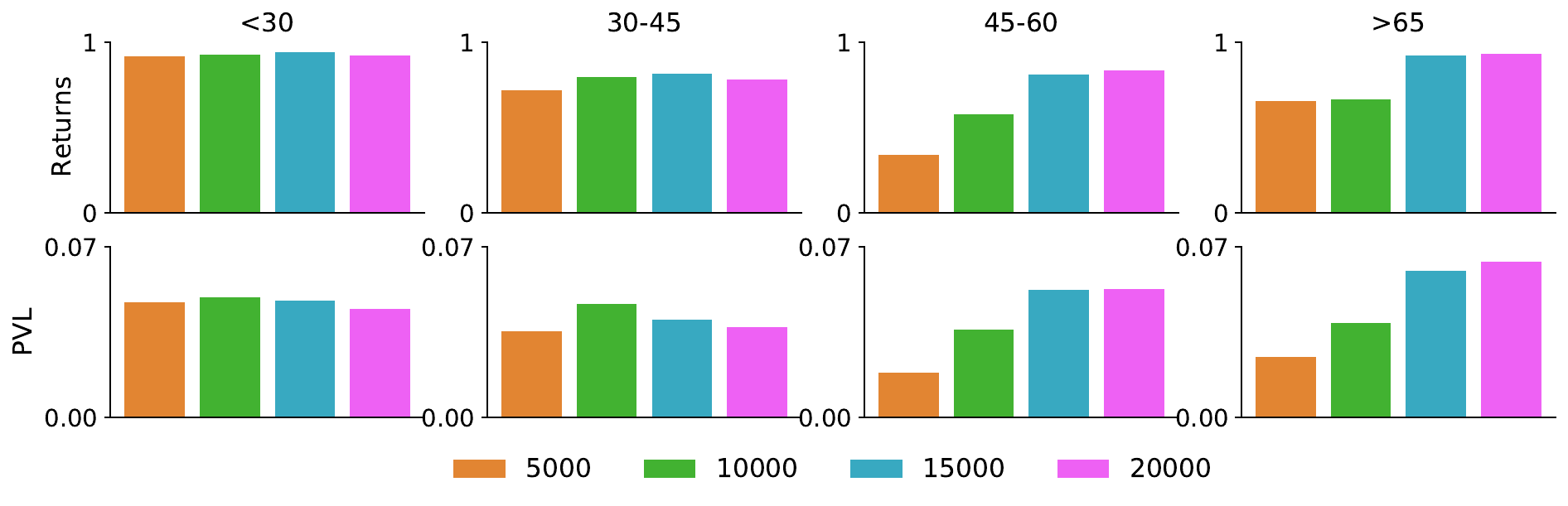}}
    \vspace{-5mm}
    \caption{Aggregate metrics for each band of generations. For example, ``30-45'' refers to all the levels between generation 30-45. The later generation levels are harder for the early agents to solve, while the early agents have higher return and PVL for the earlier levels.
    }
    \label{figure:mg_maze_frontier_mean}
    \end{minipage}
\end{figure}

\begin{figure}[h!]
    \centering
    \begin{minipage}{0.99\textwidth}
    \centering\subfigure{\includegraphics[width=.8\linewidth]{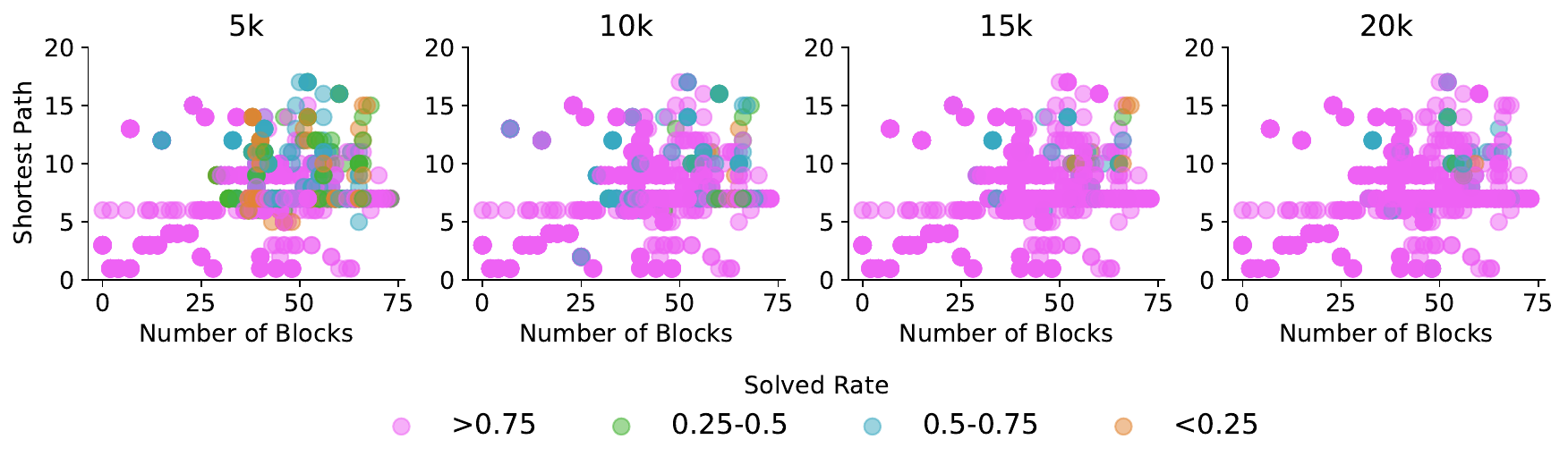}}
    \vspace{-5mm}
    \caption{How do complexity metrics relate to difficulty? The plot shows the block count and shortest path length. From left to right we evaluate the agents at four checkpoints: 5k, 10k, 15k, 20k PPO updates. The color represents the solved rate. As we see, the 5k agent is unable to solve the levels with higher block count and longer paths to the goal, while the 20k agent is able to solve almost all levels.
    }
    \label{figure:mg_maze_frontier_srvsdifficulty}
    \end{minipage}
\end{figure}

In Figure~\ref{figure:mg_maze_frontier_mean} we show all generations for the level included in Figure~\ref{figure:mg_maze_frontier}, grouped by generation. We then show the mean return and PVL for all four agent checkpoints. We find the 20k checkpoint sees the highest learning potential in levels from later generations, while the 5k checkpoint sees the lowest return on these levels. Next in Figure~\ref{figure:mg_maze_frontier_srvsdifficulty} we show data for all generations of 20 levels which were present in the 20k checkpoint replay buffer and their ancestors in the 5k checkpoint buffer. For each checkpoint, we visualize the solved rate for each of these levels based the color of the point for each level, plotted along axes corresponding to shortest path length and number of blocks. The 5k checkpoint can only reliably solve the shorter-path levels with low block count. In contrast, the 20k checkpoint performs well across all sampled levels.

\newpage
\subsection{Full Experimental Results}
\label{appendix:full_results}

\begin{figure}[H]
    \centering
    \begin{minipage}{0.99\textwidth}
    \centering\subfigure{\includegraphics[width=.99\linewidth]{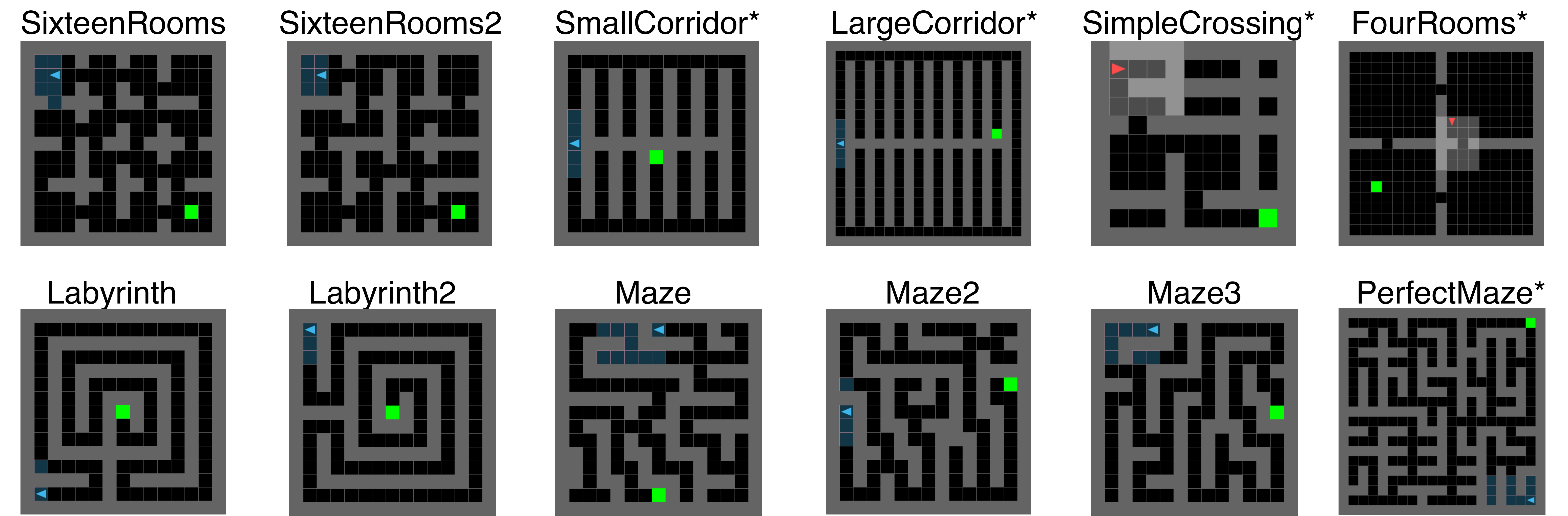}}
    \caption{MiniGrid Zero-Shot Environments. Those with an asterisk are procedurally generated: For SmallCorridor and LargeCorridor, the position of the goal can be in any of the corridors. SimpleCrossing and FourRooms are from \citet{gym_minigrid}, and PerfectMaze, from \citet{jiang2021robustplr}.}
    \label{figure:minigrid_zs_levels}
    \end{minipage}
\end{figure}

\textbf{Partially-Observable Navigation} Next we show the extended results for the MiniGrid experiments. We use a series of challenging zero-shot environments (see Figure~\ref{figure:minigrid_zs_levels}), introduced in the UED literature \citep{paired, jiang2021robustplr}. We include the full results in Table \ref{table:minigridresults} and bar plots of the same data in Figure \ref{figure:mg_zs_results}. 

\begin{figure}[H]
    \centering
    \begin{minipage}{0.99\textwidth}
    \centering\subfigure{\includegraphics[width=.55\linewidth]{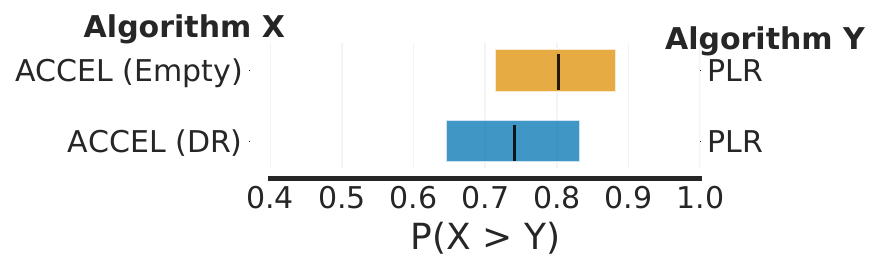}}
    \vspace{-3mm}
    \caption{Probability of improvement of \method{} over PLR across all evaluation environments in Figure~\ref{figure:minigrid_zs_levels}, using the open-source notebook from \citet{agarwal2021deep}. The probability of improvement represents the probability that Algorithm X outperforms Algorithm Y on a new task from the same distribution.}
    \label{figure:minigrid_prob_improve}
    \end{minipage}
\end{figure}

We also include an additional version of \method{} using the same generator as the DR and PLR baselines, thus editing more complex base levels. This removes the prior that it is beneficial to begin with simple empty rooms. As we see, both versions of \method{} significantly outperform the baselines. Particularly in the more complex environments like Labyrinth, we see large gains compared to baselines. Also note that PLR outperforms all other baselines, and \method{} outperforms PLR. We show report these results in the Table~\ref{figure:minigrid_zs_levels}, as well as show more robust comparison metrics provided by \texttt{rliable} \citep{agarwal2021deep} in Figure~\ref{figure:minigrid_prob_improve}.

\begin{figure}[H]
    \centering
    \begin{minipage}{0.99\textwidth}
    \centering\subfigure{\includegraphics[width=.99\linewidth]{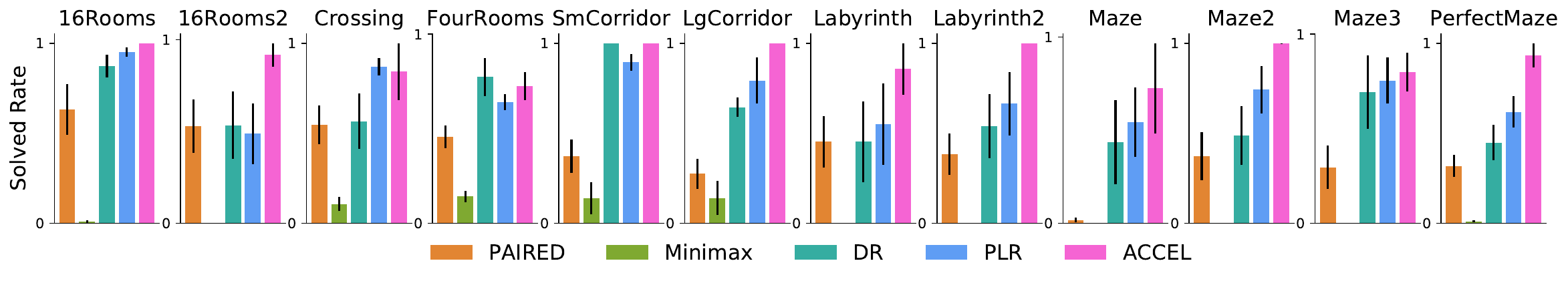}}
    \vspace{-3mm}
    \caption{Zero-shot transfer results in out-of-distribution mazes. Agents are evaluated for 100 episodes on human-designed mazes. Plots show mean and standard error for each environment, across five runs.
    }
    \label{figure:mg_zs_results}
    \end{minipage}
\end{figure}

\begin{table}[H]
\begin{center}
\caption{Zero-Shot transfer to human-designed environments. Each entry corresponds to the mean and standard error of 5 training runs, where each run is evaluated for 100 trials on each environment. $\dagger$ indicates the generator first samples the number of blocks to place in $[0, 60]$, then places that many at random locations. $\ddagger$ indicates the generator produces only empty rooms. Bold values are within one standard error of the best mean. $\star$ indicates a statistically significant improvement against PLR ($p<0.05$ via Welch's t-test). All methods are evaluated after 20k student updates, aside from PAIRED and Minimax, which are evaluated at $\approx$30k updates.
}
\label{table:minigridresults}
\scalebox{0.95}{
\begin{tabular}{ l | cccccc } 
\toprule
\textbf{Environment} & PAIRED  & Minimax & DR$\dagger$ & PLR$\dagger$ & \method{}$\dagger$  & \method{}$\ddagger$ \\ 

\midrule
16Rooms & $0.63 \pm 0.14$ & $0.01 \pm 0.01$  & $0.87 \pm 0.06$&  $0.95 \pm 0.03$ & $\mathbf{1.0 \pm 0.0}$ & $\mathbf{1.0 \pm 0.0}$ \\
16Rooms2 & $0.53 \pm 0.15$ & $0.0 \pm 0.0$ & $0.53 \pm 0.18$ & $0.49 \pm 0.17$ & $0.62 \pm 0.22$ & $\mathbf{0.92 \pm 0.06}$ \\
SimpleCrossing & $0.55 \pm 0.11$ & $0.11 \pm 0.04$ & $0.57 \pm 0.15$ & $\mathbf{0.87 \pm 0.05}$ & $\mathbf{0.92 \pm 0.08}$ & $\mathbf{0.84 \pm 0.16}$ \\
FourRooms & $0.46 \pm 0.06$ & $0.14 \pm 0.03$ & $0.77 \pm 0.1$ & $0.64 \pm 0.04$ & $\mathbf{0.9 \pm 0.08}$ & $0.72 \pm 0.07$ \\
SmallCorridor & $0.37 \pm 0.09$ & $0.14 \pm 0.09$ & $\mathbf{1.0 \pm 0.0}$ & $0.89 \pm 0.05$ & $0.88 \pm 0.11$ & $\mathbf{1.0 \pm 0.0}$ \\
LargeCorridor & $0.27 \pm 0.08$ & $0.14 \pm 0.09$ & $0.64 \pm 0.05$ & $0.79 \pm 0.13$ & $0.94 \pm 0.05$ & $\mathbf{1.0 \pm 0.0}$ \\
Labyrinth & $0.45 \pm 0.14$ & $0.0 \pm 0.0$ & $0.45 \pm 0.23$ & $0.55 \pm 0.23$ & $\mathbf{0.97 \pm 0.03}$ & $0.86 \pm 0.14$ \\
Labyrinth2 & $0.38 \pm 0.12$ & $0.0 \pm 0.0$ & $0.54 \pm 0.18$ & $0.66 \pm 0.18$ & $\mathbf{1.0 \pm 0.01}$ & $\mathbf{1.0 \pm 0.0}$ \\
Maze & $0.02 \pm 0.01$ & $0.0 \pm 0.0$ & $0.43 \pm 0.23$ & $\mathbf{0.54 \pm 0.19}$ & $\mathbf{0.52 \pm 0.26}$ & $\mathbf{0.72 \pm 0.24}$ \\
Maze2 & $0.37 \pm 0.13$ & $0.0 \pm 0.0$ & $0.49 \pm 0.16$ & $0.74 \pm 0.13$ & $0.93 \pm 0.04$ & $\mathbf{1.0 \pm 0.0}$ \\
Maze3 & $0.3 \pm 0.12$ & $0.0 \pm 0.0$ & $0.69 \pm 0.19$ & $0.75 \pm 0.12$ & $\mathbf{0.94 \pm 0.06}$ & $0.8 \pm 0.1$ \\
PerfectMaze (M) & $0.32 \pm 0.06$ & $0.01 \pm 0.0$ & $0.45 \pm 0.1$ & $0.62 \pm 0.09$ & $\mathbf{0.88 \pm 0.12}$ & $\mathbf{0.93 \pm 0.07}$ \\
\midrule
\textbf{Mean} & $0.39 \pm 0.03$ & $0.05 \pm 0.01$ & $0.62 \pm 0.05$ & $0.71 \pm 0.04$ & $\mathbf{0.88 \pm 0.04}^\star$ & $\mathbf{0.9 \pm 0.03}^\star$ \\
\bottomrule
\end{tabular}}
\end{center}
\end{table}

\textbf{BipedalWalker} For the BipedalWalker environment, we test agents on each of the individual challenges encoded in the environment parameterization. Specifically, we evaluate agents in the following four environments: 
\begin{itemize}
    \item \texttt{Stairs}: The stair height parameters are set to [2,2] with the number of steps set to 5.
    \item \texttt{PitGap}: The pit gap parameter is set to [5,5].
    \item \texttt{Stump}: The stump parameter is set to [2,2].
    \item \texttt{Roughness}: The ground roughness parameter is set to 5.
\end{itemize}
Each of these environments is visualized in Figure~\ref{figure:bipedal_trainperf} in the main paper. We also test agents on the simple \texttt{BipedalWalker-v3} environment and the more challenging \texttt{BipedalWalkerHardcore-v3} environment. For \texttt{BipedalWalkerHardcore-v3}, we note that none of our agents fully solve the environment, which is considered to be a mean reward $ge 300$ over 100 independent evaluations. To test whether it is possible with our base RL algorithm and agent model, we trained an identical PPO agent directly on the environment for 1B steps. The reward achieved was 239---indistinguishable from that achieved by \method{}, which additionally robustifies the agent against a much wider range of environments, including the individual challenges described above.

\newpage

\begin{table}[H]
\begin{center}
\caption{Test performance on a variety of challenging evaluation environments. Each entry corresponds to the mean and standard error of 10 independent runs, where each run is evaluated for 100 trials on each environment. $\dagger$ indicates the generator creates each level with obstacle parameters uniformly sampled between the corresponding minimum value of the ``Easy Init'' range and max value defined in Table~\ref{table:bipedal_params}. $\ddagger$ indicates the generator instead uniformly samples obstacle parameters within the ``Easy Init'' ranges. Bold indicates being within one standard error of the best mean. All methods are evaluated at 30k updates.
}
\label{table:bipedalresults}
\scalebox{0.85}{
\begin{tabular}{ l | ccccccc } 
\toprule
\textbf{Environment} & PAIRED  & Minimax & ALP-GMM & DR$\dagger$ & PLR$\dagger$ & \method{}$\dagger$  & \method{}$\ddagger$ \\ 

\midrule
Basic & $206.5 \pm 30.3$ & $154.3 \pm 59.2$ & $301.5 \pm 11.6$ & $261.9 \pm 19.3$ & $304.1 \pm 1.8$ & $316.9 \pm 2.1$ & $\mathbf{318.1 \pm 1.0}$ \\
Hardcore & $-47.2 \pm 10.6$ & $-44.3 \pm 1.6$ & $29.7 \pm 9.9$ & $23.8 \pm 8.3$ & $82.6 \pm 8.5$ & $163.3 \pm 30.9$ & $\mathbf{236.0 \pm 8.9}$ \\
Stairs & $-27.4 \pm 12.1$ & $-2.6 \pm 2.6$ & $22.1 \pm 6.3$ & $23.3 \pm 4.4$ & $48.0 \pm 4.3$ & $59.4 \pm 10.5$ & $\mathbf{91.7 \pm 8.9}$ \\
PitGap & $-68.2 \pm 9.7$ & $-79.3 \pm 0.5$ & $\mathbf{98.8 \pm 24.9}$ & $11.0 \pm 7.6$ & $46.2 \pm 11.3$ & $49.6 \pm 12.6$ & $\mathbf{133.3 \pm 39.1}$ \\
Stump & $-76.0 \pm 10.3$ & $-65.0 \pm 18.4$ & $-22.4 \pm 17.2$ & $-5.4 \pm 5.5$ & $7.5 \pm 6.4$ & $44.6 \pm 49.8$ & $\mathbf{188.8 \pm 10.9}$ \\
Roughness & $-5.1 \pm 25.9$ & $-1.2 \pm 7.7$ & $44.7 \pm 11.6$ & $52.3 \pm 9.0$ & $126.7 \pm 7.3$ & $211.7 \pm 21.5$ & $\mathbf{248.9 \pm 12.3}$ \\
\midrule
\textbf{Mean} & $-2.9 \pm 14.5$ & $-6.3 \pm 24.6$ & $79.1 \pm 17.5$ & $61.1 \pm 12.6$ & $102.5 \pm 13.0$ & $140.9 \pm 23.0$ & $\mathbf{202.8 \pm 13.6}$ \\
\bottomrule
\end{tabular}}
\end{center}
\end{table}

\begin{figure}[H]
    \centering
    \vspace{-8mm}
    \begin{minipage}{0.99\textwidth}
    \centering\subfigure{\includegraphics[width=.99\linewidth]{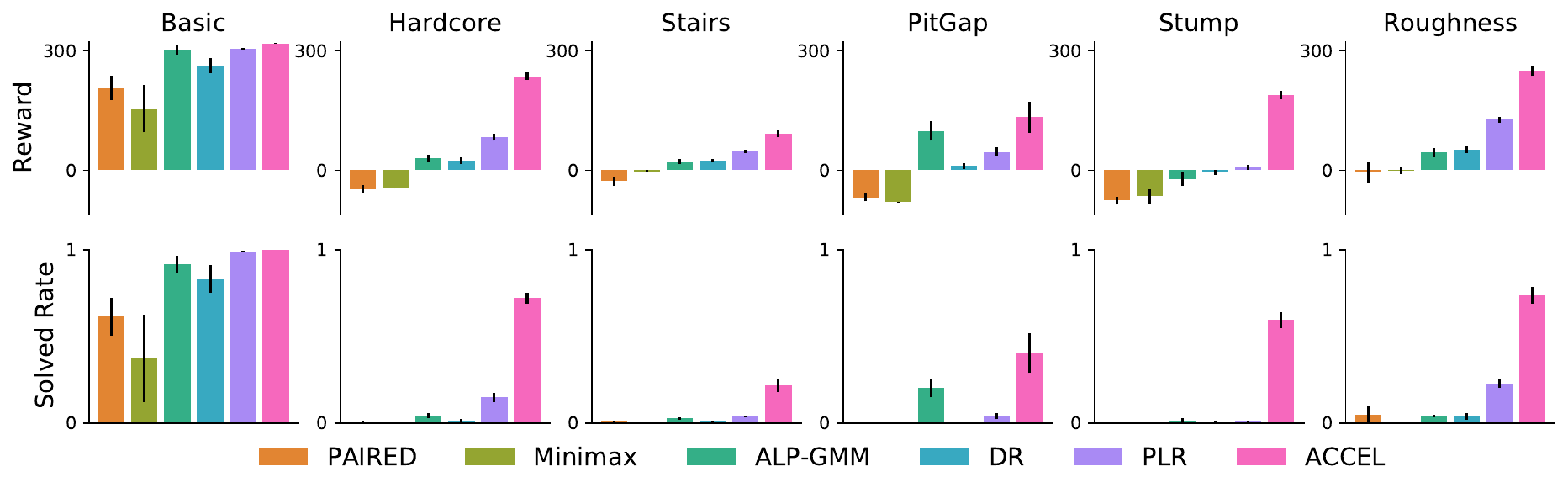}}
    \vspace{-2mm}
    \caption{Test results. Agents are evaluated for 100 episodes on a series of individual challenges, plots show mean and standard error for each environment, across ten runs.
    }
    \label{figure:mg_zs_results}
    \end{minipage}
\end{figure}

\textbf{POET Generated Levels} We also evaluated our agents on the six extremely challenging environments highlighted in the original POET paper. These represent some of the most difficult environment parameterizations produced by POET. Since each run of POET has a population of 20 agents, it is not clear if a single agent from any of their runs can solve more than one of these challenges.  In addition, POET only solves a single fixed seed of these environments. Instead, we report the mean performance over 100 samples for each given parameterization, for all 10 runs of \method{}. We report the mean and max performance across all training seeds and trials for each environment parameterization in Table~\ref{table:poet_roseplot_test}.

\begin{table}[H]
\begin{center}
\caption{Test performance on extremely challenging levels produced by POET. For each level, we run 100 trials with different random seeds. Mean shows the mean performance across all 10 ACCEL runs and trials. Max shows the best performance out of all runs and trials for each environment. 
}
\label{table:poet_roseplot_test}
\scalebox{0.87}{
\begin{tabular}{ l | cccccc } 
\toprule
 & 1a  & 1b & 2a & 2b & 3a & 3b   \\ 
\midrule
\textbf{Mean} & $0.01$ & $0.01$ & $0.00$ & $0.03$ & $0.01$ & $0.12$ \\
\textbf{Max} & $0.03$ & $0.05$ & $0.00$ & $0.08$ & $0.03$ & $0.31$ \\
\bottomrule
\end{tabular}}
\end{center}
\end{table}

\subsection{Testing the Limits of Current Approaches} 

\begin{figure}[h!]
    \centering
    \begin{minipage}{0.95\textwidth}
    \centering\subfigure{\includegraphics[width=.65\linewidth]{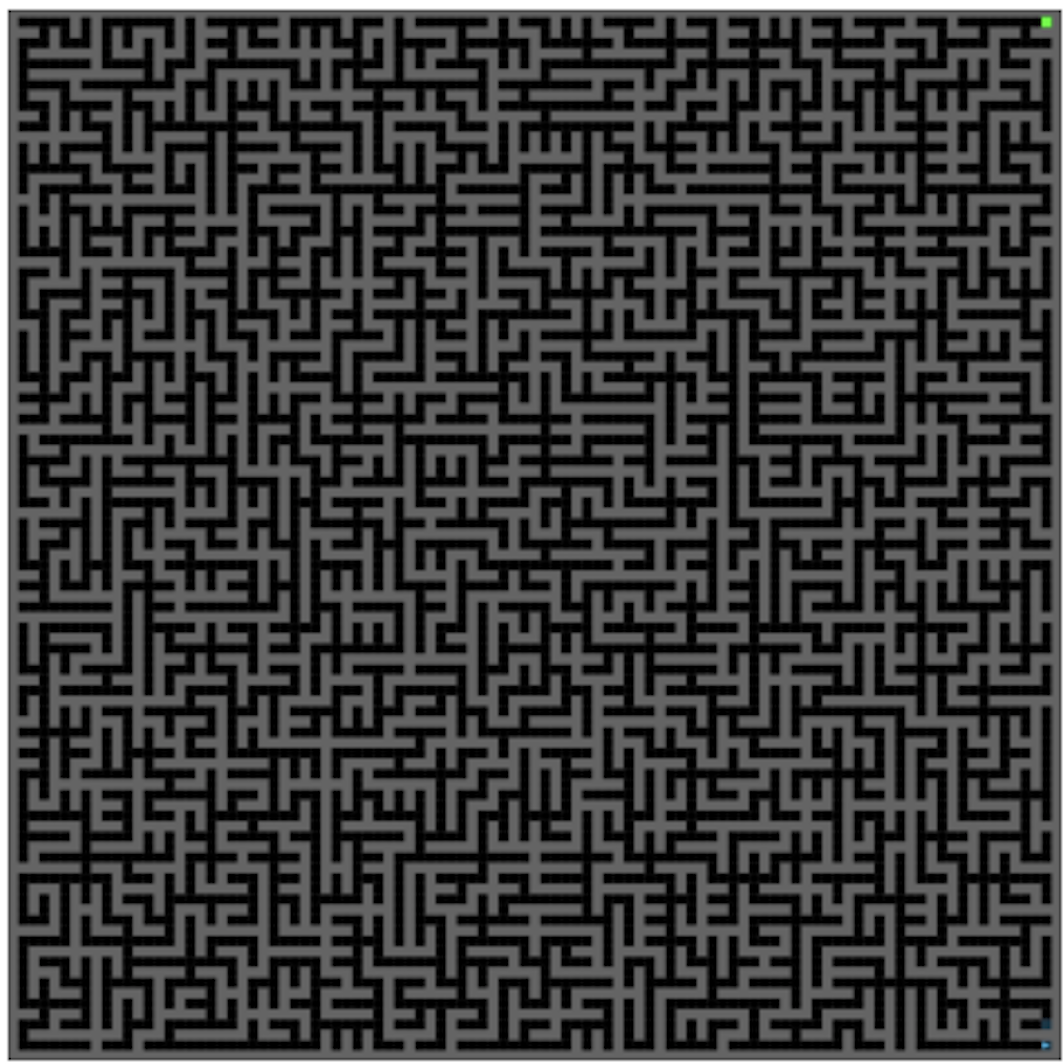}}
    \caption{PerfectMazesXL. A 101x101 procedurally-generated MiniGrid environment. The agents have to transfer zero-shot from training in a 15x15 grid. This environment is challenging even for humans, since the agent only has a partially observable view, it requires memorizing the current location at all times to ensure exploring all corners of the grid.}
    \label{figure:minigrid_zs_extralarge}
    \end{minipage}
\end{figure}

In Section~\ref{sec:experiments}, we showed the zero-shot performance for \method{} and baseline methods on a $51\times51$ procedurally-generated maze. \method{}, where \method{} saw over 50\% mean success rate across training runs. We further test \method{}, using both an empty generator and the typical DR generator, as well as DR and PLR, on an even larger 101x101 maze, shown in Figure~\ref{figure:minigrid_zs_extralarge}. Such a large partially-observable maze would be challenging even for humans. On this larger maze, the performance of all methods is significantly weaker, with DR and PLR achieving a mean success rate of 4\%. However, \method{} still outperforms all baselines, achieving 8\% and 7\% mean success rates when using the empty and DR generators respectively. 

\subsection{Additional Experiments}
\label{sec:additional_experiments}

In this section we present a series of additional experiments to better understand the performance of \method{}.

\textbf{Ablation Studies} To investigate the importance of various design choices for \method{}, we consider a series of ablations:
\begin{itemize}
    \item \textbf{ACCEL} The full \method{} method using a DR generator.
    \item \textbf{Learned Editor} Same as the first condition, but the editor uses RL to optimize an editing policy that seeks to maximize the positive value loss of the resulting levels.
    \item \textbf{No Editor} This is an ablation on the core editing mechanism, where replace the editing step in the first condition with simply sampling an equivalent number of additional levels from the DR generator.
\end{itemize}

\begin{table}[h!]
\begin{center}
\caption{Zero-shot transfer to human-designed environments. Each entry is the mean and standard error of five independent runs, where each run is evaluated for 100 trials on each environment. All methods use a DR generator that places between 0 and 60 blocks.
}
\label{table:minigridablations}
\scalebox{0.85}{
\begin{tabular}{ l | ccc } 
\toprule
\textbf{Test Environment} & ACCEL &Learned Editor & No Editor \\ 
\midrule 
16Rooms & $1.0 \pm 0.0$  & $0.9 \pm 0.07$ & $0.84 \pm 0.06$ \\
16Rooms2 &  $0.51 \pm 0.28$ & $0.41 \pm 0.19$ & $0.68 \pm 0.18$ \\
SimpleCrossing & $0.8 \pm 0.05$  & $0.9 \pm 0.1$ & $0.75 \pm 0.05$ \\
FourRooms &  $0.85 \pm 0.05$  & $0.88 \pm 0.04$ & $0.88 \pm 0.05$ \\
SmallCorridor &  $0.72 \pm 0.1$  & $0.6 \pm 0.17$ & $0.7 \pm 0.18$ \\
LargeCorridor &  $0.91 \pm 0.05$ & $0.56 \pm 0.18$ & $0.63 \pm 0.18$ \\
Labyrinth &  $0.98 \pm 0.02$ & $0.99 \pm 0.01$ & $0.67 \pm 0.19$ \\
Labyrinth2 &  $0.97 \pm 0.03$  & $0.7 \pm 0.15$ & $0.48 \pm 0.2$ \\
Maze &  $0.78 \pm 0.21$ & $0.57 \pm 0.18$ & $0.15 \pm 0.08$ \\
Maze2 &  $0.5 \pm 0.24$  & $0.65 \pm 0.15$ & $0.23 \pm 0.15$ \\
Maze3 & $0.79 \pm 0.14$  & $0.95 \pm 0.05$ & $0.56 \pm 0.17$ \\
\midrule
\textbf{Mean} & $0.79 \pm 0.04$  & $0.74 \pm 0.04$ & $0.58 \pm 0.05$ \\
\bottomrule
\end{tabular}}
\end{center}
\end{table}
Each condition is an ablation of the full ACCEL method using a DR generator, sampling levels from the DR distribution at the start of 10\% of new episode rollouts. We train each ablation for 10k PPO updates and evaluate each on the zero-shot maze environments. The results in Table \ref{table:minigridablations} show that using a learned editor that seeks to maximize the PVL of the resulting levels degrades zero-shot performance. Note however that each of these ablations, making use of editing, still outperform the next best baseline, PLR, which sees a mean solved rate of 0.69 over all zero-shot environments. Finally, the No Editor ablation performs worse than PLR, showing that \method{}'s strong performance derives from level editing.

\begin{figure}[h!]
    \centering
    \begin{minipage}{0.99\textwidth}
    \centering\subfigure{\includegraphics[width=.42\linewidth]{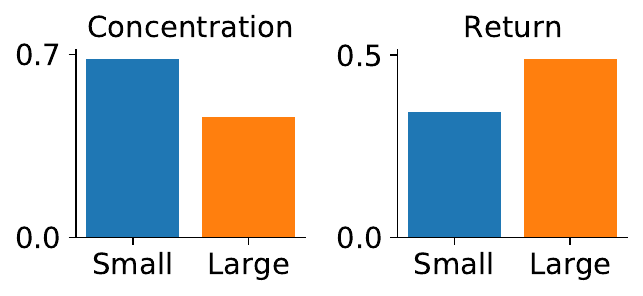}}
    \caption{Replay buffer diversity vs. return in the lava environment. On the left we show the concentration of the replay buffer, measured as the percentage of the top-100 high-regret levels that can be produced by just ten parents. On the right we compare the average return on ten-tile test environments. Here, \emph{small} corresponds to a buffer of size 4k, with no generator, while \emph{large} indicates a buffer of size 10k, using a generator 10\% of the time.}
    \label{figure:diversity}
    \end{minipage}
\end{figure}

\textbf{Diversity of the Level Buffer} we compare a buffer of size 4k with no DR sampling against our method with a buffer size of 10k and 10\% sampling. The plots show the proportion of the top 200 levels produced by just ten initial generator levels, with a significant increase for the smaller buffer. We also compare the performance of the two agents on test levels with ten tiles, showing clear outperformance for the lower concentration agent. It is likely that hyperparameters alone will not be sufficient if we want to scale \method{} to more open-ended domains, which we leave to future work.

\section{Implementation Details}

In this section we detail the training procedure for all our experiments. All training runs used a single V100 GPU and Intel Xeon E5-2698 v4 CPUs. Our ACCEL implementation directly builds on top of the Robust PLR codebase, available at \url{https://github.com/facebookresearch/dcd}. For PAIRED and Minimax in the maze environment, we report the results from \citet{jiang2021robustplr}. 

\subsection{Environment Details}
\label{sec:env_details}

\textbf{Learning with Lava} The MiniHack environment is an open-source Gym environment \citep{Gym}, which wraps the game of NetHack via the NetHack Learning Environment \citep{kuettler2020nethack}. MiniHack allows users (or agents) the ability to fully specify environments leveraging the full NetHack runtime. For our experiments we use a simple $7\times7$ grid and allow the agent to place lava tiles in any location. The DR agent samples the number of blocks in $[0,20]$. The reward is sparse, with the agent receiving $+1$ reward for reaching the goal and a per timestep penalty of $-0.01$.

\textbf{Partially-Observable Navigation} Each maze consists of a $15 \times 15$ grid, where each tile can contain a block, the goal, the agent, or navigable space. The agent receives a reward of $1-T/T_{\textnormal{max}}$ upon reaching the goal, where $T$ is the episode length and $T_{\textnormal{max}}$ is the maximum episode length (set to 250 at training). The agent receives a reward of 0 if it fails to reach the goal.

\textbf{BipedalWalker} We use a modified version of the \texttt{BipedalWalkerHardcore} environment from OpenAI Gym. The agent receives a 24-dimensional proprioceptive state corresponding to inputs from its lidar sensors, angles, and contacts. The agent does not have access to its positional coordinates. The action space is four continuous values that control the torques of its four motors. The environment design space is shown in Table \ref{table:bipedal_params}, where we show the value of the initial environment parameterization for \method{}, the edit size, and the maximum values. In this environment, the UED parameters correspond to the ranges of level properties, uniformly sampled to define each level. Thus, combined with a random seed, the UED parameters determine a specific level. Parameters for DR levels are sampled between zero and the maximum value. For PLR, we combine the environment parameterization with the specific seed of the sampled level, ensuring deterministic generation of the replayed level. \method{} makes each edit by randomly sampling one of the eight environment parameters and adding or subtracting the corresponding edit size listed in Table  \ref{table:bipedal_params} from the parameter value.

\begin{table}[H]
\begin{center}
\caption{Environment design space for the BipedalWalker environment. The UED parameters correspond to the min and max values for each level property. When a specific level is created, each property (i.e. obstacle size) is sampled from the corresponding range. 
}
\label{table:bipedal_params}
\scalebox{0.95}{
\begin{tabular}{ l ccccc } 
\toprule
& \textbf{Stump Height} &\textbf{Stair Height}  & \textbf{Stair Steps} & \textbf{Roughness} & \textbf{Pit Gap}  \\ 
\midrule
\textbf{Easy Init} & [0,0.4] & [0,0.4] & 1 & Unif(0, 0.6) & [0,0.8] \\
\textbf{Edit Size} & 0.2 & 0.2 & 1 & Unif(0, 0.6) & 0.4 \\
\textbf{Max Value} & [5,5] & [5,5] & 9 & 10 & [10,10] \\
\bottomrule
\end{tabular}}
\end{center}
\end{table}

\begin{table}[H]
\begin{center}
\caption{Total number of environment steps for a given number of student PPO updates.
}
\label{table:stepcount}
\scalebox{0.95}{
\begin{tabular}{ ll | cc } 
\toprule
\textbf{Environment} & \textbf{PPO Updates} & PLR & ACCEL \\ 
\midrule
MiniGrid & 20k & 327M & 369M \\
BipedalWalker & 30k & 1.96B & 2.07B \\
\bottomrule
\end{tabular}}
\end{center}
\end{table}

\subsection{Environment Design Procedure}

\textbf{Generating new levels} For a fair comparison to the PAIRED level generation procedure, DR is implemented by sampling a uniformly random teacher policy to output actions that set the environment parameters, thereby designing each level. Under PAIRED, this policy is no longer uniformly random, but rather optimized to maximize the estimated regret (e.g. PVL) incurred by the student agent on the resulting levels. The environment design procedure for the lava and maze domains is as follows: For each timestep the teacher receives an observation consisting of a map of the entire level and takes chooses a tile in the grid. For the first $N$ steps, where $N$ is teacher's budget of blocks (or lava tiles) the teacher always places a block (or lava tile). In the last two time steps, the teacher chooses a location for the goal and agent. This procedure reflects the approach taken in several recent works \citep{paired, plr, jiang2021robustplr, pcgrl}. For BipedalWalker, the teacher generates each level by choosing a random value between the minimum value of the ``Easy Init'' range in Table~\ref{table:bipedal_params} and the maximum value for each environment parameter. A random integer is then generated to seed the procedural content generation algorithm, which takes the sampled parameters to produce the level. 

\textbf{Editing levels} In lava levels, edits only add or remove obstacle tiles (i.e. lava or block tiles), while in MiniGrid mazes, edits can also alter the goal location. If an edit places a lava or block tile in the current goal or agent position, then the new tile replaces the goal or agent, which is randomly relocated  after applying all remaining edits. 

\subsection{Hyperparameters}
\label{sec:hparams}

The majority of our hyperparameters are inherited from previous works such as \citep{paired, plr, jiang2021robustplr}, with a few small changes. For the lava grid in MiniHack we use the agent model from the NetHack paper \citep{kuettler2020nethack}, using the \texttt{glyphs} and \texttt{blstats} as observations. The agent has both a global and a locally cropped view (produced using the coordinates in the \texttt{blstats}). 

For MiniHack we conduct a grid search across the level replay buffer size $\{4000,10000\}$ for both PLR and \method{}, and for \method{} we sweep across the edit method in \{random, positive value loss\}, where the latter option equates to a learned editor trained with RL to maximize the positive value loss. For MiniGrid we use the replay buffer size from \citet{jiang2021robustplr} and only conduct the \method{} grid search over the edit objective, again sweeping across \{random, positive value loss\} and replay rate from \{0.8, 0.9\}. For MiniGrid, we follow the protocol from \citet{jiang2021robustplr} and select the best hyperparameters using the validation levels \{16Rooms, Labyrinth, Maze\}. The final hyperparameters chosen are shown in Table \ref{table:hyperparams}.

For BipedalWalker we used the continuous control policy from the open source implementation of PPO from \citet{pytorchrl}, as well as many of the hyperparameters used in the recommended settings for MuJoCo. This involves a simple feedforward neural network with two hidden layers of size 64 and tanh activations. We tuned the hyperparameters for our base agent using domain randomization, and conducted a sweep over the learning rate \{3e-4, 3e-5\}, PPO epochs $\{5,20\}$, entropy coefficient \{0, 1e-3\} and number of minibatches $\{4,32\}$, using the validation performance on BipedalWalkerHardcore. We then used these base agent configurations for all UED algorithms. For PLR we further conducted a sweep over the buffer size $\{1000, 5000\}$, replay rate $\{0,9, 0.5\}$ and staleness coefficient $\{0.3, 0.5, 0.7\}$, using the same settings found for both PLR and \method{}. Finally, for \method{} we swept over number of edits in \{1, 2, 3, 4\}.

\begin{table}[h!]
\caption{Hyperparameters used for training each method in each environment.}
\label{table:hyperparams}
\begin{center}
\scalebox{0.88}{
\begin{tabular}{lccc}
\toprule
\textbf{Parameter} & MiniHack (Lava) & MiniGrid & BipedalWalker \\
\midrule
\emph{PPO} & \\
$\gamma$ & 0.995  & 0.995 & 0.99 \\
$\lambda_{\text{GAE}}$ & 0.95  & 0.95 & 0.9 \\
PPO rollout length & 256  & 256 & 2000 \\
PPO epochs & 5  & 5 & 5 \\
PPO minibatches per epoch & 1 & 1 & 32 \\
PPO clip range & 0.2 & 0.2 & 0.2 \\
PPO number of workers & 32 & 32 & 16 \\
Adam learning rate & 1e-4  & 1e-4 & 3e-4 \\
Adam $\epsilon$ & 1e-5 & 1e-5 & 1e-5 \\
PPO max gradient norm & 0.5 & 0.5 & 0.5 \\
PPO value clipping & yes & yes & no \\
return normalization & no  & no & yes \\
value loss coefficient & 0.5  & 0.5 & 0.5 \\
student entropy coefficient & 0.0  & 0.0 & 1e-3 \\
generator entropy coefficient & 0.0 & 0.0 & 0.0 \\

\addlinespace[10pt]
\emph{\method{}} & & \\
Edit rate, $q$ & 1.0 & 1.0 & 1.0  \\
Replay rate, $p$ & 0.9 & 0.8 & 0.9 \\
Buffer size, $K$ & 10000 & 4000 & 1000 \\
Scoring function & positive value loss & positive value loss & positive value loss \\
Edit method & positive value loss & random & random \\
Number of edits & 5 & 5 & 3 \\
Prioritization & rank  & rank & rank \\
Temperature, $\beta$ & 0.3  & 0.3 & 0.1 \\
Staleness coefficient, $\rho$ & 0.3 & 0.3 & 0.5 \\

\addlinespace[10pt]
\emph{PLR} & & \\
Scoring function & positive value loss & positive value loss  & positive value loss\\
Replay rate, $p$ & 0.5 & 0.5 & 0.5 \\
Buffer size, $K$ & 10000 & 4000 & 1000 \\

\bottomrule 
\end{tabular}
}
\end{center}
\end{table}

\end{document}